\documentclass{article}

\usepackage[final]{corl_2022} 

\title{Learning Riemannian Stable Dynamical Systems\\ via Diffeomorphisms}

\author{
     Jiechao Zhang\textsuperscript{\ensuremath{1,2}}
     \quad
     Hadi Beik-Mohammadi\textsuperscript{\ensuremath{1,2}}
     \quad
     Leonel Rozo\textsuperscript{\ensuremath{1}}
 	\\
 	\textsuperscript{\ensuremath{1}}Bosch Center for Artificial Intelligence. Renningen, Germany.
 	\\
 	\textsuperscript{\ensuremath{2}.} Karlsruhe Institute of Technology (KIT), Karlsruhe, Germany.
 	\\
 	\href{mailto:leonel.rozo@de.bosch.com}{\textrm{leonel.rozo@de.bosch.com}} 
 	\quad
 	\href{mailto:hadi.beik-mohammadi@de.bosch.com}{\textrm{hadi.beik-mohammadi@de.bosch.com}}
}

\usepackage{amsmath} 
\usepackage{amssymb}  
\usepackage{amsthm}
\usepackage{siunitx}
\usepackage{adjustbox}
\usepackage{bm}  
\usepackage{graphics} 
\usepackage{epsfig} 
\usepackage{wrapfig}
\usepackage{subcaption}
\usepackage{caption}
\usepackage{subcaption}
\usepackage[font=small,labelfont=small]{caption}
\usepackage{color}
\usepackage[ruled, linesnumbered, vlined]{algorithm2e}
\usepackage{sidecap}
\usepackage{tabularx}
\usepackage{booktabs} 

\newtheorem{theorem}{Theorem}

\graphicspath{{Images/}}

\newcommand{\SphereManifold}{\mathcal{S}}
\newcommand{\RtimeS}{\mathbb{R}^3 \times \mathcal{S}^3}
\newcommand{\trsp}{\mathsf{T}} 
\newcommand{\euclideanspace}{\mathbb{R}}
\newcommand{\manifold}{\mathcal{M}}
\newcommand{\tangentspace}[1]{\mathcal{T}_{#1}\mathcal{M}}
\newcommand{\expmap}[2]{\text{Exp}_{#1}(#2)}  
\newcommand{\logmap}[2]{\text{Log}_{#1}(#2)}  
\newcommand{\prltrsp}[3]{\Gamma_{#1 \rightarrow #2}\big(#3\big)}  
\newcommand{\diffeomorphism}{\psi}
\newcommand{\proju}{\operatorname{proju}}
\newcommand{\manifolddist}[2]{d_{\manifold}(#1, #2)}
\begin{document}
\maketitle


\begin{abstract}
Dexterous and autonomous robots should be capable of executing elaborated dynamical motions skillfully.
Learning techniques may be leveraged to build models of such dynamic skills.
To accomplish this, the learning model needs to encode a stable vector field that resembles the desired motion dynamics. 
This is challenging as the robot state does not evolve on a Euclidean space, and therefore the stability guarantees and vector field encoding need to account for the geometry arising from, for example, the orientation representation.
To tackle this problem, we propose learning Riemannian stable dynamical systems (RSDS) from demonstrations, allowing us to account for different geometric constraints resulting from the dynamical system state representation.
Our approach provides Lyapunov-stability guarantees on Riemannian manifolds that are enforced on the desired motion dynamics via diffeomorphisms built on neural manifold ODEs.
We show that our Riemannian approach makes it possible to learn stable dynamical systems displaying complicated vector fields on both illustrative examples and real-world manipulation tasks, where Euclidean approximations fail. 
\end{abstract}



\section{Introduction}
The promise of having fully-autonomous robots performing a large variety of tasks implies that robots should be able to execute highly-dynamic motions. 
The inherent complexity of these movements makes hand coding an infeasible approach. Therefore learning techniques arise as a potential solution.
In particular, learning dynamic robot motions from human demonstrations is a promising approach to build models of dynamic skills in an intuitive, sample-efficient and quick manner.
However, learning dynamical motions is not trivial as the learning model requires to provide stability guarantees, which is also an intrinsic property in human motion generation~\citep{Burdet06:StabilityHumanMotion}. 
In this context, most research works have focused on learning time-invariant stable dynamical systems for goal-driven motions (a.k.a point-to-point movements) with Lyapunov-stability guarantees~\citep{KhansariZadeh2011:StableEstimatorDS,Khansari-Zadeh2014:lyapunovStabilityDS,Neumann2014:SDS_diffeomorphism,Rana2020:EuclideanizingFlows,Zhang2022:accurateSDS}. 

\citet{KhansariZadeh2011:StableEstimatorDS} proposed one of the first approaches to learn stable dynamical systems from human demonstrations by imposing quadratic Lyapunov non-linear constraints on the model parameters' optimization, which limited the range of possible learnable motions.
As the class of stable dynamical systems constrained by a predefined Lyapunov function is a subset of all possible stable dynamical systems, this limits the learned model accuracy~\citep{Neumann2014:SDS_diffeomorphism}. 
To improve accuracy, a more general parametric control Lyapunov function~\citep{Sontag83:CLF} can be learned from demonstrations~\citep{Khansari-Zadeh2014:lyapunovStabilityDS, Ravanbakhsh19:CLFwithLfD, Rodriguez2022:LyaNet}.
The trade-off between stability and accuracy motivated the use of diffeomorphisms~\citep{Neumann2014:SDS_diffeomorphism, Rana2020:EuclideanizingFlows, Zhang2022:accurateSDS, Urain20:ImitationFlow, Urain2021:LearningVFonLieGroups, Zhi22:DiffeoImitation}, which leveraged a more general class of stable dynamical systems.
Their main idea is to design or learn a canonical Lyapunov-stable dynamical system on a latent space and use a diffeomorphic mapping to transform the demonstrations to the latent space so that they are consistent with the desired Lyapunov-stable behavior.
Thus, the modeling accuracy depends on the expressiveness of the diffeormorphic function, often modeled by a neural network
~\citep{Rana2020:EuclideanizingFlows,Zhang2022:accurateSDS,Urain20:ImitationFlow,Urain2021:LearningVFonLieGroups}.

Most of aforementioned works assume that the training data lie in the Euclidean space~\citep{Neumann2014:SDS_diffeomorphism, Rana2020:EuclideanizingFlows, Zhang2022:accurateSDS, Urain20:ImitationFlow,Zhi22:DiffeoImitation}, with the exception of~\citet{Urain2021:LearningVFonLieGroups}, which severely limits their use in real applications.
For instance, there are various types of representation for the robot end-effector's orientation, namely unit quaternions in the $3$-Sphere manifold~\citep{Wen1991:Quaternion}, and rotation matrices in the special orthogonal group ($\operatorname{SO}(3)$) manifold~\citep{Gu1988:OrientationRep}, which do not lie in the Euclidean space.
Accounting for the data geometry has proven critical when learning and optimizing movement primitives on quaternion space~\citep{Ude14ICRA,Zeestraten17:Riemannian,Koutras20:OrientationDMP,Rozo2021:OrientationProMP,Jaquier21:MaternGaBO}, as relying on Euclidean approximations leads to modeling distortions and compromises extrapolation.   
The importance of geometry-aware methods when learning dynamical systems was recently addressed in~\citep{Urain2021:LearningVFonLieGroups}, where a stable dynamical system was learned via diffeomorphisms over Lie groups.  
Although Lie theory has been exploited to operate with data of specific geometries~\citep{Sola18:LieTheory}, a potential limitation is that not all types of manifolds arising in robotics can be easily endowed with a Lie group structure (e.g., the space of symmetric positive-definite matrices (SPD)). 

A more general solution based on Riemannian geometry~\citep{DoCarmo92:RiemannManifold} is proposed in this paper. 
We consider dynamical systems evolving on a Riemannian manifold.
This arises two main challenges: \emph{(1)} designing a canonical Lyapunov-stable dynamical system on Riemannian manifolds, and \emph{(2)} learning a diffeomorphism that accounts for the Riemannian geometry.
To address these challenges, we leverage the Lyapunov stability analysis on Riemannian manifolds~\citep{Pait10:RiemannianLyapunov,Jaquier2021:ManipulabilityLearning}.  
Moreover, we exploit neural ordinary differential equations (ODEs) on Riemannian manifolds~\citep{lou2020:NeuralMODE, mathieu2020:RCNormalizingFlows} for constructing the diffeomorphism to learn a Riemannnian stable dynamical system (RSDS).
Unlike previous works using diffeomorphisms~\citep{Neumann2014:SDS_diffeomorphism, Rana2020:EuclideanizingFlows, Zhang2022:accurateSDS}, our approach extends this concept to systems evolving on Riemannian manifolds.
In contrast to works assuming Riemannian manifolds that are diffeomorphic to the Euclidean space~\citep{Gemici16:NFonRiemannianManif}, or manifold-specific diffeomorphisms built on specialized neural networks~\citep{Rezende20:NFsToriSphere}, our approach leverages a general formulation to construct diffeomorphisms based on solutions of ODEs evolving on arbitrary Riemannian manifolds~\citep{lou2020:NeuralMODE}.
Our approach is conceptually similar to the Lie-group method introduced in~\citep{Urain2021:LearningVFonLieGroups}, as both explicitly consider the data geometry to design technically-sound learning models via diffeomorphisms. 
However, our Riemannian formulation substantially differs from~\citep{Urain2021:LearningVFonLieGroups} in its technical development, and provides a more general approach that may be exploited for a variety of Riemannian manifolds. 

In summary, we propose a method to learn Riemannian stable dynamical systems from demonstrations. 
Our approach provides Lyapunov-stability guarantees on Riemannian manifolds (see \S~\ref{subsec:GeodesicVF}) that are enforced on the desired motion dynamics via diffeomorphisms built on neural manifold ODEs (see~\S~\ref{subsec:DiffeomorphRM} and~\ref{subsec:DiffInvDiffeomorph}).  
Through a set of evaluations on the $2$-sphere manifold $\SphereManifold^2$, presented in \S~\ref{sec:result}, we show that our Riemannian approach is able to learn complicated dynamical systems, in contrast to Euclidean approximations which fail to encode stable vector fields.
Also, we learn realistic motion skills on a $7$-DoF robotic manipulator featuring complex full-pose trajectories on $\RtimeS$.

\section{Background}
\label{sec:background}

\subsection{Dynamical Systems and Lyapunov Stability}
\label{subsec:DSlyapunov}
We here give a short review of Lyapunov stability in the Euclidean setting.
Let us assume an autonomous dynamical system $\dot{\bm{x}} = f(\bm{x})$, with a single equilibrium point $\bm{x}^*$, where $\bm{x} \in \mathbb{R}^n$ is the state variable.
Consider a potential function $V(\bm{x}(t))$ describing the energy of such a system.
If this system loses energy over time and the energy is never restored, the system must eventually reach some final resting state.
This idea is formally described as (see~\citep{Slotine91:AppliedNonlinearCtrl} for details):
\begin{theorem}[Lyapunov Stability]
\label{th:lyapunov_stability}
A dynamical system $\dot{\bm{x}} = f(\bm{x})$ is globally asymptotically stable at $\bm{x}^*$ if there exists a continuously differentiable Lyapunov function $V(\bm{x}): \mathbb{R}^n \rightarrow \mathbb{R}$ such that 
\begin{equation}
        V(\bm{x}^*) = 0 , \quad \dot{V}(\bm{x}^*) = 0 , \quad V(\bm{x}) > 0 ,\  \forall \ \bm{x} \neq \bm{x}^* , \quad  \dot{V}(\bm{x}) < 0 , \ \forall \  \bm{x} \neq \bm{x}^* .
    \label{eq:lyapunov_conditions}
\end{equation}
\end{theorem}
From Theorem~\ref{th:lyapunov_stability} we know that we can always find a Lyapunov function that fulfills these four conditions in Eq.~\ref{eq:lyapunov_conditions} for a globally asymptotically stable dynamical system.
\vspace{.5cm}

\subsection{Riemannian Manifolds}
\label{subsec:RiemanianManif}
A smooth manifold $\manifold$ can be seen as a set of points that locally, but not globally, resemble the Euclidean space $\euclideanspace^d$~\citep{DoCarmo92:RiemannManifold,Lee18:RiemannManifold}. 
An abstract definition of a manifold specifies the topological, differential and geometric structure by using \emph{charts}, which are maps between parts of $\manifold$ to $\euclideanspace^d$.
The collection of these charts (a.k.a. local parameterizations) is called \emph{atlas}.
More formally, a chart on a smooth manifold $\manifold$ is a diffeomorphic mapping (i.e. a bijective and differentiable function) $\varphi: U \to \tilde{U}$ from an open set $U \subset \manifold$ to an open set $\tilde{U} \subseteq \euclideanspace^d$ (see Fig.~\ref{fig:coordinate_chart} in App.~\ref{app:RiemannianManif}).
Moreover, the transition map between two intersecting sets $U_1$ and $U_2$, given by $\varphi_1 \circ \varphi_2^{-1}$ or $\varphi_2 \circ \varphi_1^{-1}: \mathbb{R}^d \rightarrow \mathbb{R}^d$ is also a diffeomorphism.
The smooth structure of $\manifold$ makes it possible to take derivatives of curves on the manifold, leading to tangent vectors in $\euclideanspace^d$.
The set of tangent vectors of all curves at $\bm{x} \in \manifold$ spans a $d$-dimensional affine subspace of $\euclideanspace^d$, which is known as the \emph{tangent space} $\tangentspace{\bm{x}}$ of $\manifold$ at $\bm{x}$.
The collection of all tangent spaces of $\manifold$ is the \emph{tangent bundle} $\tangentspace{} = \bigsqcup_{\bm{x} \in \manifold} \tangentspace{x}$.
Therefore, a velocity vector $\dot{\bm{x}}$ at $\bm{x} \in \manifold$ lies on $\tangentspace{x}$, and consequently a vector field on $\manifold$ lies on $\tangentspace{}$.

The above definitions do not provide the mechanisms to measure how curved $\manifold$ is, or to compute distances on $\manifold$. 
To do so, we can endow $\manifold$ with a \emph{Riemannian metric}, which is a family of inner products $g_{\bm{x}}: \tangentspace{\bm{x}} \times  \tangentspace{\bm{x}} \rightarrow \euclideanspace$ associated to each point $\bm{x} \in \manifold$.
As a result, a \emph{Riemannian manifold} $(\manifold, g)$ is a smooth manifold endowed with a Riemannian metric~\citep{Lee18:RiemannManifold}. 
To operate with Riemannian manifolds, it is common practice to exploit the Euclidean tangent spaces. 
To do so, we resort to mappings back and forth between $\tangentspace{\bm{x}}$ and $\manifold$ using the exponential and logarithmic maps.
The exponential map $\expmap{\bm{x}}{\bm{u}}: \tangentspace{\bm{x}} \to \manifold$ maps a point $\bm{u}$ in the tangent space of $\bm{x}$ to a point $\bm{y}$ on the manifold, so that it lies on the geodesic starting at $\bm{x}$ in the direction $\bm{u}$, and such that the geodesic distance $d_{\manifold}(\bm{x}, \bm{y}) = d_{\euclideanspace}(\bm{x}, \bm{u})$. 
The inverse operation is the logarithmic map $\logmap{\bm{x}}{\bm{y}}: \manifold \to \tangentspace{\bm{x}}$.
We provide all the necessary operations in App.~\ref{app:RiemannianManif}.
\vspace{.5cm}

\subsection{Diffeomorphism}
\label{subsec:Diffeomorphism}
A diffeomorphism $\diffeomorphism: \manifold \rightarrow \mathcal{N}$ is a smooth bijective mapping between two smooth manifolds which preserves the topological properties of $\manifold$, and whose inverse $\diffeomorphism^{-1}$ is also smooth.
When learning stable dynamical systems, diffeomorphisms can be exploited to impose Lyapunov stability guarantees by transferring a manually-designed stable dynamical system on $\mathcal{N}$ to the desired manifold $\manifold$. 
We focus on constructing learnable diffeomorphisms that resemble continuous normalizing flows (CNFs)~\citep{chen2018:NeuralODE,Grathwohl19:FFJORD,Finlay2020:JacobianRegularization}, which are bijective and bidirectionally differentiable mappings, and have been recently exploited on density estimation problems~\citep{Rezende2015:VIwithNF,Papamakarios21:NormalizingFlows,Kobyzev2021:IntroductionNFs}.
We here exploit them to learn diffeomorphic mappings between Riemannian manifolds. 

Most CNFs are constructed from \emph{neural ordinary differential equation} (Neural ODEs) in Euclidean space~\citep{chen2018:NeuralODE,Grathwohl19:FFJORD,Finlay2020:JacobianRegularization}, with the exception of \emph{neural manifold ordinary differential equations} (Neural MODEs) on Riemannian manifolds~\citep{lou2020:NeuralMODE, mathieu2020:RCNormalizingFlows}.
Generally, Neural ODEs parametrize the dynamics of a hidden variable using a continuous-time ODE represented by a neural network, as follows,
\begin{equation}
    \dot{\bm{z}}(t) = f_{\bm{\theta}}(\bm{z}(t), t) ,
    \label{eq:neuralode}
\end{equation}
where $\bm{z} \in \euclideanspace^d$ is the state variable and $f_{\bm{\theta}}: \euclideanspace^d \times \euclideanspace \rightarrow \euclideanspace^d$ is a neural network.
According to~\citet{mathieu2020:RCNormalizingFlows} (see Theorem~\ref{th:vector_flows} in App.~\ref{app:NeuralMODEs}), we can extend CNFs to the Riemannian setting, where the state variable $\bm{z} \in \mathcal{M}$ and the vector field $f_{\bm{\theta}}: \mathcal{M} \times \mathbb{R} \rightarrow \tangentspace{}$.
As a result, we can use~\eqref{eq:neuralode} as a manifold ODE, whose initial value problem (IVP) solution results in a diffeormorphic mapping $\psi_{\bm{\theta}}: \manifold \rightarrow \mathcal{N}, \ \bm{x} = \bm{z}(t_s) \in \manifold$ and $\bm{y}=\bm{z}(t_e) \in \mathcal{N}$.
i.e. $\bm{y} = \psi_{\bm{\theta}}(\bm{x}) = \bm{x} + \int_{\tau=t_s}^{t_e} f_{\bm{\theta}}(\bm{z}(\tau), \tau) d\tau$.
To solve the IVP on $\manifold$, we leverage integrators on Riemannian manifolds based on the local representation via coordinate charts~\citep{hairer2011:ODEManifolds}.
This method uses a local representation of $\mathcal{M}$ defined by a coordinate map $\varphi: \mathcal{M} \supseteq U \rightarrow \tilde{U} \subseteq \mathbb{R}^d$ with coordinates $\bm{w}(t) = \varphi(\bm{z}(t))$. 
Computing integrators on $\manifold$ can be approximated by solving an equivalent ODE in $\euclideanspace^d$
\begin{equation}
    \dot{\bm{w}}(t) = D_{\varphi^{-1}(\bm{w}(t))}\varphi \circ f_{\bm{\theta}}(\varphi^{-1}(\bm{w}(t)), t) ,
    \label{eq:equivalentode}
\end{equation}
where $D_{\varphi^{-1}(\bm{w}(t))}\varphi$ represents the differential of $\varphi$ at $\varphi^{-1}(\bm{w}(t))$
(see App.~\ref{app:NeuralMODEs} and~\ref{app:integrators_manifolds} for details).
Additionally, we use the adjoint method~\citep{chen2018:NeuralODE} on Riemannian manifolds~\citep{lou2020:NeuralMODE} to compute gradients, which can also be used for calculating differentials. Consider a loss function $\mathcal{L} : \mathcal{M} \rightarrow \mathbb{R}$, in order to compute the derivative of $\mathcal{L}$ with respect to an intermediate variable $\boldsymbol{z}(t)$ of the manifold ODE, we can solve $\dot{\bm{a}}(t)^\trsp = - \bm{a}(t)^\trsp D_{\bm{z}(t)}  f_{\bm{\theta}}(\bm{z}(t), t)$, where $\bm{a}(t)^\trsp := D_{\bm{z}(t)} \mathcal{L}$  (as detailed in App.~ \ref{app:adjoint_method}).
\vspace{2cm}


\section{Learning Riemannian Stable Dynamical Systems}
\label{sec:approach}
\begin{wrapfigure}[13]{r}{0.46\linewidth}
    \centering
    \includegraphics[trim={0.0cm 0.0cm 0.0cm .5cm}, width=0.44\textwidth, ]{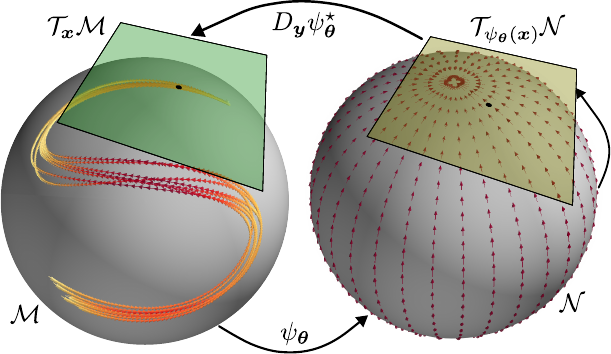}
    \caption{Architecture of a diffeomorphism-based stable vector field on the Riemannian manifold $\SphereManifold^2$.}
    \label{fig:diffeomorphism_based_svf}
\end{wrapfigure}

We here introduce our approach for learning stable dynamical systems on Riemannian manifolds from demonstrated point-to-point motions.
First, let us consider that the recorded demonstrations follow a dynamical system $\Dot{\bm{x}} = f(\bm{x})$, where the state $\bm{x}$ evolves on a Riemannian manifold $\mathcal{M}$ with velocity $\dot{\bm{x}} \in \tangentspace{\bm{x}}$.
This dynamical system is equivalent, under a change of coordinates, to another system defined on a latent Riemannian manifold $\mathcal{N}$.
Under the diffeomorphism $\psi_{\bm{\theta}}: \bm{x} \mapsto \bm{y} \in \mathcal{N}$, parameterized by $\bm{\theta}$, we map the observed states $\bm{x}$ onto $\mathcal{N}$.
Then, we evaluate the canonical stable vector field $g_{\bm{\gamma}}(\bm{y})$ to obtain the velocity $\dot{\bm{y}} \in \mathcal{T}_{\bm{y}} \mathcal{N}$.
Finally, we leverage the \emph{pullback operator} $D_{\bm{y}}\psi_{\bm{\theta}}^\star$ to project $\dot{\bm{y}}$ back to the tangent space $\tangentspace{\bm{x}}$.
The whole procedure can be expressed as follows, 
\begin{equation}
    \dot{\bm{x}} = (D_{\bm{y}}\psi_{\bm{\theta}}^\star \circ g_{\bm{\gamma}} \circ \psi_{\bm{\theta}})(\bm{x}) \\ =  D_{\bm{y}}\psi_{\bm{\theta}}^\star (\dot{\bm{y}}) ,
    \label{eq:diffeomorphism_based_svf_equation}
\end{equation}
which is illustrated in Fig.~\ref{fig:diffeomorphism_based_svf} (a proof is given in App.~\ref{app:stabilityAnalysis}).
In the sequel, we describe how we design a Lyapunov-stable vector field $g_{\bm{\gamma}}$ on $\mathcal{N}$ to provide stability guarantees on the learned dynamical system.
Later, we explain how to compute the diffeomorphism between the target and latent manifolds.
Finally, we introduce two different methods to compute the pullback operator $D_{\bm{y}}\psi_{\bm{\theta}}^\star$.
\vspace{-0.15cm}

\subsection{Lyapunov-stable Geodesic Vector Fields}
\label{subsec:GeodesicVF}
To design a stable vector field on the Riemannian manifold $\mathcal{N}$, we enforce the canonical dynamical system to follow geodesic curves that converge to a single equilibrium.
Such a vector field can be designed via the logarithmic map.
Specifically, given an equilibrium point $\bm{y}^* \in \mathcal{N}$, the corresponding velocity vector $\dot{\bm{y}} \in \mathcal{T}_{\bm{y}}\mathcal{N}$ can be computed as $\dot{\bm{y}} = g_{\bm{\gamma}}(\bm{y}) = k_{\bm{\gamma}}(\bm{y}) g_n(\bm{y})$ with the normalized geodesics vector field $ g_n(\bm{y}) := \frac{\operatorname{Log}_{\bm{y}}(\bm{y}^*)}{\lVert \operatorname{Log}_{\bm{y}}(\bm{y}^*) \rVert_2}$.
This implies that the direction of tangent vectors is fully specified by $\operatorname{Log}_{\bm{y}}(\bm{y}^*)$, while their magnitude depends on the scaling factor $k_{\bm{\gamma}}: \mathbb{R}^n \supset \mathcal{N} \rightarrow \mathbb{R}_{\geq 0}$.
We can prove the stability of this geodesic vector field by choosing the Lyapunov function $V(\bm{y}):=\langle F, F \rangle_{\bm{y}^*}$ with $F = \logmap{\bm{y}^*}{\bm{y}}$, and applying Theorem~\ref{th:stability_geodesics_vf} for Lyapunov stability on Riemannian manifolds, as detailed in App.~\ref{app:stabilityAnalysis}.
Given that our geodesic vector field is Lyapunov stable, we can easily prove that the desired dynamical system is also globally asymptotically stable by defining a new valid Lyapunov function $\tilde{V}(\bm{x}) := V(\psi_{\bm{\theta}}(\bm{x}))$ via the diffeomorphism $\psi_{\bm{\theta}}$, with a single equilibrium point $\bm{x}^* = \psi_{\bm{\theta}}^{-1}(\bm{y}^*) \in \mathcal{M}$.
As $\psi_{\bm{\theta}}$ preserves the topological properties of $\mathcal{N}$, the equilibrium point $\bm{x}^*$ is also globally asymptotically stable on $\mathcal{M}$ (see App.~\ref{app:stabilityAnalysis} for the proof).
Note that for certain Riemannian manifolds, it is only possible to guarantee \emph{quasi-global} stability guarantees due to the Poincaré-Hopf theorem (see App.~\ref{app:stabilityAnalysis} for details).

Note that we separate the parameterization for the magnitude and direction of vector fields to improve the expressiveness of our framework.
By relocating the scaling factor $k_{\bm{\gamma}}$ and normalizing the vector fields governed by~\eqref{eq:diffeomorphism_based_svf_equation}, we can obtain our final RSDS learning framework
\begin{equation}
    \dot{\bm{x}} = \hat{k}_{\bm{\gamma}}(\bm{x}) \frac{(D_{\bm{y}}\psi_{\bm{\theta}}^\star \circ  g_n \circ \psi_{\bm{\theta}})(\bm{x})}{\lVert (D_{\bm{y}}\psi_{\bm{\theta}}^\star \circ  g_n \circ \psi_{\bm{\theta}})(\bm{x}) \rVert_2} ,
    \label{eq:final_diffeomorphism_based_vf_equation}
\end{equation}
where $\hat{k}_{\bm{\gamma}}(\bm{x})$ is the new positive scaling factor that fully determines the magnitude of the learned vector fields. 
In App.~\ref{app:finalFramework}, we prove that the models~\eqref{eq:diffeomorphism_based_svf_equation} and \eqref{eq:final_diffeomorphism_based_vf_equation} are equivalent.

\subsection{Diffeomorphisms on Riemannian Manifolds}
\label{subsec:DiffeomorphRM}
Given the final RSDS in~\eqref{eq:final_diffeomorphism_based_vf_equation} and $M$ demonstrations, the goal of learning stable dynamics on a Riemannian manifold reduces to learning $\psi_{\bm{\theta}}$, computing its pullback operator $D_{\bm{y}}\psi_{\bm{\theta}}^\star$, and subsequently estimating $\hat{k}_{\bm{\gamma}}(\bm{x})$. 
However, due to the geometric constraints arising from $\mathcal{M}$, learning a diffeomorphism and calculating the corresponding pullback operator are non-trivial problems. 
To address them, we leverage Neural MODEs~\citep{lou2020:NeuralMODE} to build the diffeomorphism $\psi_{\bm{\theta}}$. 
Unlike~\citep{lou2020:NeuralMODE}, we propose a novel approach to compute the pullback operator by reversing the time interval of the ODE integration (see \S~\ref{subsec:DiffInvDiffeomorph}), avoiding to explicitly compute the corresponding inverse. 
We also propose a method to design Lyapunov-stable geodesic vector fields on a Riemannian manifold, which are leveraged to provide stability guarantees on the learned dynamical system, as explained in \S~\ref{subsec:GeodesicVF}.

According to Theorem~\ref{th:vector_flows} in App.~\ref{app:NeuralMODEs}, the dynamics $f_{\bm{\theta}}$ of Neural MODEs only has to be a $\mathcal{C}^1$ function.
To compute the diffeomorphism with a parametric Neural MODE, we solve an integration problem based on the local parameterization $\bm{w}(t)= \varphi(\bm{z}(t))$ (described in App.~\ref{app:integrators_manifolds}).
Using this method requires the selection of coordinate charts, which can be created via the exponential map $\varphi_i^{-1} = \operatorname{Exp}_{\bm{z}_i}$ and logarithmic map $\varphi_i = \operatorname{Log}_{\bm{z}_i}$, similarly to~\citep{lou2020:NeuralMODE}.
Under this choice of coordinate mapping and given a fixed number of charts $k$, the diffeomorphism $\psi_{\bm{\theta}}: \bm{x} = \bm{z}_0 \mapsto \bm{z}_k = \bm{y}$, obtained via integration on the manifold can be then viewed as the composition of blocks of solving Neural ODEs and chart transitions defined as,
\begin{equation}
    \begin{split}
    \psi_{\bm{\theta}} = \operatorname{Exp}_{\bm{z}_{k-1}} \circ \hat{\psi}_{\bm{\theta},k-1} \circ \operatorname{Log}_{\bm{z}_{k-1}} \circ \ldots \circ \operatorname{Exp}_{\bm{z}_0} \circ \hat{\psi}_{\bm{\theta},0} \circ \operatorname{Log}_{\bm{z}_0} , \quad \textrm{with} \\
    \hat{\psi}_{\bm{\theta}, i}(\bm{w}_i(t_{i, s})) = \bm{w}_i(t_{i, s}) + \int_{\tau=t_{i, s}}^{t_{i, e}} D_{\varphi_i^{-1}(\bm{w}_i(\tau))}\varphi_i \circ f_{\boldsymbol{\theta}}(\varphi_i^{-1}(\bm{w}_i(\tau)), \tau) d\tau ,
    \end{split}
    \label{eq:dynamic_chart_composition}
\end{equation}
where $i$ is the chart index, $t_{i, s}$ and $t_{i, e}$ are the starting and end time for $i^{th}$ chart. 
$\hat{\psi}_{\bm{\theta}, i}$ defines a diffeomorphism computed by the classical ODE solver on the tangent space (i.e. Euclidean space) and provides the solution of the IVP of the equivalent ODE~\eqref{eq:equivalentode}. 

\subsection{Differential of the Inverse Diffeomorphism}
\label{subsec:DiffInvDiffeomorph}
We are now left with the problem of computing the pullback operator $D_{\bm{y}}\psi_{\bm{\theta}}^\star$ in~\eqref{eq:diffeomorphism_based_svf_equation}, which maps the latent velocity $\dot{\bm{y}}$ back to the original tangent space $\tangentspace{\bm{x}}$.
This operator can be considered as the inverse mapping of the differential $D_{\bm{x}} \psi_{\bm{\theta}}:  \tangentspace{\bm{x}} \rightarrow \mathcal{T}_{\psi_{\bm{\theta}}(\bm{x})} \mathcal{N}$.
As we already have the diffeomorphism $\psi_{\bm{\theta}}$, the straightforward solution is to compute its derivatives and then obtain the required inverse.
Nevertheless, under the Riemannian setting, particularly for $d$-dimensional submanifolds $\manifold^d$ embedded in $\mathbb{R}^n$, computing the inverse directly becomes problematic due to the geometric constraints arising from $\manifold$. 
Next, we provide two methods to deal with this problem.

\paragraph{Pullback operator via constrained optimization:}
\label{subsubsec:pullbackConstrainedOpt}
Instead of naively differentiating through the ODE solver of $\psi_{\bm{\theta}}$, we can use the adjoint method to calculate the differential of a diffeomorphism constructed by a Neural MODE.
Assuming that we have the differential $D_{\bm{x}} \psi_{\bm{\theta}}$ (as computed in Algorithm~\ref{alg:differential_diffeomorphism} in App.~\ref{app:adjoint_method}), the connection between tangent vectors $\dot{\bm{x}}$ and $\dot{\bm{y}}$ can be written as $D_{\bm{x}} \psi_{\bm{\theta}}(\bm{x}) \dot{\bm{x}} = \dot{\bm{y}}$.
In the Euclidean case, we can directly compute $\dot{\bm{x}} = (D_{\bm{x}} \psi_{\bm{\theta}}(\bm{x}))^{-1}\dot{\bm{y}}$.
However, under the Riemannian setting, computing the inverse $(D_{\bm{x}} \psi_{\bm{\theta}}(\bm{x}))^{-1}$ often leads to a loss of rank in the matrix representation of $D_{\bm{x}} \psi_{\bm{\theta}}(\bm{x})$ for an embedded submanifold $\mathcal{M}^d$ due to the intrinsic geometric constraints of $\bm{x}$.
We address this problem by introducing geometric constraints that allow us to compute $\dot{\bm{x}}$ on $\tangentspace{\bm{x}}$.
For example, for manifold $\SphereManifold^d$, the tangent vector $\dot{\bm{x}}$ is orthogonal to $\bm{x}$, that is $\bm{x}^\trsp \dot{\bm{x}}=0$.
Hence, we can find a solution by solving a constrained optimization problem, from which the pullback operator $D_{\bm{y}} \psi_{\bm{\theta}}^\star$ is obtained as,
\begin{equation}
    D_{\bm{y}} \psi_{\bm{\theta}}^\star = \left[ D_{\bm{x}} \psi_{\bm{\theta}}(\bm{x})^\trsp D_{\bm{x}} \psi_{\bm{\theta}}(\bm{x}) + \bm{x} \bm{x}^\trsp \right]^{-1} D_{\bm{x}} \psi_{\bm{\theta}}(\bm{x})^\trsp .
    \label{eq:pullback_operator_sphere_contrained_problem}
\end{equation}
The full derivation and discussions can be found in App.~\ref{app:PullbackConstrainedOpt}.
Note that $D_{\bm{y}} \psi_{\bm{\theta}}^\star$ in~\eqref{eq:pullback_operator_sphere_contrained_problem} is specific to hypersphere manifolds due to the choice of constraints. 
Thus, this constrained optimization approach does not easily scale to compute the pullback operator for arbitrary Riemannian manifolds. 

\paragraph{Pullback operator via the adjoint method:}
\label{subsubsec:pullbackAdjoint}
To generalize the computation of the pullback operator for arbitrary Riemannian manifolds, we introduce a new approach based on a modified version of the adjoint method.
By reversing the integration time interval (i.e. from $[t_s, t_e]$ to $[t_e, t_s$]), we can determine the inverse diffeomorphism $\psi_{\bm{\theta}}^{-1}$, which is a distinct benefit of Neural MODEs.
Thus, the pullback operator $D_{\bm{y}} \psi_{\bm{\theta}}^\star$ can be viewed as the differential of the inverse diffeomorphism $D_{\bm{y}} (\psi_{\bm{\theta}}^{-1})$.
Furthermore, we leverage the adjoint method to compute the differential of $\psi_{\bm{\theta}}^{-1}$
using the adjoint ODE
$\dot{\bm{A}}^*(t) = - \bm{A}^*(t) D_{\bm{z}(t)}  f_{\bm{\theta}}(\bm{z}(t), t)$, with $\bm{A}^*(t):=D_{\bm{z}(t)} (\psi_{\bm{\theta}}^{-1})$.
Due to the availability of starting states $\bm{z}(t_s) = \bm{x}$ and $\bm{A}^*(t_s) = \bm{I}_n$, we can integrate both the Neural MODE~\eqref{eq:neuralode} and adjoint ODE to get the $D_{\bm{y}} (\psi_{\bm{\theta}}^{-1})$.
For clarification, we provide Algorithm~\ref{alg:differential_inv_diffeomorphism} in App.~\ref{app:PullbackAdjoint} for computing $D_{\bm{y}} (\psi_{\bm{\theta}}^{-1})$.
Although, we can use dynamic charts method to solve the Neural MODE, dealing with the adjoint ODE dynamics is still not straightforward. 
The main challenge is to compute the differential of the vector fields on the Riemannian manifold $D_{\bm{z}(t)}  f_{\bm{\theta}}(\bm{z}(t), t)$, despite it is nothing but partial derivatives in the Euclidean case.
To avoid directly computing the differential of vector fields on $\manifold$, we adopt an approach similar to~\eqref{eq:dynamic_chart_composition}, such that a component $\bm{z}_i = (\operatorname{Exp}_{\bm{z}_i} \circ \hat{\psi}_{\bm{\theta},i}^{-1} \circ \operatorname{Log}_{\bm{z}_i}) \bm{z}_{i+1}$ for computing the inverse diffeomorphism has its differential as, 
\begin{equation}
    D_{\bm{z}_{i+1}} \bm{z}_i = D_{\bm{w}_i(t_{i, s})} \operatorname{Exp}_{\bm{z}_i} \circ D_{\bm{w}_i(t_{i, e})} \hat{\psi}_{\bm{\theta},i}^{-1} \circ D_{\bm{z}_{i+1}} \operatorname{Log}_{\bm{z}_i} ,
    \label{eq:inv_block_eq}
\end{equation}
where $D_{\bm{w}_i(t_{i,e})} \hat{\psi}_{\bm{\theta},i}^{-1}$ boils down to partial derivatives (the proof of~\eqref{eq:inv_block_eq} is provided in App.~\ref{app:PullbackAdjoint}).


\section{Experiments}
\label{sec:result}
We evaluate our method in two settings: reproducing trajectories based on the LASA dataset~\citep{Lee18:RiemannManifold} projected on $\SphereManifold^2$; and reproducing real dynamic motions learned from demonstrations. 
To show the importance of incorporating geometry, we compare against a baseline method similar to Euclideanizing flows~\citep{Rana2020:EuclideanizingFlows}, that is implemented using CNFs with Neural ODEs in $\euclideanspace^n$ for illustrative experiments on a modified LASA dataset. We refer to this baseline as \textit{EuclideanFlow} and our model as \textit{RSDS}.

\begin{figure}[t!]
    \includegraphics[width =.24\linewidth]{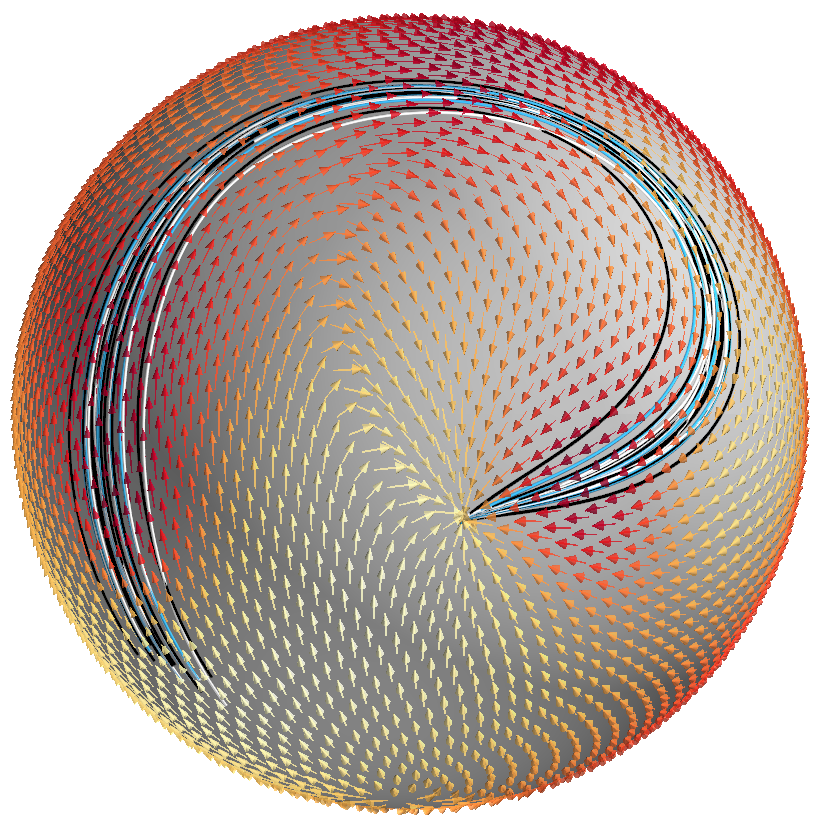}
    \includegraphics[width = .24\linewidth]{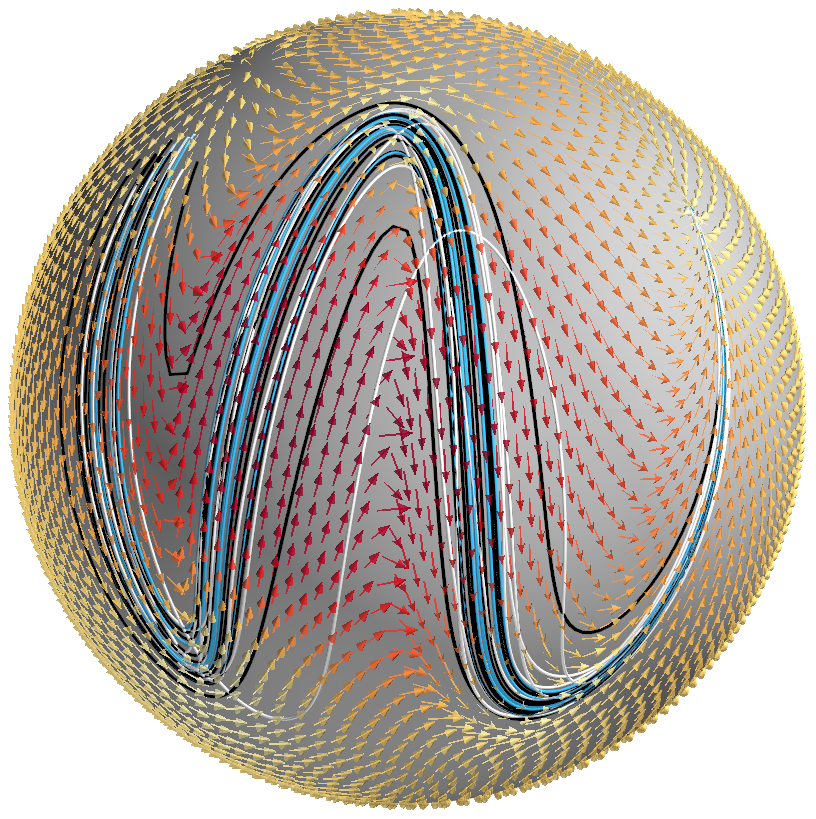}
    \includegraphics[width =.24\linewidth]{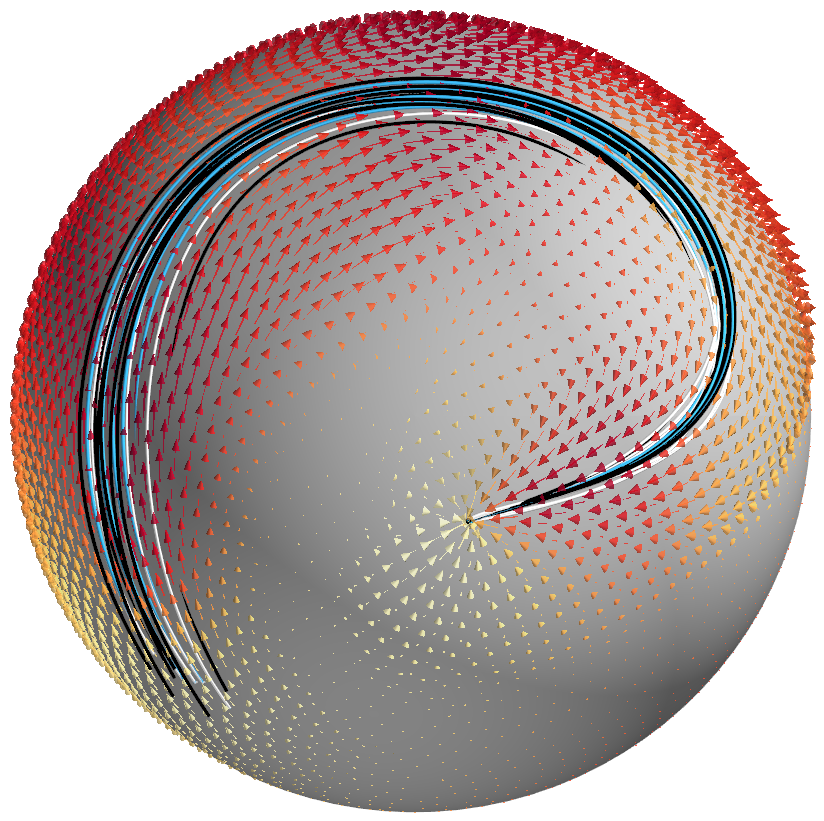}
    \includegraphics[width = .24\linewidth]{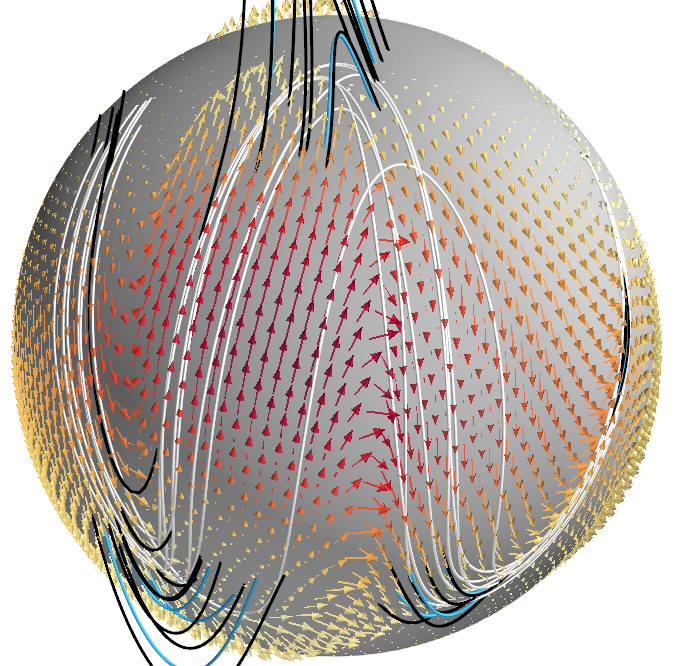}
    \caption{Experiments on LASA dataset on $\SphereManifold^2$ ($\mathsf{P}$ and $\mathsf{W}$ letters): Demonstrations (white), learned vector fields, and reproductions (black and blue). Blue trajectories start at the same initial points as the demonstrations, while the black ones depart from randomly-sampled points around the initial points of the demonstrations. The first two plots show the results for our RSDS approach, and the last two display the \textit{EuclideanFlow} results.}           
    \label{fig:lasa}
\end{figure}

\subsection{LASA dataset on $\SphereManifold^2$}
\label{subsec:ExpSphere}
\paragraph{Architectures:}
For \textit{EuclideanFlow}, we use a fully-connected neural network with an input vector on $\euclideanspace^{4}$ (i.e., the $3$-dimensional state $\bm{x}$ and time), and $3$ hidden layers each, with $32$ hidden units for $\SphereManifold^2$.
We use $\operatorname{tanh}$ as activation function to guarantee a $\mathcal{C}^1$-bounded mapping for modeling the Neural ODE. 
The RSDS architecture has an additional projection operator $\proju$ on the head of the network to impose the output on the tangent space of manifolds.
The scaling factor $\hat{k}_{\bm{\gamma}}$ is generated using a network composed of an RBF layer and a linear layer without bias. 
The network architectures are depicted in App.~\ref{app:NetworkArchiecture} (Fig.~\ref{fig:net_structures_framework}). 
For all the experiments, we use an Euler ODE solver with step size of $1/32$, and $4$ coordinate charts for the integration. 
For each dataset, one trajectory is used for testing and the remaining ones as training set. 
All models are trained using the ADAM optimizer~\citep{kingma2014:ADAM} with learning rate of $10^{-3}$ (decaying by factor $0.1$ after $1000$ epochs) over $2000$ epochs. 

\begin{figure}[t!]
    \includegraphics[width =.33\linewidth]{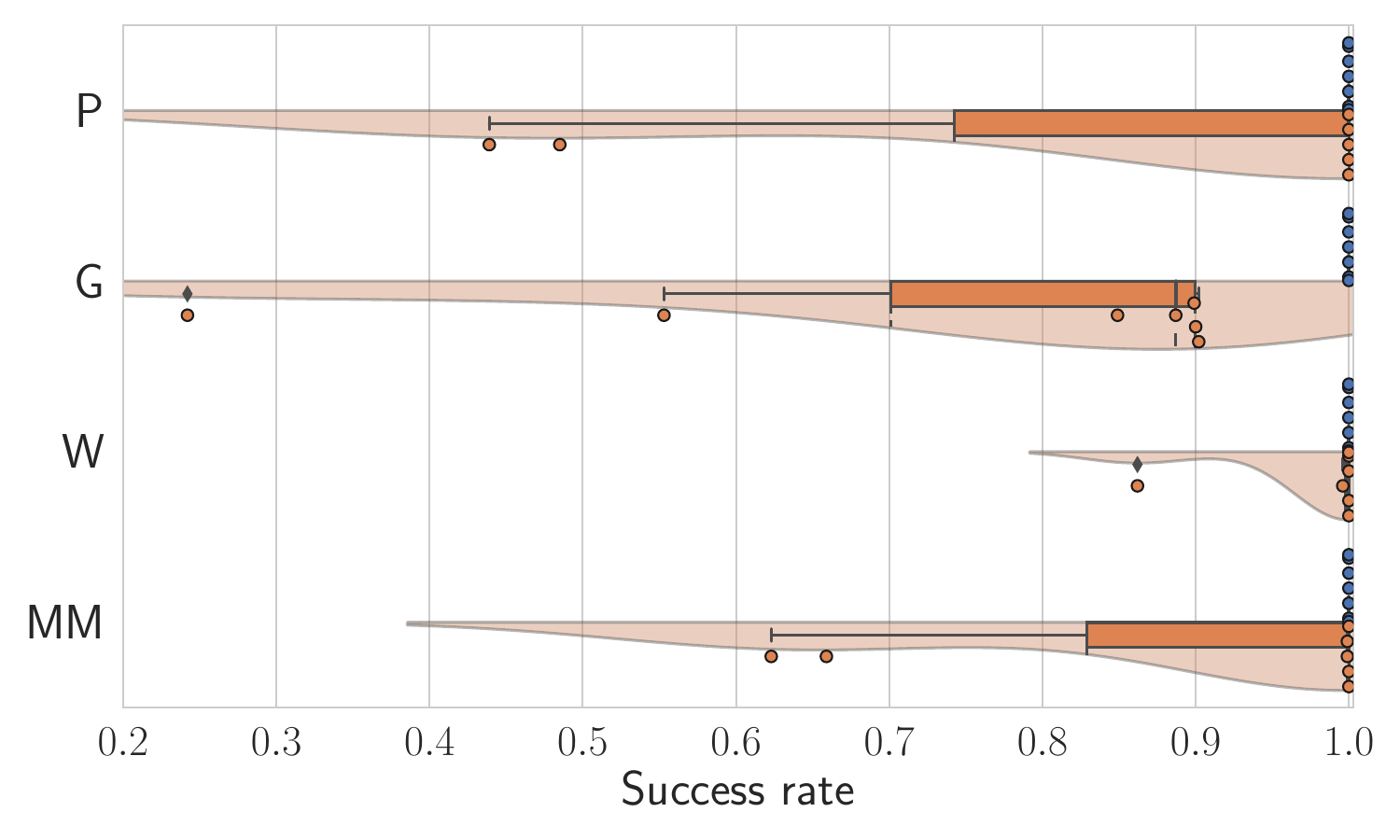}
    \includegraphics[width =.33\linewidth]{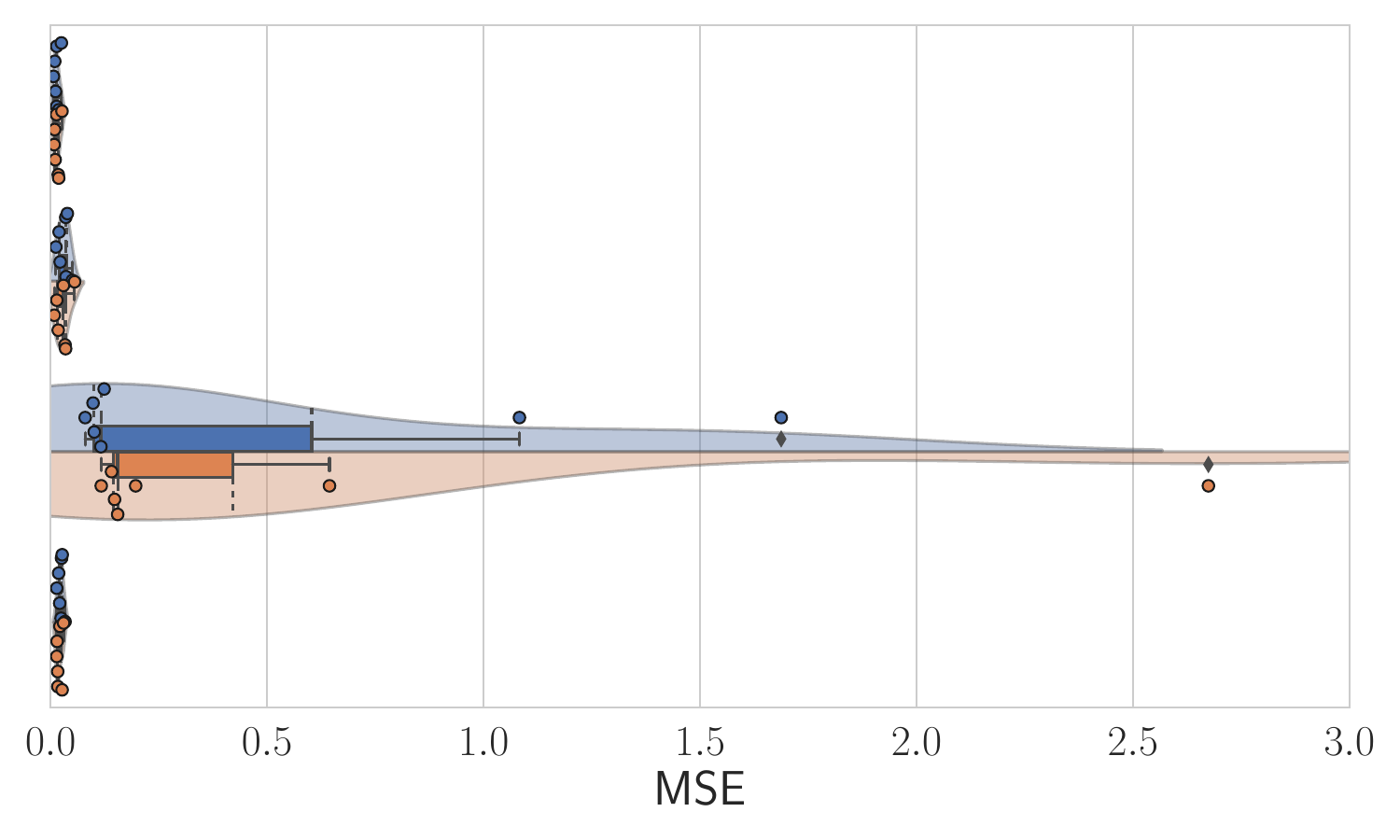}
    \includegraphics[width =.33\linewidth]{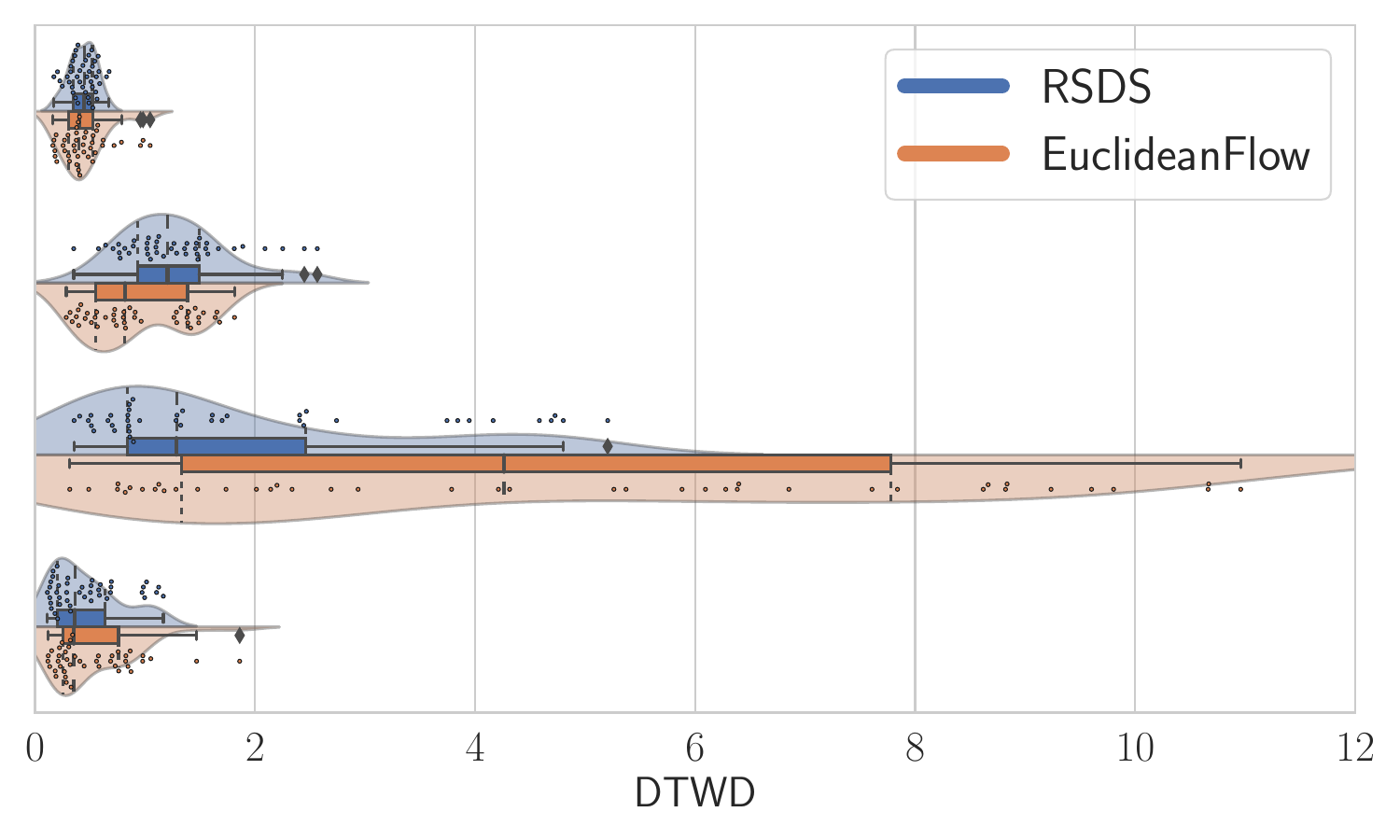}
    \caption{\textit{Left}: Average success rate of RSDS and \emph{EuclideanFlow} over randomly-sampled initial points on $\SphereManifold^2$ using $7$ different trained models indicated as points. \textit{Middle:} Average mean square error (MSE) between observed and predicted velocity over data points in the test trajectories indicated as points. \textit{Right:} Dynamic time warping distance (DTWD) between demonstrations and reproductions.} 
    \label{fig:metrics}
\end{figure}

To illustratively show how RSDS and \emph{EuclideanFlow} perform, we use a modified version of the LASA dataset of hand-written letters~\citep{Lemme2015:LasaDataset}, whose trajectories are projected on $\SphereManifold^2$.
The corresponding vector field can be easily computed from the projected trajectories by using the logarithmic map. 
As a result, we have a new dataset $\{\{\bm{x}_{m,t}, \Dot{\bm{x}}_{m,t}\}_{t=1}^{T_m}\}_{m=1}^M$ of $M$ demonstrations for each letter, with positions $\bm{x} \in \SphereManifold^2$ and velocities $\Dot{\bm{x}} \in \mathcal{T}_{\bm{x}}\SphereManifold^2$, from which we learn stable dynamical systems. 
Figure~\ref{fig:lasa} shows the resulting demonstrations as white curves for datasets $\mathsf{P}$ and $\mathsf{W}$, the corresponding learned vector fields and the reproduced trajectories.
Note that we provide more results in App.~\ref{app:extended_results} and comparisons to an alternative method to \textit{EuclideanFlow}, which projects trajectories onto the manifold after computing the integration in Euclidean space.
Concerning the reproductions, the blue and black trajectories are rollouts starting from the same initial position as the demonstrations and from randomly-sampled points around them, respectively.
Regarding our RSDS approach (first two plots in Fig.~\ref{fig:lasa}), it is evident that all blue and black rollouts closely match the demonstrations and converge to the equilibrium. 
In contrast, the \textit{EuclideanFlow} reproductions  constantly leave the manifold since there are no mechanisms accounting for the inherent geometric constraints of the data (see last two plots in Fig.~\ref{fig:lasa}). 
The bald regions on the manifold where the velocity vectors point inwards towards the sphere's center are also evidence of this phenomenon.
We also provide quantitative metrics for accuracy comparisons.
Figure~\ref{fig:metrics}-\emph{right}, and -\emph{middle} show the \textit{dynamic time warping distance} (DTWD) as a measure of reproduced position trajectory accuracy, and the \textit{mean squared error} (MSE) of the velocities reproduction. 
These metrics show that RSDS outperforms \textit{EuclideanFlow} for the most complex trajectories, e.g. the $\mathsf{W}$ dataset.
Although Fig.~\ref{fig:metrics} shows that both models seem to perform well on the $\mathsf{P}$, $\mathsf{G}$, and $\mathsf{Multi Models}$ datasets, as pointed out before, Fig.~\ref{fig:lasa} displays that the \textit{EuclideanFlow} trajectories do not obey the data geometry. 

\begin{figure}[t!]
    \includegraphics[width =.24\linewidth]{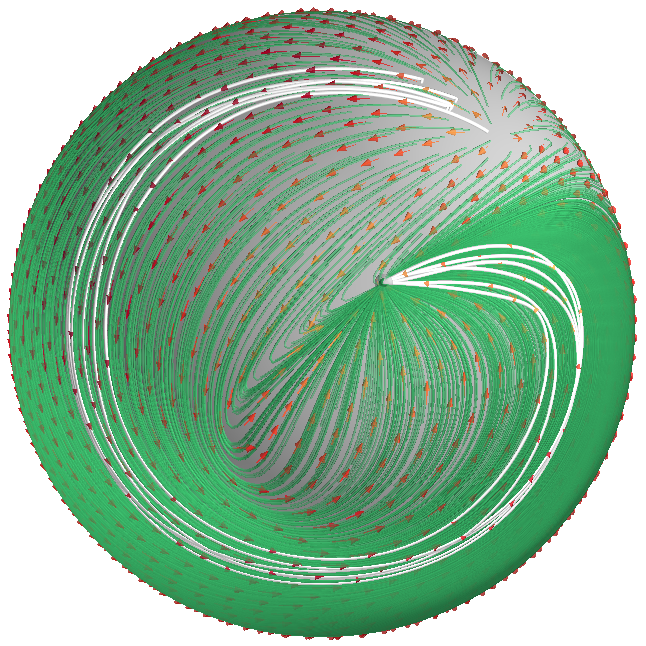}
    \includegraphics[width =.24\linewidth]{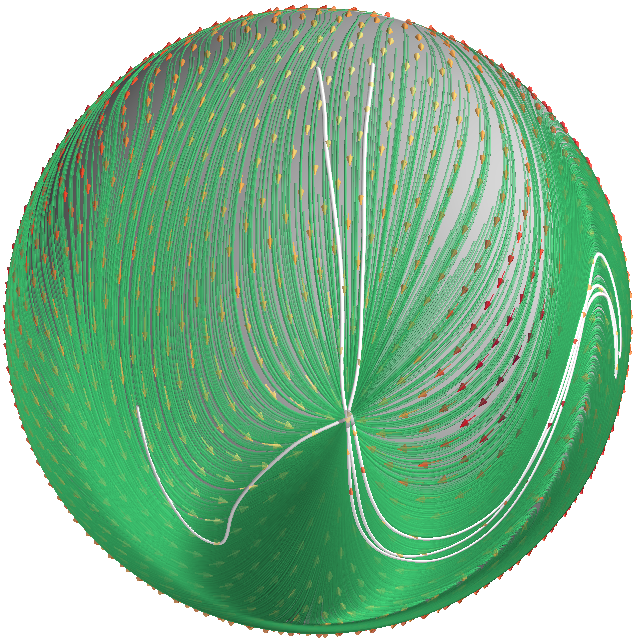}
    \includegraphics[width =.24\linewidth]{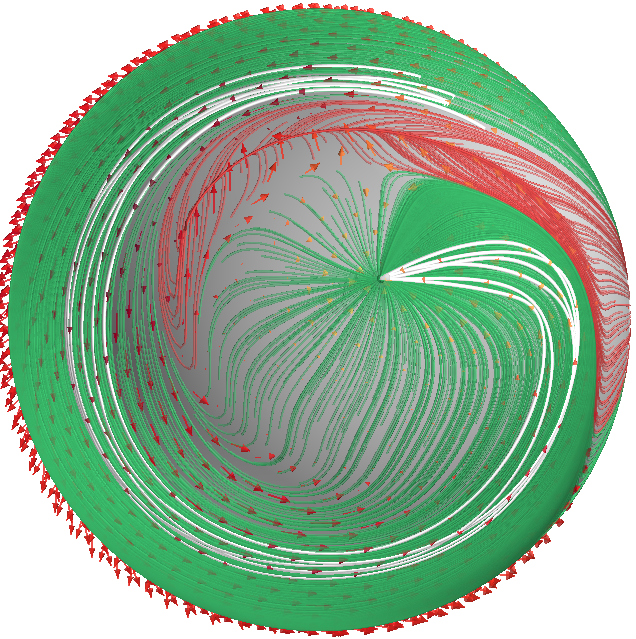}
    \includegraphics[width =.24\linewidth]{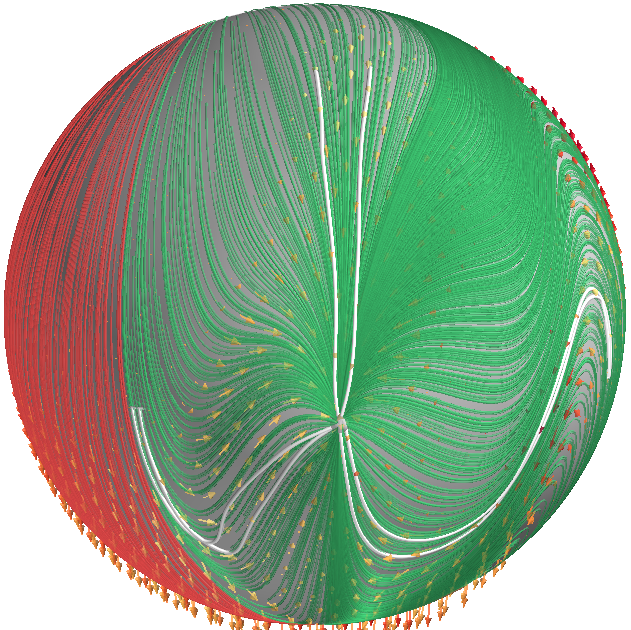}
    \caption{Stability of reproductions on $\SphereManifold^2$ for $\mathsf{G}$ and $\mathsf{Multi Models}$: $1000$ trajectories starting from uniformly-sampled points.
    The successful and failed trajectories are depicted in green and red, respectively.
    The first two spheres from the left correspond to RSDS reproductions while the other two relate to \textit{EuclideanFlow} results. For \textit{EuclideanFlow}, we first project the vector fields onto $\SphereManifold^2$ and then compute the integration trajectories.} 
    \label{fig:stability}
\end{figure}

Secondly, we evaluate the stability of both approaches. For a fair comparison, we first project the vector fields onto $\SphereManifold^2$ and compute the integration trajectories for \textit{EuclideanFlow}. 
To quantitatively assess this, we measure the stability of the learned vector fields (i.e. convergence to the equilibrium), by uniformly sampling $1000$ initial points on $\SphereManifold^2$ and counting the number of trajectories that successfully converge.
This procedure is repeated for $7$ different trained models, with the average success rate computed over the initial points. 
Using one of the trained models, Fig.~\ref{fig:stability} shows green and red curves representing successful and failed trajectories, respectively. 
It is evident that a large number of the \textit{EuclideanFlows} trajectories failed to converge despite the projection, however all the \textit{RSDS} trajectories succeeded.
This result is supported by the success rate metric displayed in Fig.~\ref{fig:metrics}--\emph{left}.
These results show that accounting for the data geometry, as in our RSDS approach, is crucial to provide stability guarantees of the learned dynamical system.
Additionally, we also compare the learning efficiency between the two methods, where RSDS generally requires fewer training epochs than \textit{EuclideanFlow}. 
Nevertheless, each training epoch of \textit{RSDS} is more computationally expensive due to manifold operations. 
The details of these results are provided in App.~\ref{app:extended_results} (Fig.~\ref{fig:loss_comparison_appendix}). 

\subsection{Real robot experiments on $\RtimeS$}
\begin{figure}[t!]
    \includegraphics[width =.245\linewidth]{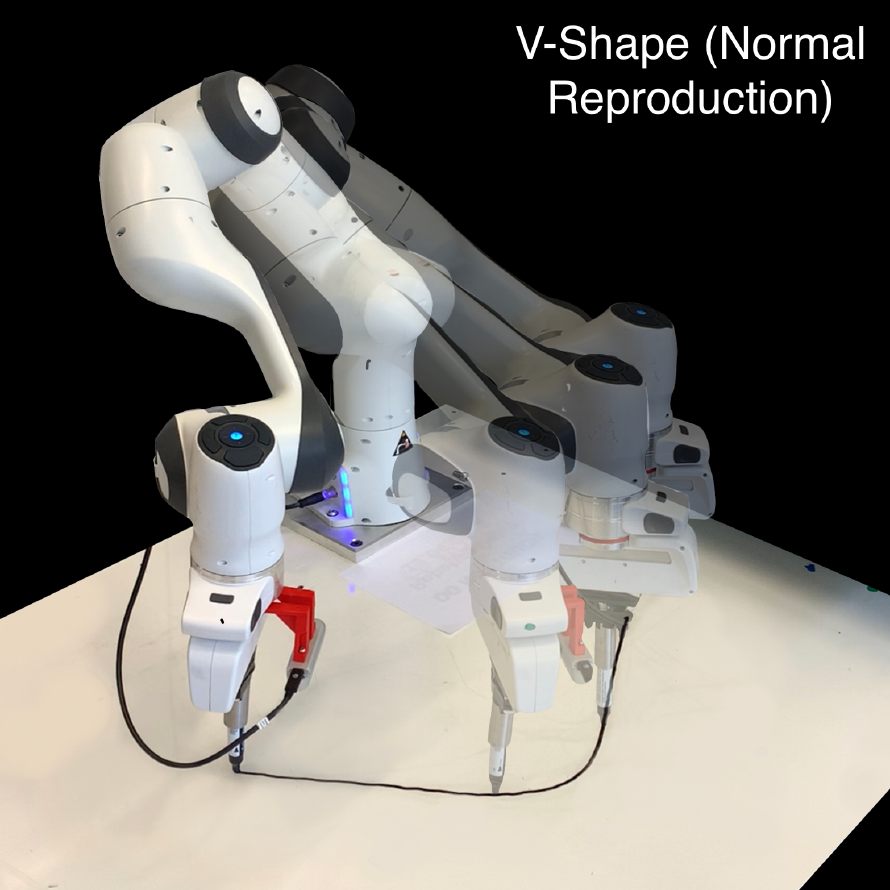}
    \includegraphics[width =.245\linewidth]{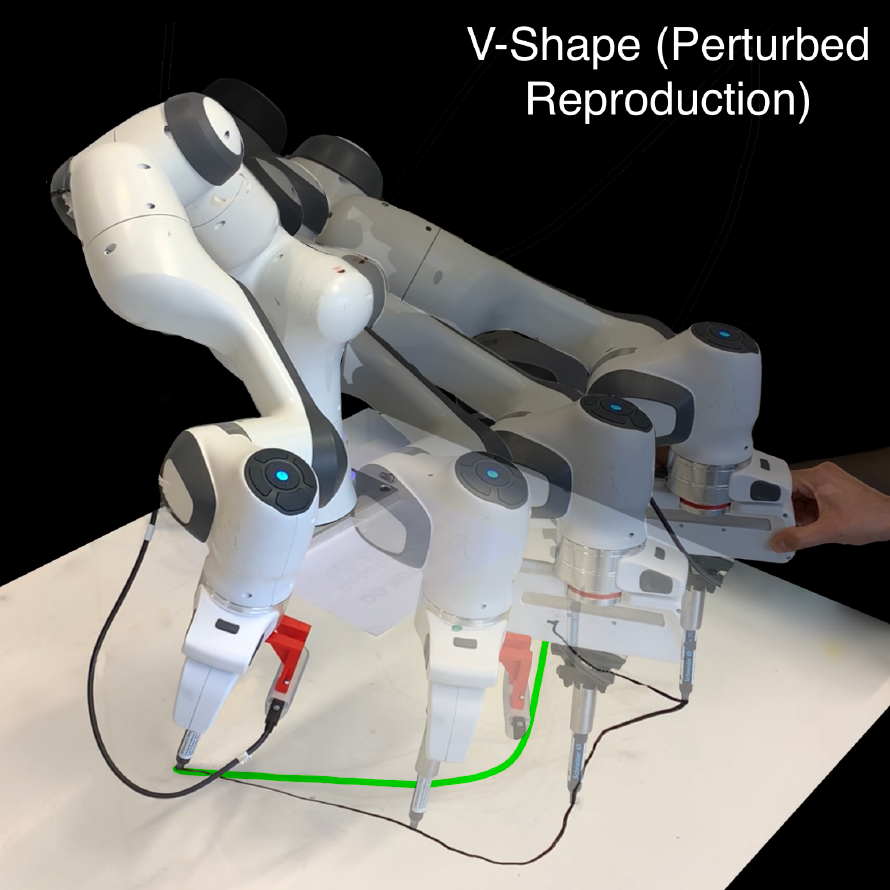}
    \includegraphics[width =.245\linewidth]{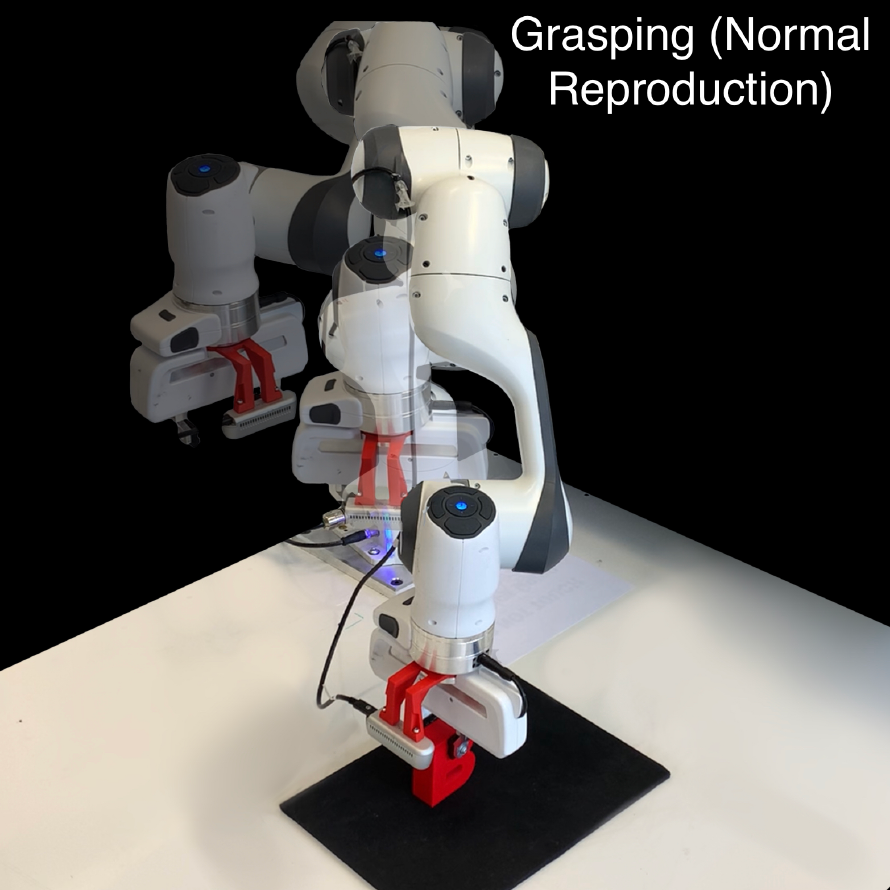}
    \includegraphics[width =.245\linewidth]{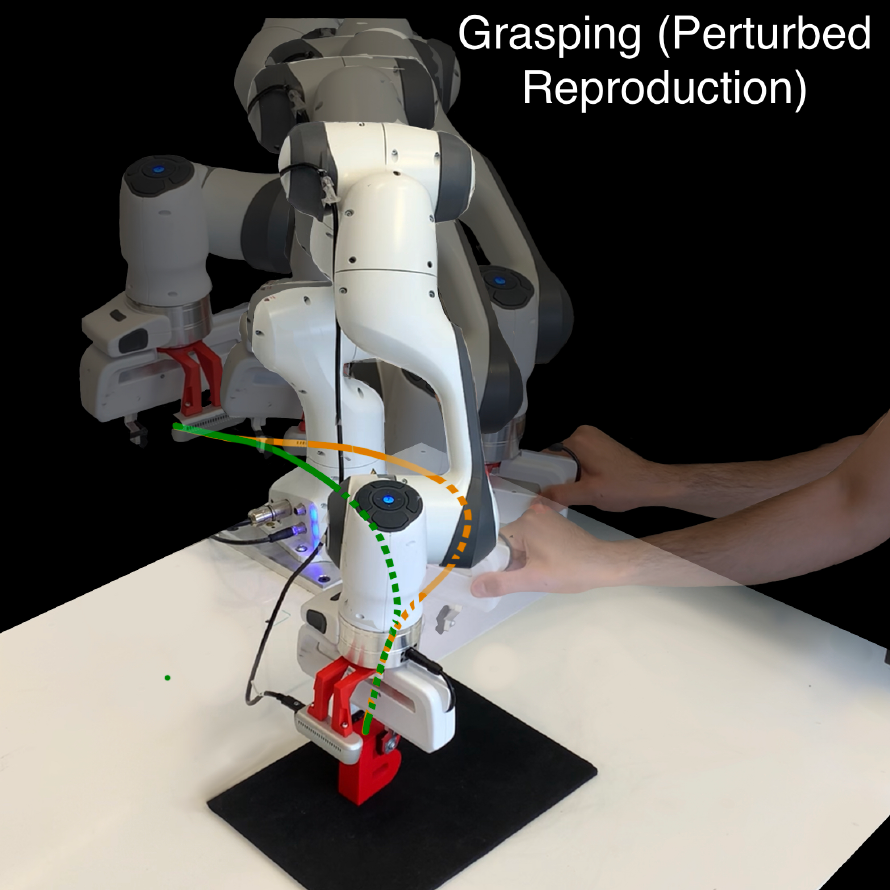}
    \caption{$\mathsf{DrawingTask}$ and $\mathsf{GraspingTask}$: Time evolution of the reproductions is depicted by superimposed images from different time frames. The transparent robots depict the trace of the motion trajectory. The second plot for each task displays the task reproduction under perturbation, where the unperturbed reproduction is depicted as a green trajectory for reference.} 
    \label{fig:robot_exp}
\end{figure}

We evaluated two different manipulation tasks using the $7$-DOF Franka-Emika Panda robot: \emph{(1)} a $\mathsf{GraspingTask}$ with $90$ degrees rotation, and \emph{(2)} a $\mathsf{V}$-shape $\mathsf{DrawingTask}$. 
For both experiments, we collected $10$ kinesthetic demonstrations of the robot end-effector motion as position-velocity trajectories at a frequency of \SI[group-digits=false]{10}{\hertz}.
Here, we show that RSDS can indeed be used in real-world applications. 
We train our model on the $\RtimeS$ manifold accounting for position and orientation of robot end-effector with the same network architecture as our illustrative experiments, except that we use $16$ hidden units for faster computations.
As shown in Fig.~\ref{fig:robot_exp}, while following the $\mathsf{V}$-shape curve in the $\mathsf{DrawingTask}$, the end-effector always faces the moving direction; similarly, when approaching the object in the $\mathsf{GraspingTask}$, the gripper rotates $90$ degrees.
These motion patterns require synchronized position and orientation trajectories, which is only attainable by training a model with a state variable on a product manifold, i.e., $\bm{x} \in \RtimeS$. 
To deploy the reproduced motion on the robot, we numerically integrated the desired velocity vector $\hat{\bm{x}} \in \tangentspace{\Dot{\bm{x}}}$ online, and used it as reference for a Cartesian impedance controller.

As observed in Fig.~\ref{fig:robot_exp}, the reproductions governed by the vector fields learned with our RSDS model accurately imitate the demonstrated motion patterns and converge to the goal position. 
Furthermore, these experiments incorporated some target shifts to test if our model could cope with them without retraining. 
We further evaluated the stability of the learned vector fields by perturbing the robot during the task reproduction. 
As Fig.~\ref{fig:robot_exp} shows, after perturbing the robot, the end-effector still follows an alternative trajectory computed from the learned vector field (see black and orange trajectories for $\mathsf{DrawingTask}$ and $\mathsf{GraspingTask}$, respectively).

\label{subsec:ExpReal}


\section{Discussion}
\label{sec:discussion}
We introduced a new approach RSDS to accurately learn vector fields on Riemannian manifolds while ensuring global asymptotic stability, which can not be achieved without taking into account the underlying geometry structure of the data. 
Our model inherits all the advantages of stable dynamical systems, such as high robustness against environmental perturbations.
To our knowledge, RSDS is the first to leverage neural ODEs on Riemannian manifolds to learn Lyapunov-stable Riemannian dynamical systems.
Moreover, RSDS builds on a new methodology to compute the pullback operator leveraging the characteristics of neural MODEs.
Our framework is generic and can theoretically be used to learn vector fields on any Riemannian manifolds with defined exponential and logarithmic maps.
As future work, we will leverage RSDS to learn vector fields on other Riemannian manifolds such as the manifold of symmetric-positive-definite (SPD) matrices $\mathcal{S}_{++}^d$, which is relevant in manipulability learning~\citep{Jaquier2021:ManipulabilityLearning} and video tracking~\citep{Cheng13:SPDdynamicSystem}.

\paragraph{Limitations:}
Due to the complexity of the Riemannian operators and the Neural MODE solvers, our framework runs relatively slowly, making it unsuitable for hard real-time applications.
This problem can be alleviated by switching to faster ODE solvers after training, which allows us to accelerate the query time at the expense of precision.
To improve the accuracy and stability of solving Neural MODEs, we can take advantage of techniques such as regularization~\citep{Finlay2020:JacobianRegularization} and recording checkpoint for the forward mode~\citep{Zhuang2020:ACA_gradientEstimation}.
In addition, since we leverage the Lyapunov stability to a single fixed point, the model may still reproduce some trajectories that are inconsistent with the trend of demonstration data due to lack of information for points far from demonstrations.
It may be worthwhile exploring other stability criteria, such as contraction analysis~\citep{Dawson2022:SafeCW}, to ensure the incremental exponential stability of trajectories with respect to each other on the manifold.


\clearpage
\acknowledgments{J. Zhang was supported by the Bosch Center for Artificial Intelligence (BCAI) as a master thesis student.} 


\bibliography{References}  

\clearpage
\appendix
\section{Extended background}
\subsection{Riemannian manifolds}
\label{app:RiemannianManif}
\begin{wrapfigure}[11]{r}{0.56\linewidth}
    \centering
    \includegraphics[clip, trim={40mm, 40mm, 40mm, 55mm}, width=0.54\textwidth]{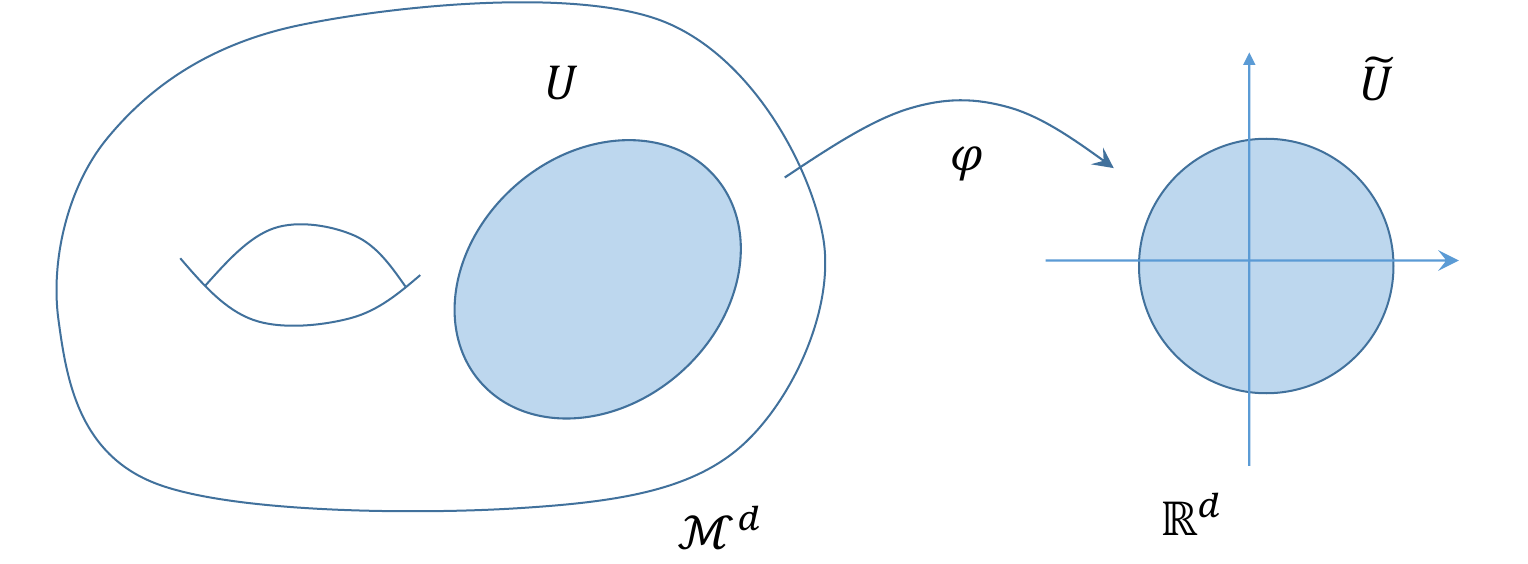}
    \caption{Coordinate chart $\varphi$ applied on a smooth manifold $\manifold^d$ from an open set $U$ to $\tilde{U}$.}
    \label{fig:coordinate_chart}
\end{wrapfigure}
A key definition in the development of our approach is the notion of a \emph{chart.}
Formally speaking, a chart on a smooth manifold $\manifold$ is a diffeomorphic mapping $\varphi: U \to \tilde{U}$ from an open set $U \subset \manifold$ to an open set $\tilde{U} \subseteq \euclideanspace^d$, see Fig.~\ref{fig:coordinate_chart} for an illustration.

As briefly introduced in~\S~\ref{sec:background}, to operate with Riemannian manifolds, it is common practice to exploit the Euclidean tangent spaces. 
We resort to mappings back and forth between $\tangentspace{\bm{x}}$ and $\manifold$ using the exponential and logarithmic maps, namely, $\expmap{\bm{x}}{\bm{u}}: \tangentspace{\bm{x}} \to \manifold$ and $\logmap{\bm{x}}{\bm{y}}: \manifold \to \tangentspace{\bm{x}}$.
Another relevant operation is the parallel transport $\prltrsp{\bm{x}}{\bm{y}}{\bm{u}}: \tangentspace{\bm{x}}\to\tangentspace{\bm{y}}$, which moves manifold elements lying on different tangent spaces along geodesics on $\manifold$.
Finally, as described in \S~\ref{app:finalFramework}, we used the projection operator to project Euclidean vectors to the tangent space of the manifold. 
Formally, this orthogonal projection is defined as $\proju_{\bm{x}}(\bm{v}): \euclideanspace^n \to \tangentspace{\bm{x}}$, which is used to compute the Riemannian gradient as the orthogonal projection of the ``Euclidean'' gradient to the manifold tangent space.
The specific operations for the unit hypersphere, used in this work, as provided in Table~\ref{tab:SphereOperations}.

\setlength{\extrarowheight}{2pt}
\begin{table}[h]
	\centering
	\begin{tabularx}{\linewidth}{|l|X|}
		\hline
		\textbf{Operator} & \textbf{Formula} \\ [0.3ex]
		\hline 
		$\manifolddist{\bm{x}}{\bm{y}}$ & $\arccos(\bm{x}^\trsp \bm{y})$ \\ [0.3ex] 
		\hline
		$\expmap{\bm{x}}{\bm{u}}$ & $\bm{x}\cos(\|\bm{u}\|) + \bm{\overline{u}}\sin(\|\bm{u}\|)$ with $\bm{\overline{u}}=\frac{\bm{u}}{\|\bm{u}\|}$\\ [0.3ex]
		\hline 
		$\logmap{\bm{x}}{\bm{y}}$ & $ d_{\mathcal{M}}(\bm{x},\bm{y}) \, \frac{\bm{y} - \bm{x}^\trsp \bm{y} \, \bm{x}}{\|\bm{y} - \bm{x}^\trsp \bm{y} \, \bm{x}\|}$\\ [0.4ex]
		\hline 
		$\prltrsp{\bm{x}}{\bm{y}}{\bm{u}}$ & $\left( -\bm{x} \sin(\|\bm{v} \|) \bm{\overline{v}}^\trsp + \bm{\overline{v}} \cos(\|\bm{v}\|) \bm{\overline{v}}^\trsp) + (\bm{I}_n - \bm{\overline{v}} \bm{\overline{v}}^\trsp) \right) \bm{u}$,  $\bm{\overline{v}}=\frac{\bm{v}}{\|\bm{v}\|}$, $\bm{v} = \logmap{\bm{x}}{\bm{y}}$\\  [0.4ex]
		\hline 
		$\proju_{\bm{x}}(\bm{v})$ & $(\bm{I}_n - \bm{x} \bm{x}^\trsp) \bm{v}$\\ [0.4ex]
		\hline 
	\end{tabularx}
	\caption{Principal operations on $\SphereManifold^d$. For details, see~\citep{Boumal2022:IntoManifoldOpt}}
	\label{tab:SphereOperations}
\end{table}

\subsection{Neural MODEs for Diffeomorphism Learning}
\label{app:NeuralMODEs}
A diffeomorphism can be constructed by solving an initial value problem (or integral) for Neural ODEs~\citep{chen2018:NeuralODE}.
Futhermore, we can extend this idea to the manifold setting following~\citep{lou2020:NeuralMODE, mathieu2020:RCNormalizingFlows}.
To do so, we start from the following theorem from~\citet{mathieu2020:RCNormalizingFlows}:

\begin{theorem}[Vector flows]
\label{th:vector_flows}
Let $\mathcal{N}$ be a smooth complete manifold and $f_{\bm{\theta}}$ be a
$\mathcal{C}^1$-bounded time-dependent vector field. 
Then there exists a global flow $\psi_{\bm{\theta}}: \mathcal{N} \times \mathbb{R} \rightarrow \mathcal{N}$ such that
for each $t \in \mathbb{R}$, the map $\psi_{\bm{\theta}}(\cdot, t): \mathcal{N} \rightarrow \mathcal{N}$ is a $\mathcal{C}^1$-diffeomorphism (i.e. $\mathcal{C}^1$ bijection with $\mathcal{C}^1$ inverse).
\end{theorem}
Accordingly, we can compute the diffeomorphism on the manifold $\mathcal{N}$ by solving an IVP of Neural MODEs~\eqref{eq:neuralode}, where the integration is done over a manifold instead of an Euclidean space.
During computation of the IVP, we can select the time interval $[t_s, t_e]$ as $[0, 1]$ without loss of generality. 
As a result, the diffeomorphism $\psi_{\bm{\theta}}$ and inverse diffeomorphism $\psi_{\bm{\theta}}^{-1}$ can be computed by integration forwards and backwards, respectively, as follows,
\begin{equation}
        \bm{y} = \psi_{\bm{\theta}}(\bm{x}) = \bm{x} + \int_{0}^{1} f_{\bm{\theta}}(\bm{z}(\tau), \tau) d\tau , \quad
        \bm{x} = \psi_{\bm{\theta}}^{-1}(\bm{y}) = \bm{y} + \int_{1}^{0} f_{\bm{\theta}}(\bm{z}(\tau), \tau) d\tau ,
    \label{eq:diffeomorphism_and_inv_diffeormorphism}
\end{equation}
where $\bm{x}=\bm{z}(t_s)$ and $\bm{y}=\bm{z}(t_e)$ are two points on the manifold $\mathcal{N}$.
Note that in this section we view the diffeomorphism $\psi_{\bm{\theta}}$ as a function on the Riemannian manifold $(\mathcal{N}, h)$.
However, under this diffeomorphic mapping, we actually create a new Riemannian manifold $(\mathcal{M}, \hat{h})$ with $\hat{h}$ being a pullback metric $\hat{h}_{\bm{x}}(\bm{u}, \bm{v}) = h_{\psi_{\bm{\theta}}(\bm{x})}(D_{\bm{x}}\psi_{\bm{\theta}} \  \bm{u}, D_{\bm{x}}\psi_{\bm{\theta}} \  \bm{v})$.
From this perspective, the diffeomorphism $\psi_{\bm{\theta}}$ can be regarded as a mapping from a Riemannian manifold $(\mathcal{M}, \hat{h})$ to $(\mathcal{N}, h)$, that is $\psi_{\bm{\theta}}: \mathcal{M} \rightarrow \mathcal{N}$, with $(\mathcal{M}, \hat{h})$ and $(\mathcal{N}, h)$ being isometric Riemannian manifolds.

\subsection{Integrators on manifolds}
\label{app:integrators_manifolds}
Finding a numerical solution for integrators on Riemannian manifolds is not straightforward and needs some additional considerations. 
The majority of existing ODE solvers are designed and carefully optimized for Euclidean spaces (e.g. Euler, Runge-Kutta, and adaptive solvers), which can be modified to compute the integral on Riemannian manifolds.
In the following, we discuss two approaches for integration on Riemannian manifolds: \emph{projection methods} and \emph{integrators based on local coordinates}~\citep{hairer2011:ODEManifolds}.

\begin{wrapfigure}[9]{r}{0.5\linewidth}
    \centering
    \includegraphics[clip, trim={120mm, 80mm, 100mm, 75mm}, width=0.48\textwidth]{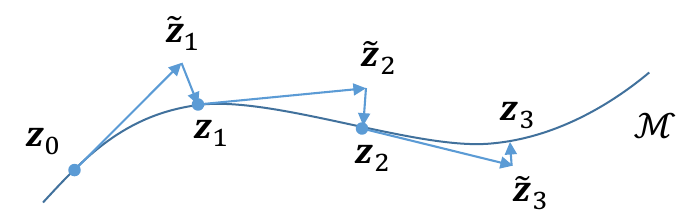}
    \caption{Standard projection method}
    \label{fig:standard_projection_method}
\end{wrapfigure}
\paragraph{Projection methods:}
Assume we have a Neural MODE given by~\eqref{eq:neuralode} on a $\mathcal{M}$, where $f_{\bm{\theta}}(\cdot, t) \in \mathcal{T}_{\bm{z}(t)}\mathcal{M}$ for all $\bm{z}(t) \in \mathcal{M}$. 
The standard projection methods compute a one-step integral with an arbitrary numerical integrator in Euclidean space and then projects the value onto the manifold $\mathcal{M}$, as depicted in Fig.~\ref{fig:standard_projection_method}.
For example, an Euler solver is selected with step size $dt$ which computes a single step from the initial point $\bm{z}_0$. 
The next state $\bm{\tilde{z}}_1$ is then computed as $\bm{\tilde{z}}_1 = \bm{z}_0 + f_{\bm{\theta}}(\bm{z}_0, 0) dt$.
Finally, $\bm{\tilde{z}}_1$ is projected back onto the manifold by leveraging the exponential map. 

\paragraph{Integrators on local coordinates:}
These methods consider a local representation of the manifold $\mathcal{M}$, which defines a coordinate map $\varphi: \mathcal{M}^d \supseteq U \rightarrow \tilde{U} \subseteq \mathbb{R}^d$, as described in \S~\ref{sec:background}, with coordinates $\bm{w}(t) = \varphi(\bm{z}(t))$. 
Since this coordinate map is a diffeomorphism, we can compute its inverse $\varphi^{-1}$. 
The differential equation for local coordinates $\bm{w}(t)$ is then given by~\eqref{eq:equivalentode} in Euclidean space.
This equation states that the differential of the coordinate map $D_{\varphi^{-1}(\bm{w}(t))}\varphi$ must be computed to project the vector fields on the manifold onto its coordinate codomain $\tilde{U}$.
In this way, we can solve the equivalent ODE in~\eqref{eq:equivalentode} with classical ODE solvers in Euclidean space, as follows,
\begin{equation}
    \bm{w}(t_e) = \bm{w}(t_s) + \int_{t_s}^{t_e} D_{\varphi^{-1}(\bm{w}(\tau))}\varphi \circ f_{\bm{\theta}}(\varphi^{-1}(\bm{w}(\tau)), \tau) d\tau .
\end{equation}
As a result, the solution of the manifold ODE~\eqref{eq:neuralode} and the equivalent ODE in~\eqref{eq:equivalentode} are connected via coordinate maps $\varphi$, so that any approximation $\bm{w}(t_n)$ provides an approximation for $\bm{z}(t_n)$.
Theoretically there are infinite possible choices of local coordinates which can be selected before or during the integration process.
However, for complex Riemmanian manifolds that are not globally diffeomorphic to the Euclidean space, it is impossible to find a single coordinate chart to cover the whole manifold.
Therefore, different coordinate charts must be selected during the integration process.
Combining the equivalent ODE and multiple choices of coordinates charts leads to the so-called approach \emph{Dynamic Chart Method}~\citep{lou2020:NeuralMODE}.
We take advantage of this method to efficiently perform integration on manifolds. 

\subsection{Adjoint Method}
\label{app:adjoint_method}
To incorporate Neural MODEs into deep learning frameworks, it is crucial to compute their gradients efficiently.
According to~\citet{chen2018:NeuralODE}, the adjoint sensitivity method allows us to treat the ODE solvers as a black box and compute the corresponding gradients by solving a second adjoint ODE backwards in time, instead of directly differentiating through ODE solvers (which is the na\"ive solution).
This approach is significantly more memory efficient, especially for adaptive ODE solvers, which adjust the step size on the fly. 
Furthermore, the adjoint method can be generalized for ODEs on arbitrary manifolds~\citet{lou2020:NeuralMODE}.

Formally, let us consider a loss function $\mathcal{L} : \mathcal{M}^d \rightarrow \mathbb{R}$.
To compute gradients of $\mathcal{L}$ with respect to any value of the state variable $\bm{z}(t)$ of the manifold ODE, we can define an \emph{adjoint variable} $\bm{a}(t)^\trsp := D_{\bm{z}(t)} \mathcal{L} $, subject to,
\begin{equation}
     \dot{\bm{a}}(t)^\trsp = - \bm{a}(t)^\trsp D_{\bm{z}(t)}  f_{\bm{\theta}}(\bm{z}(t), t) .
     \label{eq:adjointode}
\end{equation}
If the loss $\mathcal{L}$ merely depends on the end value $\bm{z}(t_e)$, to get the gradient of $\mathcal{L}$ w.r.t. the starting value $\bm{z}(t_s)$, we solve the end value problem (EVP) of the adjoint ODE~\eqref{eq:adjointode}.
Since this adjoint ODE contains two variables: $\bm{z}(t)$ and $\bm{a}(t)^\trsp$, solving the EVP of it requires knowledge about $\bm{z}(t)$ over the entire time span, which leads to solving a combination of the adjoint ODE~\eqref{eq:adjointode} and the Neural MODE~\eqref{eq:neuralode} backwards.
Furthermore, the end values are needed for both variables $\bm{z}$ and $\bm{a}$ to solve ODEs backwards from end time $t_e$ to starting time $t_s$.
The end value $\bm{z}(t_e)$ needs to be computed first through the forward integration of the Neural MODE, while $\bm{a}(t_e)^\trsp = D_{\bm{z}(t_e)} \mathcal{L}$ can be obtained via classical backpropagation, since the loss $\mathcal{L}$ is only a function of $\bm{z}(t_e)$. 

Finally, to compute the differential of the diffeomorphism $\psi_{\bm{\theta}}$ with respect to the starting value $\bm{z}(t_s)$ (which will help us compute the pullback operator in \S~\ref{subsec:DiffInvDiffeomorph}), we consider a special loss $\mathcal{L}_i = q_i(\bm{z}(t_e))$, where $q_i$ is the $i$-th component of a function $q: \mathcal{N}^d \rightarrow \mathbb{R}^d$.
We introduce the function $q$ to help us construct a valid loss $\mathcal{L}_i = q_i \circ \psi_{\bm{\theta}}$, which maps from $\mathcal{M}^d$ to $\mathbb{R}$.
Furthermore, we define $\bm{l} = (\mathcal{L}_1, \ldots, \mathcal{L}_d)^\trsp = q(\bm{z}(t_e)) = (q \circ \psi_{\bm{\theta}})(\bm{z}(t_s))$. 
Given the previous definition, we can define $\tilde{\bm{A}}(t) = D_{\bm{z}(t)} \bm{l}$ as an adjoint variable as follows,
\begin{equation}
    \tilde{\bm{A}}(t) = D_{\bm{z}(t)} \bm{l} = D_{\bm{z}(t)} (q \circ \psi_{\bm{\theta}}) = D_{\bm{z}(t_e)} q \circ  D_{\bm{z}(t)} \psi_{\bm{\theta}} = D_{\bm{z}(t_e)} q \circ \bm{A}(t) ,
\end{equation}
where $\bm{A}(t)$ is the derivative of the diffeomorphism with respect to $\bm{z}(t)$. Consequently, we build the corresponding adjoint ODE,
\begin{equation}
    \begin{split}
        \dot{\tilde{\bm{A}}}(t) &= - \tilde{\bm{A}}(t) \circ D_{\bm{z}(t)}  f_{\bm{\theta}}(\bm{z}(t), t) , \\
        D_{\bm{z}(t_e)} q \circ \dot{ \bm{A}}(t) &= - D_{\bm{z}(t_e)} q \circ \bm{A}(t) \circ D_{\bm{z}(t)}  f_{\bm{\theta}}(\bm{z}(t), t), \\
        \dot{\bm{A}}(t) &= - \bm{A}(t) \circ D_{\bm{z}(t)}  f_{\bm{\theta}}(\bm{z}(t), t) .
    \end{split}
\end{equation}
Moreover, we have initial and final conditions defined as $\bm{A}(t_s) = D_{\bm{z}(t_s)} \psi_{\bm{\theta}}$ and $\bm{A}(t_e) = D_{\bm{z}(t_e)} \bm{z}(t_e) = \bm{I}$.
As a result, to compute $\bm{A}(t_s)$, we can construct an ``augmented'' ODE composed of the Neural MODE~\eqref{eq:neuralode} and the adjoint ODE $ \dot{ \bm{A}}(t) = - \bm{A}(t) D_{\bm{z}(t)}  f_{\bm{\theta}}(\bm{z}(t), t)$.
After computing the IVP of the Neural MODE to obtain $\bm{z}(t_e)$, we can then compute the differential of the diffeomorphism $D_{\bm{x}} \psi_{\bm{\theta}}$ by solving the EVP of the augmented ODE, as shown in Algorithm~\ref{alg:differential_diffeomorphism}.

{\SetAlgoNoLine%
\begin{algorithm}
\caption{Differential of diffeomorphism constructed by a Neural MODE}
\label{alg:differential_diffeomorphism}
\SetKw{Return}{return}
\SetKwProg{Def}{def}{:
}{end}
\SetKwInOut{Input}{Input}
\Input{start time $t_s$, start end $t_e$, final states $\bm{z}(t_e)$ and $\bm{A}(t_e) = \bm{I}$}
\Indp $\bm{s}_0 = [\bm{z}(t_e), \bm{I}]$ \\
$\%$ augmented ODE using~\eqref{eq:neuralode} and adjoint ODE with variable $\bm{A}$ \\
\Def{augmented{\_}dynamics$([\bm{z}(t), \bm{A}(t)], t)$}{\Return{$\left[f_{\bm{\theta}}(\bm{z}(t), t), -\bm{A}(t) D_{\bm{z}(t)}  f_{\bm{\theta}}(\bm{z}(t), t)\right]$} }   
$\left[\bm{z}(t_s), D_{\bm{z}(t_s)} \bm{l}\right]$ = ManifoldODESolve$(\bm{s}_0, \text{augmented\_dynamics}, t_e, t_s)$ \\
\Indm \Return{$\bm{z}(t_s)$,  $D_{\bm{z}(t_s)} \bm{l}$}
\end{algorithm}
}
To emphasize once more, solving this augmented ODE system backwards requires knowing the end value $\bm{z}(t_e)$, which can only be obtained by solving the first ODE (Neural MODE) forwards.
In other words, this process is essentially similar to the classical neural network backpropagation in the sense that we must first perform forward pass through the network and then back-propagate the error to obtain the gradients.

\section{Stability Analysis of RSDS}
\label{app:stabilityAnalysis}
\subsection{Lyapunov stability on Riemannian Manifolds}
Before we prove the Lyapunov stability of RSDS~\eqref{eq:diffeomorphism_based_svf_equation}, we first introduce the following theorem
\begin{theorem}[Stability of geodesic vector fields]
\label{th:stability_geodesics_vf}
Let $\mathcal{N}$ be a Riemannian manifold with logarithmic map $\operatorname{Log}_{\bm{y}}: \mathcal{N} \rightarrow \mathcal{T}{\bm{y}} \mathcal{N}$ at $\bm{y} \in \mathcal{N}$. 
A dynamical system on this manifold can be formulated as,  
\begin{equation}
    \dot{\bm{y}} = k(\bm{y}) \operatorname{Log}_{\bm{y}}(\bm{y}^*) ,
    \label{eq:geodesics_vf}
\end{equation}
where $k(\bm{y}) > 0 \ \forall \bm{y} \neq \bm{y}^*$ and $\bm{y}^*$ is the origin (equilibrium) point on $\mathcal{N}$. 
The velocity $\dot{\bm{y}}$ of such a dynamical system points along the geodesic curves converging to $\bm{y}^*$.
Moreover, let a Lyapunov function on such a manifold be formulated as $V(\bm{y}):=\langle  F , F \rangle_{\bm{y}^*}$, where $F = \operatorname{Log}_{\bm{y}^*}(\bm{y})$ defines a vector field composed of the initial velocities of all geodesics departing from the origin $\bm{y}^*$.
This Lyapunov function $V$ satisfies, 
\begin{equation*}
        \quad V(\bm{y}^*)=0, \quad  \dot{V}(\bm{y}^*) = 0 , \quad  V(\bm{y})>0, \  \forall \ \bm{y} \neq \bm{y}^* , \quad  \dot{V}(\bm{y}) < 0, \  \forall \ \bm{y} \neq \bm{y}^*  .
\end{equation*}
As a result, the above dynamical system is globally asymptotically stable on the Riemannian manifold $\mathcal{N}$ with a single equilibrium point $\bm{y}^*$.
\end{theorem}
The full proof of Theorem~\ref{th:stability_geodesics_vf} can be found in~\citep{Jaquier2021:ManipulabilityLearning}. 
Note that the dynamical system defined by $\dot{\bm{y}} = g_{\bm{\gamma}}(\bm{y}) = k_{\bm{\gamma}}(\bm{y}) g_n(\bm{y})$ as in \S~\ref{subsec:GeodesicVF}, corresponds to the one defined in Theorem~\ref{th:stability_geodesics_vf} by substituting $k(\bm{y})$ with $ \frac{k_{\bm{\gamma}}(\bm{y})}{\lVert \operatorname{Log}_{\bm{y}}(\bm{y}^*) \rVert_2}$.
Therefore, it is easy to observe that our canonical vector field on $\mathcal{N}$ is also globally asymptotically stable at the attractor point $\bm{y}^*$.
To analyze the stability properties of the dynamical system~\eqref{eq:diffeomorphism_based_svf_equation} on the manifold $\mathcal{M}$, we can construct a new Lyapunov function $\tilde{V}(\bm{x}) := V(\psi_{\bm{\theta}}(\bm{x}))$ by applying the diffeomorphism $\psi_{\bm{\theta}}$.
Since this diffeomorphism $\psi_{\bm{\theta}}$ is a one-to-one and onto mapping, there exists also a single equilibrium point on $\mathcal{M}$, i.e $\bm{x}^* = \psi_{\bm{\theta}}^{-1}(\bm{y}^*)$.
Then, we can prove that $\bm{x}^* \in \mathcal{M}$ is globally asymptotically stable.

To verify this condition, we first prove that the dynamics under the coordinate change with diffeomorphism $\psi_{\bm{\theta}}$ indeed corresponds to~\eqref{eq:diffeomorphism_based_svf_equation}.
Given $\bm{y} = \psi_{\bm{\theta}}(\bm{x}) \ \mathrm{and} \  \dot{\bm{y}} = g_{\bm{\gamma}}(\bm{y})$, it follows that,
\begin{equation*}
    \dot{\bm{y}} = D_{\bm{x}} \psi_{\bm{\theta}}(\bm{x}) \circ \dot{\bm{x}}, \quad \text{then,} \quad
    \dot{\bm{x}} = (D_{\bm{x}} \psi_{\bm{\theta}}(\bm{x}))^{-1} \dot{\bm{y}} = \left( (D_{\bm{x}} \psi_{\bm{\theta}}(\bm{x}))^{-1} \circ g_{\bm{\gamma}} \circ \psi_{\bm{\theta}} \right) (\bm{x}).
\end{equation*}
Note that the pullback operator $D_{\bm{y}}\psi^*$ indeed corresponds to $D_{\bm{x}} (\psi_{\bm{\theta}}(\bm{x}))^{-1}$ as discussed in \S~\ref{subsubsec:pullbackConstrainedOpt}.
Then, we compute the time derivative of the Lyapunov function $\tilde{V}(\bm{x}(t))$ as follows,
\begin{equation}
    \begin{split}
        \frac{d\tilde{V}(\bm{x}(t))}{dt} 
        &= \frac{d (\tilde{V} \circ (\psi_{\bm{\theta}}^{-1} \circ \psi_{\bm{\theta}}) \circ \bm{x})(t) }{dt} = \frac{d (\tilde{V} \circ \psi_{\bm{\theta}}^{-1}) \circ (\psi_{\bm{\theta}} \circ \bm{x})(t)}{dt} \\
        &= \frac{d (V \circ \bm{y}) (t)}{dt} = \frac{d V (\bm{y}(t))}{dt} ,
    \end{split}
    \label{eq:derivative_new_lyapunov}
\end{equation}
where we first introduced the identity mapping $\bm{I} = \psi_{\bm{\theta}}^{-1} \circ \psi_{\bm{\theta}}$ and then applied some regrouping.
This implies that the new Lyapunov function $\tilde{V}$ satisfies all Lyapunov conditions defined as,
\begin{equation*}
    \begin{split}
        & \tilde{V}(\bm{x}^*) = V(\psi_{\bm{\theta}}(\bm{x}^*)) =  V(\bm{y}^*) = 0 , \quad
        \dot{\tilde{V}}(\bm{x}^*) = \dot{V}(\psi_{\bm{\theta}}(\bm{x}^*)) =  \dot{V}(\bm{y}^*) = 0 , \\
        & \tilde{V}(\bm{x}) = 
        V(\psi_{\bm{\theta}}(\bm{x})) = V(\bm{y}) > 0, \ \forall \ \bm{x} \neq \bm{x}^* , \quad 
        \dot{\tilde{V}}(\bm{x}) = \dot{V}(\psi_{\bm{\theta}}(\bm{x})) = \dot{V}(\bm{y}) < 0, \  \forall \ \bm{x} \neq \bm{x}^* .
    \end{split}
\end{equation*}
We have therefore proved that our RSDS provides globally asymptotically stable vector fields on Riemannian manifolds.
From the above derivation, we also showed that a diffeomorphism can be used to describe a \emph{change of coordinates} for Riemannian manifolds, as briefly mentioned in \S~\ref{sec:background}. 
Thus, we can regard $\bm{x}$ and $\bm{y}$ as two coordinate representations for the same underlying dynamical system on $\mathcal{M}$.
Hence, the original dynamical system with coordinates $\bm{x} \in \manifold$ is characterized by the same stability properties as the canonical dynamical system with coordinates $\bm{y}$ on $\mathcal{N}$.

\subsection{Quasi-global Asymptotic Stability}
For Riemannian manifolds of particular topologies, it is not possible to guarantee \emph{global} stability. 
To begin with, the Poincaré-Hopf theorem establishes that for any vector field $v$ on $\mathcal{M}$, the sum of the Poincaré indices over all the isolated zeroes is equal to the Euler characteristic $\chi(\mathcal{M})$. 
Global convergence to a single equilibrium point means a single zero with index $1$. 
However, for compact manifold such as the $2$-sphere, which has Euler characteristic $2$, there must exist at least another zero. 
This means that global asymptotically stability is not achievable for such a manifold.

To address this problem, we assume the following: Let us consider again the case of the Sphere manifold. 
For a geodesic vector field defined on the Sphere, the remaining isolated zero corresponds to the cut locus of the attractor point $\operatorname{Cut}(\bm{x}^*)$, where the vector field converges to. 
Our assumption is that a geodesic vector field converges to a single attractor $\bm{x}^*$ for all points on $\mathcal{M}$ except the attractor's cut locus $\operatorname{Cut}(\bm{x}^*)$. 
Such an assumption implies that theoretically we may only guarantee ``\emph{quasi-global} asymptotic stability''. 
Although our assumption might be restrictive when designing geodesic vector fields on compact Riemannian manifolds (e.g., the Sphere), there exist several non-compact manifolds without cut locus (e.g., the manifold of symmetric positive definite matrices)~\citep{Pennec20:ManifoldValuedImagesProc}, for which the aforementioned assumption may not be necessary.

\section{Final RSDS framework}
\label{app:finalFramework}
We here prove the final RSDS equation~\eqref{eq:final_diffeomorphism_based_vf_equation} is equivalent to the original diffeormorphism-based learning framework~\eqref{eq:diffeomorphism_based_svf_equation} and give more details about network structures for each component.
First, let us summarize our framework as discussed in \S~\ref{sec:approach}.
Given a manifold $\mathcal{M}$, we can evaluate the velocity $\dot{\bm{x}} \in \mathcal{T}_{\bm{x}}\mathcal{M}$ at location $\bm{x} \in \mathcal{M}$ using~\eqref{eq:diffeomorphism_based_svf_equation} and the canonical dynamics defined in \S~\ref{subsec:GeodesicVF}.
As a result, we obtain the whole RSDS framework formulated as
\begin{equation*}
    \dot{\bm{x}} = (D_{\bm{y}}\psi_{\bm{\theta}}^\star \circ g_{\bm{\gamma}} \circ \psi_{\bm{\theta}})(\bm{x}) =  D_{\bm{y}} \psi_{\bm{\theta}}^\star (\dot{\bm{y}}) \quad
    \mathrm{with} \quad     g_{\bm{\gamma}}(\bm{y}) = 
    \begin{cases}
        k_{\bm{\gamma}}(\bm{y}) \frac{\operatorname{Log}_{\bm{y}}(\bm{y}^*)}{\lVert \operatorname{Log}_{\bm{y}}(\bm{y}^*) \rVert_2}, \  \forall \  \bm{y} \neq \bm{y}^* \\
        \bm{0}, \ \bm{y} = \bm{y}^*
    \end{cases} ,
\end{equation*}
where 
\begin{itemize}
    \item $\psi_{\bm{\theta}}: \mathcal{M} \rightarrow \mathcal{N}, \bm{x} \mapsto \bm{y}$ is a learnable diffeomorphism parameterized by the Neural MODE,
    \item $g_{\bm{\gamma}}: \mathcal{N} \rightarrow \mathcal{T}_{\bm{y}}\mathcal{N}$ is a parameterized geodesic vector field with the equilibrium point $\bm{y}^*$,
    \item $D_{\bm{y}}\psi_{\bm{\theta}}^\star: \mathcal{T}{\bm{y}}\mathcal{N} \rightarrow \mathcal{T}{\bm{x}}\mathcal{M}$ is the pullback operator,
    \item $k_{\gamma}(\bm{y})$ is a positive scaling factor $\forall \  \bm{y} \in \mathcal{N}$.
\end{itemize}
This learning framework is mainly parametrized by $\bm{\theta}$ (corresp. the diffeomorphism) and $\bm{\gamma}$ (corresp. the scaling factor of the canonical dynamics).
These two parameters sets provide large capacity to learn very complex vector fields.
However, the smoothness of the learned vector fields can not be easily guaranteed since it is affected by the aforementioned parameters, which are separately constructed with two independent neural networks.
Additionally, since the diffeomorphism is responsible for both the direction and magnitude of the vector field, the diffeomorphism network $\psi_{\bm{\theta}}$ needs to find a compromise, which potentially weakens the capacity to learn the direction of the vector field.
To address these two issues, we introduce the following improvements.
We reparameterize the direction and magnitude of the learned vector fields separately, as follows
\begin{equation}
    \begin{split}
        \dot{\bm{x}} &= (D_{\bm{y}}\psi_{\bm{\theta}}^\star \circ k_{\bm{\gamma}} g_n \circ \psi_{\bm{\theta}})(\bm{x}) = D_{\bm{y}}\psi_{\bm{\theta}}^\star \  k_{\bm{\gamma}}(\bm{y}) \  g_n(\bm{y}) ,\\
        &= k_{\bm{\gamma}}(\bm{y}) \  D_{\bm{y}}\psi_{\bm{\theta}}^\star \  g_n(\bm{y})
        = \tilde{k}_{\bm{\gamma}}(\bm{x}) \  (D_{\bm{y}}\psi_{\bm{\theta}}^\star \circ  g_n \circ \psi_{\bm{\theta}})(\bm{x}),
    \end{split}
\end{equation}
where $g_n: \mathcal{N} \rightarrow \mathcal{T}{\bm{y}}\mathcal{N}$ denotes the normalized geodesics vector field as described in \S~\ref{subsec:GeodesicVF}, and $\tilde{k}_{\bm{\gamma}}(\bm{x}) = k_{\bm{\gamma}}(\psi_{\bm{\theta}}^{-1}(\bm{y}))$ is an intermediate variable only for the purpose of the derivation.
We then redefine a new scaling factor $\hat{k}_{\bm{\gamma}}(\bm{x}):= \tilde{k}_{\bm{\gamma}}(\bm{x}) \lVert (D_{\bm{y}}\psi_{\bm{\theta}}^\star \circ  g_n \circ \psi_{\bm{\theta}})(\bm{x}) \rVert_2$, which leads to 
\begin{equation}
        \dot{\bm{x}}
        = \tilde{k}_{\bm{\gamma}}(\bm{x}) (D_{\bm{y}}\psi_{\bm{\theta}}^\star \circ  g_n \circ \psi_{\bm{\theta}})(\bm{x}) = \hat{k}_{\bm{\gamma}}(\bm{x}) \frac{(D_{\bm{y}}\psi_{\bm{\theta}}^\star \circ  g_n \circ \psi_{\bm{\theta}})(\bm{x})}{\lVert (D_{\bm{y}}\psi_{\bm{\theta}}^\star \circ  g_n \circ \psi_{\bm{\theta}})(\bm{x}) \rVert_2} ,
\end{equation}
which is the proposed RSDS~\eqref{eq:final_diffeomorphism_based_vf_equation}, as defined in \S~\ref{sec:approach}.
Since there are no fundamental changes on the learning framework with the above modifications, the stability properties remains unchanged as long as the newly defined scaling factor $\hat{k}_{\bm{\gamma}}(\bm{x})$ is strictly positive.

To ensure the smoothness of the magnitude part and provide good extrapolation in regions beyond the demonstrations, we leverage RBF networks to parametrize $\hat{k}_{\bm{\gamma}}(\bm{x})$.
Specifically, we define the scaling factor $\hat{k}_{\bm{\gamma}}$ as $\hat{k}_{\bm{\gamma}}(\bm{x}) = e^{\kappa_{\bm{\gamma}}(\bm{x}) + \epsilon}$, with $\kappa_{\bm{\gamma}}: \mathbb{R}^n \supset \mathcal{M} \rightarrow \mathbb{R}$, and $\epsilon$ being a very small constant to ensure positive definiteness numerically.
Note that the RBF network needs to be adapted to the Riemannian setting as $\bm{x} \in \mathcal{M}$,. 
To do so, we replace the usual $L_2$-norm by the Riemannian distance,  
\begin{equation}
    \kappa_{\bm{\gamma}}(\bm{x}) = \sum_i w_i \cdot \phi\left(\frac{d_{\bm{x}}(\bm{x}, \bm{c}_i)}{\sigma_i}\right) ,
    \label{eq:manifold_rbf_net}
\end{equation}
where $w_i$ are linear weights, $\bm{c}_i$ are the RBF centers (obtained via $k$-means on manifolds), $\sigma_i$ is similar to the standard deviation for Gaussian distributions, $\phi$ represents the RBFs, e.g. $\phi(r) = e^{-r^2}$, and $d_{\bm{x}}$ is the Riemannian distance induced by Riemannian metrics as defined in \S~\ref{sec:background} and Table~\ref{tab:SphereOperations}.

Regarding the parameterization of the diffeomorphism $\psi_{\bm{\theta}}$, we use a simple fully connected neural network (FCNN).
While a classical FCNN can be considered as a general function approximator in Euclidean space, dynamics model $f_{\bm{\theta}}$ represented as a neural MODE requires to account for constraint that the output of network must lie on the tangent space at the input.
Due to this, we add an additional projection operator $\proju$ to the last layer of the network to project its output onto the tangent space at the input location.
More formally, let us assume that states $\bm{z}(t)$ lie on the manifold $\mathcal{M}^d$ which is embedded in an ambient space $\mathbb{R}^n$.
Given a $\mathcal{C}^1$ network $\eta_{\bm{\theta}}: \mathbb{R}^n \times \mathbb{R} \rightarrow \mathbb{R}^n$, the vector field on the manifold $f_{\bm{\theta}}: \mathbb{R}^n \times \mathbb{R} \rightarrow \mathcal{T} \mathcal{M}$ can be obtained as,
\begin{equation}
    f_{\bm{\theta}}(\bm{z}(t), t) = \proju_{\bm{z}(t)} \circ \eta_{\bm{\theta}}(\bm{z}(t), t) .
    \label{eq:vf_ambient_projection}
\end{equation}
Note that the state variable $\bm{z}$ lies on a Riemannian manifold, but we do not explicitly consider this in the input layer of the network, as we are instead accounting for the ambient space in which the corresponding manifold is embedded. 

In summary, the final RSDS learning framework provides several advantages.
On the one hand, a separate magnitude parameterization $\hat{k}_{\bm{\gamma}}(\bm{x})$ allows us to design the network such as the Riemannian RBF network~\eqref{eq:manifold_rbf_net} performs better extrapolation in regions where no training data are available and guarantees smoothness.
Since $\hat{k}_{\bm{\gamma}}$ must be positive definite on the entire manifold, the Riemannian RBF network can provide a constant magnitude value $\hat{k}_{\bm{\gamma}} = 1$ with $\kappa_{\bm{\gamma}} = 0$ when $\bm{x}$ is far away from demonstrations. 
Note that we tried to use FCNN for this parameterization but it outputs very small values $\hat{k}_{\bm{\gamma}} \rightarrow 0$ for datapoints beyond the demonstrations region, despite we use an exponential function to guarantee positive definiteness.
The reason for this is that FCNN can not provide predictable values for inputs which are too dissimilar from the training data, and it turns out to output $\kappa_{\bm{\gamma}} \rightarrow - \infty$ for unseen inputs.
On the other hand, the parameterization of the diffeomorphism $\psi_{\bm{\theta}}$ can completely focus on learning the vector field directions, improving the reproduction accuracy.

\subsection{A short discussion on Riemannian approaches for robot motion generation}
Note that Riemannian geometry has been also leveraged to design robot motion policies that build on the geometry of classical mechanical systems, as proposed in recent works on Riemannian Motion Policies~\citep{Cheng21:RMPflow}, and more recently in a generalization called Geometric Fabrics~\citep{Van22:GeomFabrics}. 
In this context, RSDS may be seen as learning a control policy represented by a first-order dynamical system, which may be used as a motion policy into the RMPs framework, as RMPs provide a geometric robot motion structure that combines several motion polices with associated Riemannian metrics. 
In other words, we may fuse RSDS skills with additional motion policies, like obstacle avoidance behaviors, under the RMPs framework~\citep{Cheng21:RMPflow}. 
Note that RSDS represents a learned first-order dynamic motion policy that does not explicitly consider a specific type of underlying mechanical system. 
In contrast, RMPs build on Riemannian geometric control to design robot motion policies whose geometry is characterized by, e.g., the associated inertia matrix of the robot dynamics.

It is worth highlighting the difference on the geometric aspects regarding RMPs and RSDS. 
Broadly speaking, the Riemannian structure of RMPs arises from the Riemannian metric associated to the kinetic energy of the robot dynamics~\citep{Cheng21:RMPflow}, while the Riemannian structure of RSDS comes from the fact that the state variable $\bm{x}$ of the robot in operational space lies on a Riemannian manifold (due to the orientation representation).
Notice that these two different ways in which Riemannian geometry is leveraged in robotics are complementary to each other. 
On a related note, Riemannian geometry has been recently leveraged to formulate robot motion generation mechanisms that build on geodesics on learned Riemannian manifolds~\citep{BeikMohammadi:GeodesicMotionSkill,BeikMohammadi22:ReactiveManifolds}.
This approach differs from the aforementioned methods as the Riemannian geometry is not intrinsically given by the system dynamics or geometric constraints of the system state.
Instead, the geometric aspect arises from the assumption that trajectories of a robot motion skill define a nonlinear smooth surface that can be interpreted as a data-driven Riemannian manifold. 

\section{Computation of the Pullback Operator}
\subsection{Pullback operator via constrained optimization}
\label{app:PullbackConstrainedOpt}
Here we provide more insights about the loss of rank of the differential $D_{\bm{x}} \psi_{\bm{\theta}}$ and derive a pullback operator via constrained optimization using the $\SphereManifold^d$ manifold as a study case.
We consider a $2$-Sphere manifold embedded in $\mathbb{R}^3$, whose tangent space has the same dimensionality as the manifold, namely, $\dim(\mathcal{T}_{\bm{x}}\mathcal{M}) = 2$.
This implies that the operator $D_{\bm{x}} \psi_{\bm{\theta}}(\bm{x})$ should be a mapping between $2$-dimensional spaces.
However, since we embed the $2$-Sphere and its tangent spaces in $\mathbb{R}^3$, the matrix representation for $D_{\bm{x}} \psi_{\bm{\theta}}(\bm{x})$ is a $3 \times 3$ matrix.
This matrix is a sort of overparameterization for a mapping between $2$-dimensional spaces, and therefore it is rank-deficient and can not be directly inverted.
The rank deficiency arises from the geometric constraints of the Riemannian manifold, which impose linear dependencies in the vector space spanned by the matrix columns of $D_{\bm{x}} \psi_{\bm{\theta}}(\bm{x})$.

Given the relationship between $\dot{\bm{x}}$ and $\dot{\bm{y}}$ with $D_{\bm{x}} \psi_{\bm{\theta}}(\bm{x}) \dot{\bm{x}} = \dot{\bm{y}}$ and the geometric constraints associated to $\SphereManifold^d$ manifolds, as discussed in \S~\ref{subsubsec:pullbackConstrainedOpt}, we can compute a solution for $\dot{\bm{x}}$ via constrained optimization. 
For our study case on $\SphereManifold^2$, we augment the matrix $D_{\bm{x}} \psi_{\bm{\theta}}$ to include the associated geometric constraint, and then we compute the pseudo-inverse to obtain the final solution, as follows
\begin{equation}
    \begin{split}
        & \begin{bmatrix}
            D_{\bm{x}} \psi_{\bm{\theta}}(\bm{x}) \\
            \bm{x}^\trsp
        \end{bmatrix} 
        \dot{\bm{x}} = 
        \begin{bmatrix}
             \dot{\bm{y}} \\
             0
        \end{bmatrix} , \quad
        \begin{bmatrix}
            D_{\bm{x}} \psi_{\bm{\theta}}(\bm{x}) \\
            \bm{x}^\trsp
        \end{bmatrix}^\trsp 
        \begin{bmatrix}
            D_{\bm{x}} \psi_{\bm{\theta}}(\bm{x}) \\
            \bm{x}^\trsp
        \end{bmatrix}
        \dot{\bm{x}} = 
        \begin{bmatrix}
            D_{\bm{x}} \psi_{\bm{\theta}}(\bm{x}) \\
            \bm{x}^\trsp
        \end{bmatrix}^\trsp 
        \begin{bmatrix}
             \dot{\bm{y}} \\
             0
        \end{bmatrix} , \\
        & \left[ D_{\bm{x}} \psi_{\bm{\theta}}(\bm{x})^\trsp D_{\bm{x}} \psi_{\bm{\theta}}(\bm{x}) + \bm{x} \bm{x}^\trsp \right] \ \dot{\bm{x}} = D_{\bm{x}} \psi_{\bm{\theta}}(\bm{x})^\trsp \dot{\bm{y}} , \\
        & \dot{\bm{x}} = \left[ D_{\bm{x}} \psi_{\bm{\theta}}(\bm{x})^\trsp D_{\bm{x}} \psi_{\bm{\theta}}(\bm{x}) + \bm{x} \bm{x}^\trsp \right]^{-1} D_{\bm{x}} \psi_{\bm{\theta}}(\bm{x})^\trsp \dot{\bm{y}} , \\
        & D_{\bm{y}} \psi_{\bm{\theta}}^\star = \left[ D_{\bm{x}} \psi_{\bm{\theta}}(\bm{x})^\trsp D_{\bm{x}} \psi_{\bm{\theta}}(\bm{x}) + \bm{x} \bm{x}^\trsp \right]^{-1} D_{\bm{x}} \psi_{\bm{\theta}}(\bm{x})^\trsp .
    \end{split}
    \label{eq:derivation_augmened_linear_system}
\end{equation}

This pullback operator $D_{\bm{y}} \psi_{\bm{\theta}}^\star$ is only specific to $\SphereManifold^d$, since the geometric constraint considered here is only valid for this type of manifold.
This means that if we need to learn stable vector fields on different Riemannian manifolds, we will need to manually define the corresponding constraints.
These constraints can be highly non-linear for other Riemannian manifolds (e.g., matrix manifolds such as the space of symmetric positive-definite matrices), which may not accept a close-form solution as our previous study case.
Therefore, although this approach is a potential solution, it is neither general nor scalable for computing the pullback operator.

\subsection{Pullback operator via modified adjoint method}
\label{app:PullbackAdjoint}
Since we can solve an EVP of the Neural MODE~\eqref{eq:neuralode} from $t_e$ to $t_s$ to obtain $\psi_{\bm{\theta}}^{-1}$, we can consider the same adjoint method in App.~\ref{app:adjoint_method} to compute the pullback operator $D_{\bm{y}}\psi_{\bm{\theta}}^\star = D_{\bm{y} } (\psi_{\bm{\theta}}^{-1})$.
We design a new loss function $\mathcal{L}_i = q_i(\bm{z}(t_s))$, where $q_i$ is the $i$-th component of an arbitrary function $q: \mathcal{M}^d \rightarrow \mathbb{R}^d$.
Then, we define a new adjoint variable $\bm{A}^*(t) = D_{\bm{z}(t)} (\psi_{\bm{\theta}}^{-1})$, similarly to computation of differential $D_{\bm{x}} \psi_{\bm{\theta}}$ in App.~\ref{app:adjoint_method}.
As a result, we obtain an adjoint ODE $\dot{\bm{A}}^*(t) = - \bm{A}^*(t) D_{\bm{z}(t)}  f_{\bm{\theta}}(\bm{z}(t), t)$ with starting value $\bm{A}(t_s) = D_{\bm{z}(t_s)} \bm{z}(t_s) = \bm{I}$ and end value $\bm{A}^*(t_e) = D_{\bm{z}(t_e)} (\psi_{\bm{\theta}}^{-1}) = D_{\bm{y}} (\psi_{\bm{\theta}}^{-1}) $.
Then, solving the IVP of augmented ODEs composed of the Neural MODE~\eqref{eq:neuralode} and adjoint ODE with adjoint variable $\bm{A}^*(t)$ allows us to compute the pullback operator $D_{\bm{y}} (\psi_{\bm{\theta}}^{-1})$, as described in Algorithm~\ref{alg:differential_inv_diffeomorphism}.
{\SetAlgoNoLine%
\begin{algorithm}
\caption{Diffeomorphism and differential of inverse diffeomorphism} 
\label{alg:differential_inv_diffeomorphism}
\SetKw{Return}{return}
\SetKwProg{Def}{def}{:}{end}
\SetKwInOut{Input}{Input}
\Input{start time $t_s$, start end $t_e$, initial states $\bm{z}(t_s) = \bm{x}$ and $\bm{A}^* (t_s) = \bm{I}$}
\Indp $\bm{s}_0 = [\bm{x}, \bm{I}]$ \\
$\%$ augmented ODE using~\eqref{eq:neuralode} and adjoint ODE with variable $\bm{A}^*$ \\
\Def{augmented{\_}dynamics$([\bm{z}(t), \bm{A}^* (t)], t)$}{\Return{$\left[f_{\bm{\theta}}(\bm{z}(t), t), - \bm{A}^* (t) D_{\bm{z}(t)}  f_{\bm{\theta}}(\bm{z}(t), t)\right]$} }
$\left[\bm{z}(t_e),  D_{\bm{z}(t_e)} \psi_{\bm{\theta}}^\star\right]$ = ManifoldODESolve$(\bm{s}_0, \text{augmented\_dynamics}, t_s, t_e)$ \\
\Indm \Return{$\bm{z}(t_e)$,  $D_{\bm{z}(t_e)} \psi_{\bm{\theta}}^\star$}
\end{algorithm}
}

\begin{figure}[ht!]
\centering
    \includegraphics[width=.51\textwidth]{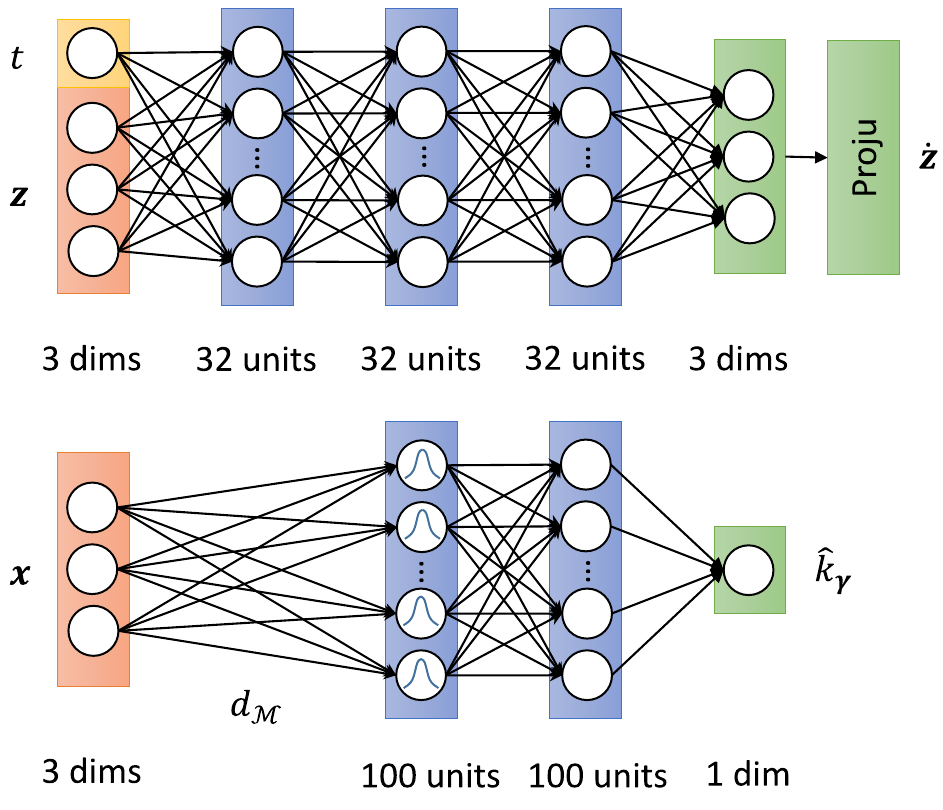}
    \includegraphics[width=.47\textwidth]{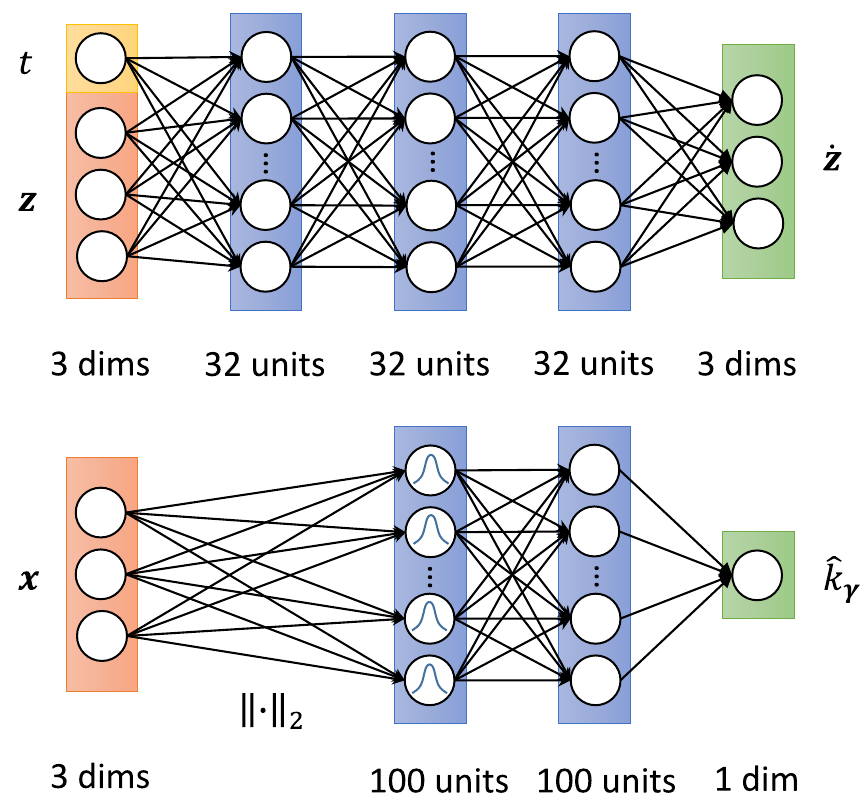}
    \caption{Neural network architectures for \textit{RSDS} and \textit{EuclideanFlow}. \textbf{Left}: In the \textit{RSDS} setting, FCNN uses an extra projection operator in the output layer, and the RBF network uses the Riemannian distance $d_{\mathcal{M}}$. \textbf{Right}: In the \textit{EuclideanFlow} setting, the RBF network uses the Euclidean distance computed by the $L_2$-norm.}
    \label{fig:net_structures_framework}
\end{figure}

In the following we provide the proof for computing the differential of the inverse diffeomorphism~\eqref{eq:inv_block_eq}.
In \S~\ref{subsec:DiffeomorphRM}, we ``discretize'' the diffeomorphism using~\eqref{eq:dynamic_chart_composition}, which overcomes the need for integrators on Riemannian manifolds.
Analogously, we can formulate the inverse diffeomorphism $\psi_{\bm{\theta}}^{-1}: \bm{y} = \bm{z}_k \mapsto  \bm{z}_0 = \bm{x}$ as,
\begin{equation}
    \begin{split}
        \psi_{\bm{\theta}}^{-1} &= \operatorname{Log}_{\bm{z}_0}^{-1} \circ \hat{\psi}_{\bm{\theta}, 0}^{-1} \circ \operatorname{Exp}_{\bm{z}_0}^{-1} \circ \ldots \circ \operatorname{Log}_{\bm{z}_{k-1}}^{-1} , \circ \hat{\psi}_{\bm{\theta}, k-1}^{-1} \circ \operatorname{Exp}_{\bm{z}_{k-1}}^{-1} ,\\
        & = \operatorname{Exp}_{\bm{z}_0} \circ \hat{\psi}_{\bm{\theta}, 0}^{-1} \circ \operatorname{Log}_{\bm{z}_0} \circ \ldots \circ \operatorname{Exp}_{\bm{z}_{k-1}} \circ \hat{\psi}_{\bm{\theta}, k-1}^{-1} \circ \operatorname{Log}_{\bm{z}_{k-1}} .
    \end{split}
    \label{eq:inv_dynamic_chart_composition}
\end{equation}
Now, we focus on a single component $\bm{z}_i = (\operatorname{Exp}_{\bm{z}_i} \circ \hat{f_i}^{-1} \circ \operatorname{Log}_{\bm{z}_i})  \bm{z}_{i+1}$ from~\eqref{eq:inv_dynamic_chart_composition}, and analyze its derivatives, i.e. $D_{\bm{z}_{i+1}} \bm{z}_i = D_{\bm{w}_i(t_{i, s})} \operatorname{Exp}_{\bm{z}_i} \circ D_{\bm{w}_i(t_{i, e})} \hat{\psi}_{\bm{\theta},i}^{-1} \circ D_{\bm{z}_{i+1}} \operatorname{Log}_{\bm{z}_i}$.
Since $\hat{\psi}_{\bm{\theta}, i}: \bm{w}_i(t_{i, s}) \mapsto \bm{w}_i(t_{i, e})$ and $\hat{\psi}_{\bm{\theta}, i}^{-1}: \bm{w}_i(t_{i, e}) \mapsto \bm{w}_i(t_{i, s})$ respectively correspond to solving the equivalent ODE \eqref{eq:equivalentode} forwards and backwards in time in a tangent space (i.e. a Euclidean space), the differential of these mappings corresponds to classical partial derivatives.
This allows us to leverage Algorithm~\ref{alg:differential_inv_diffeomorphism} with the classical ODE solver in Euclidean space to compute $\hat{\psi}_{\bm{\theta}, i}^{-1}$ and its partial derivatives.

\section{Network Architecture}
\label{app:NetworkArchiecture}
In the illustrative examples on $\SphereManifold^2$, under the \textit{EuclideanFlow} setting, we used a fully-connected neural network with an input vector on $\euclideanspace^{4}$ (i.e., the $3$-dimensional state $\bm{x}$ and time), and $3$ hidden layers each, with $32$ hidden units.
We used $\operatorname{tanh}$ as activation function to guarantee a $\mathcal{C}^1$-bounded mapping for modeling the Neural MODE. 
The RSDS architecture has an additional projection operator $\proju$ on the network head to project the output on the tangent space of manifolds (see App.~\ref{app:RiemannianManif}).
The scaling factor $\hat{k}_{\bm{\gamma}}$ is generated using a network composed of an RBF layer and a linear layer without bias. 
In the real robot experiments, we trained our model on $\RtimeS$ (i.e. position and orientation of the end-effector) with the same architecture as our illustrative experiments, except that we use $16$ hidden units for faster computations. 
The network architectures are shown in Fig.~\ref{fig:net_structures_framework}. 

\begin{figure}[t!]
	\centering
	\includegraphics[width =.24\linewidth]{Images/Experiments/Sphere/riemannianflows/P_Shape_0.png}
	\includegraphics[width =.24\linewidth]{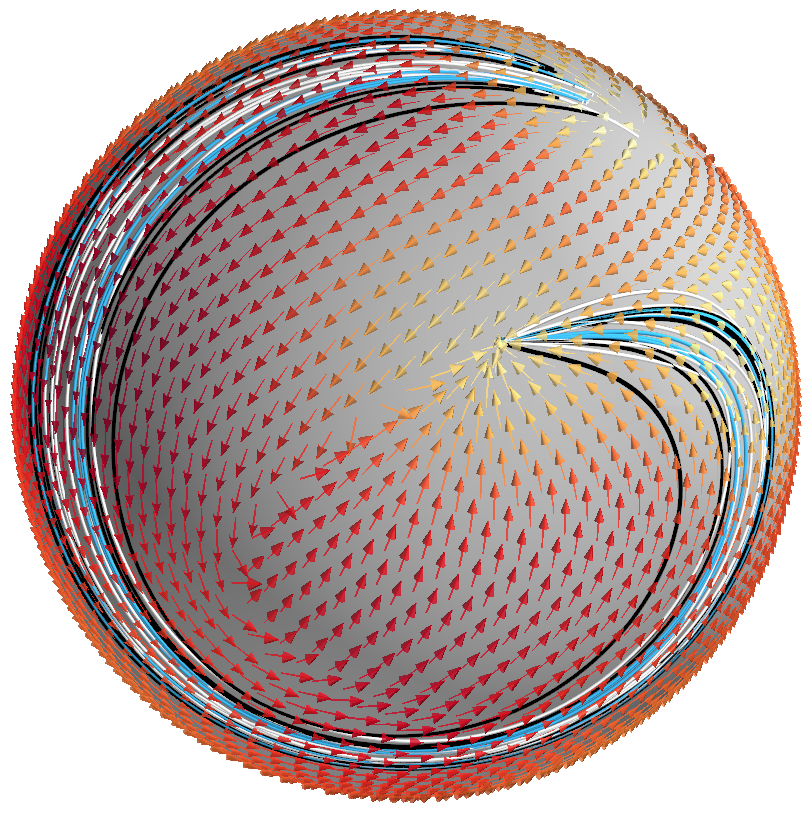}
	\includegraphics[width =.24\linewidth]{Images/Experiments/Sphere/riemannianflows/W_Shape_0.png}
	\includegraphics[width =.24\linewidth]{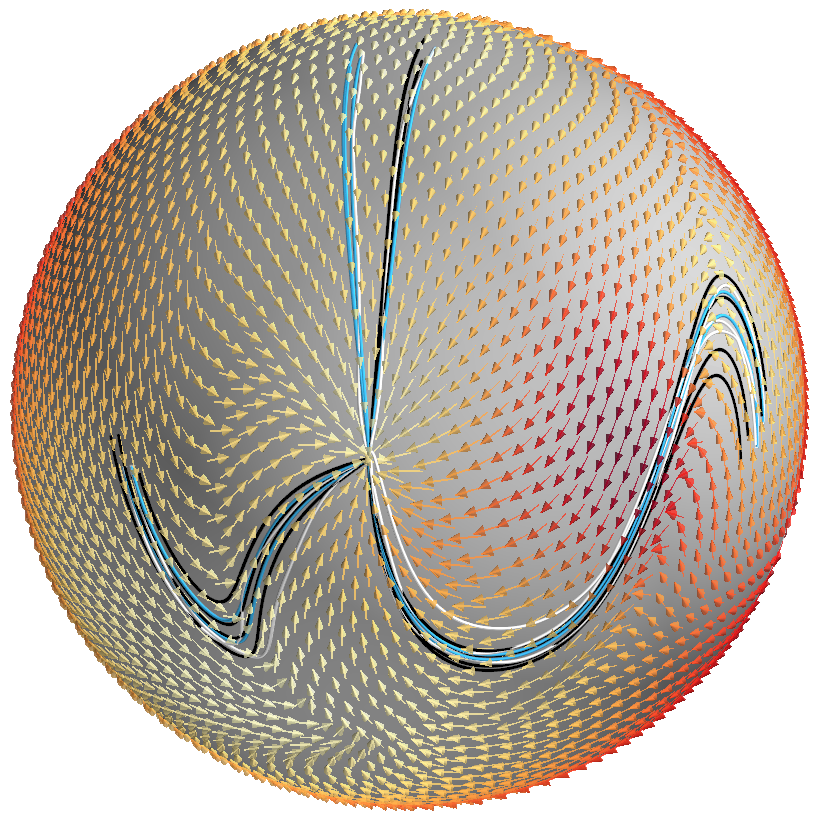}\\
	\includegraphics[width =.24\linewidth]{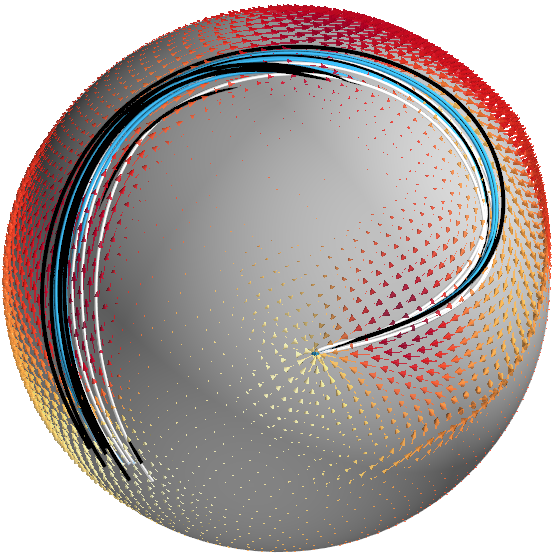}
	\includegraphics[width =.24\linewidth]{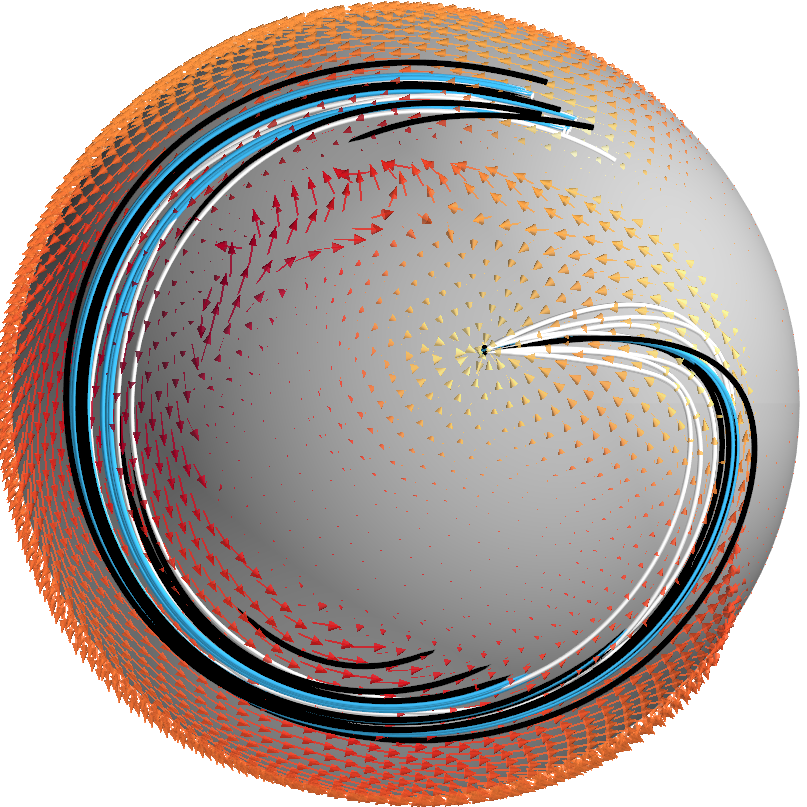}
	\includegraphics[width =.24\linewidth]{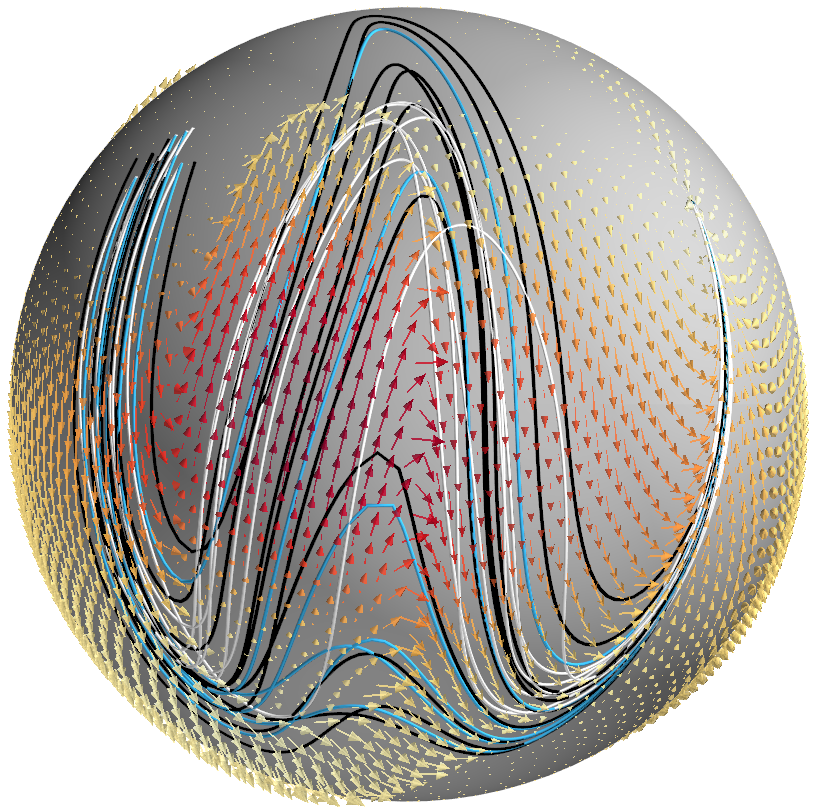}
	\includegraphics[width =.24\linewidth]{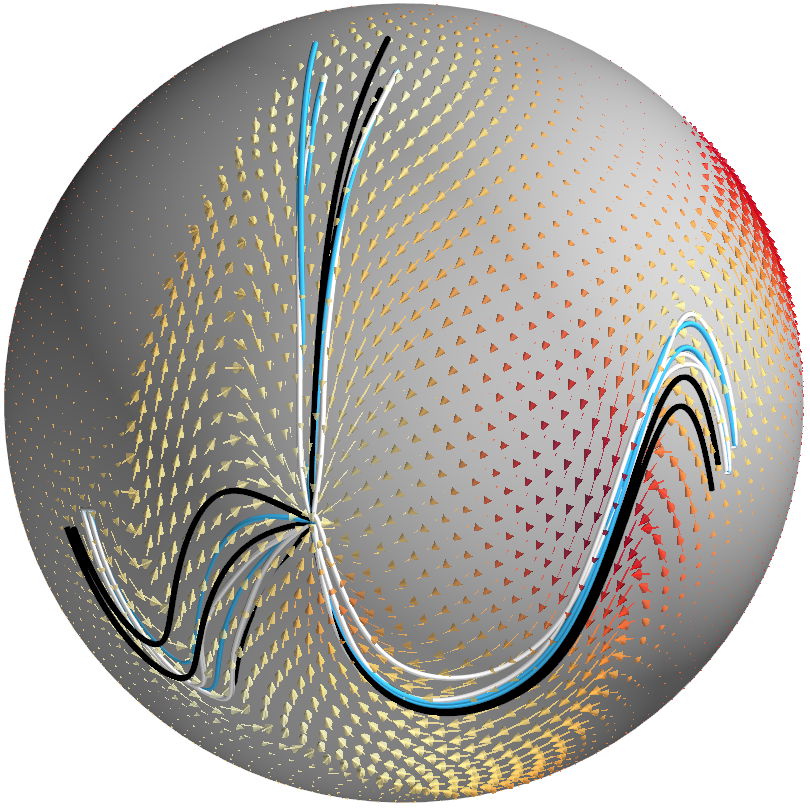}\\
	\includegraphics[width =.24\linewidth]{Images/Experiments/Sphere/euclideanflows/Normal/2/P-Shape-5.png}
	\includegraphics[width =.24\linewidth]{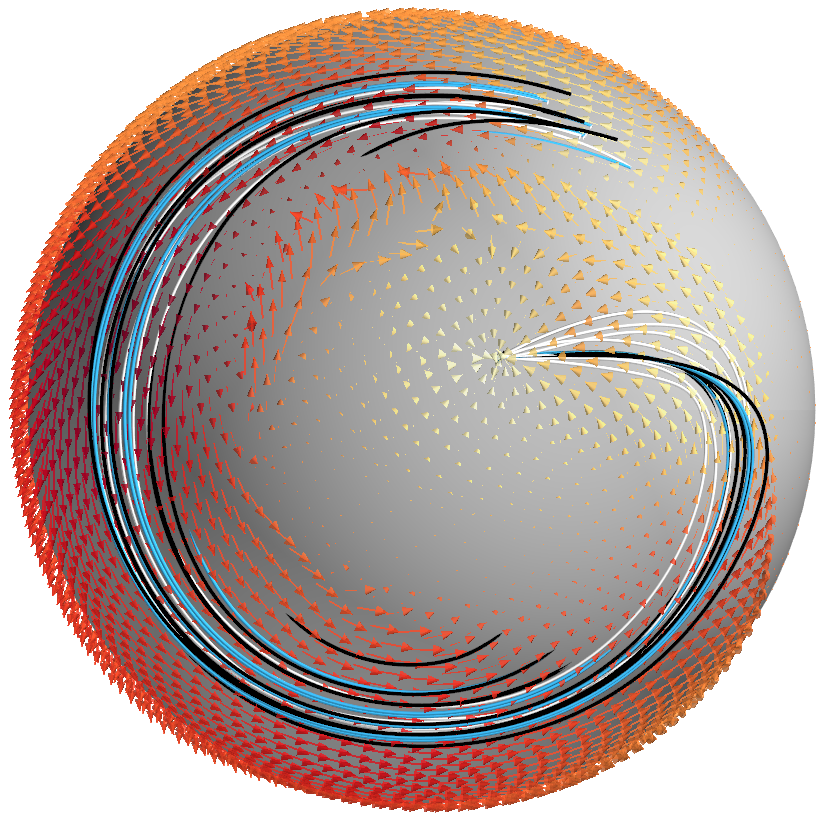}
	\includegraphics[width =.24\linewidth]{Images/Experiments/Sphere/euclideanflows/Normal/2/W_Shape_5_2_cut.png}
	\includegraphics[width =.24\linewidth]{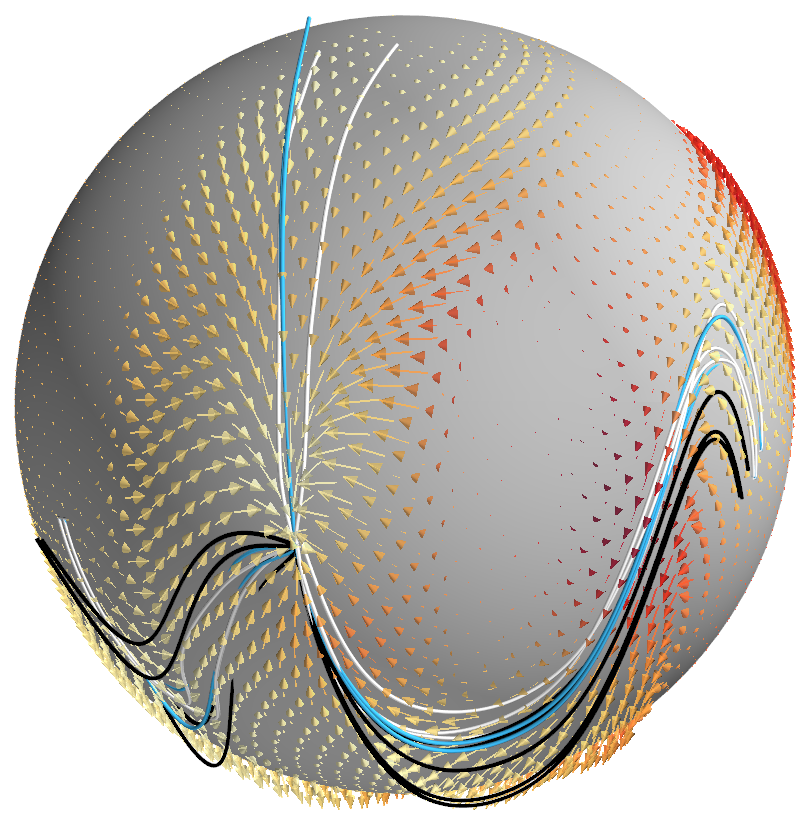}
	\caption{Experiments on LASA dataset on $\SphereManifold^2$ for datasets: $\mathsf{P}$, $\mathsf{G}$, $\mathsf{W}$, and $\mathsf{Multi Models}$. The learned vector field is depicted by arrows (color-coded based on the magnitude), and the demonstrations are shown as white curves. The blue and black trajectories are reproductions starting at the same initial points as the demonstrations and randomly-sampled points around them, respectively. The first row shows the results for the RSDS approach, and the next two rows show the \textit{Projected EuclideanFlow} and \textit{EuclideanFlow} results.}           
	\label{fig:lasa_extended}
\end{figure}

\section{Extended results}
\label{app:extended_results}
\subsection{LASA dataset on $\SphereManifold^2$}
\begin{wrapfigure}[16]{r}{0.5\linewidth}
    \centering
    \includegraphics[width=0.47\textwidth]{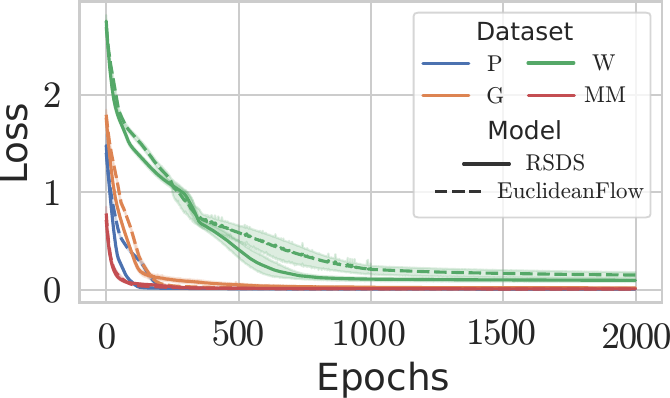}
    \caption{Loss values over $2000$ epochs for datasets: $\mathsf{P}$, $\mathsf{G}$, $\mathsf{W}$, and $\mathsf{Multi Models}$. Several models were trained for each dataset, with the lines and the shaded regions representing the mean and standard deviation.}
    \label{fig:loss_comparison_appendix}
\end{wrapfigure}

\begin{figure}[t!]
\centering
    \includegraphics[width =.24\linewidth]{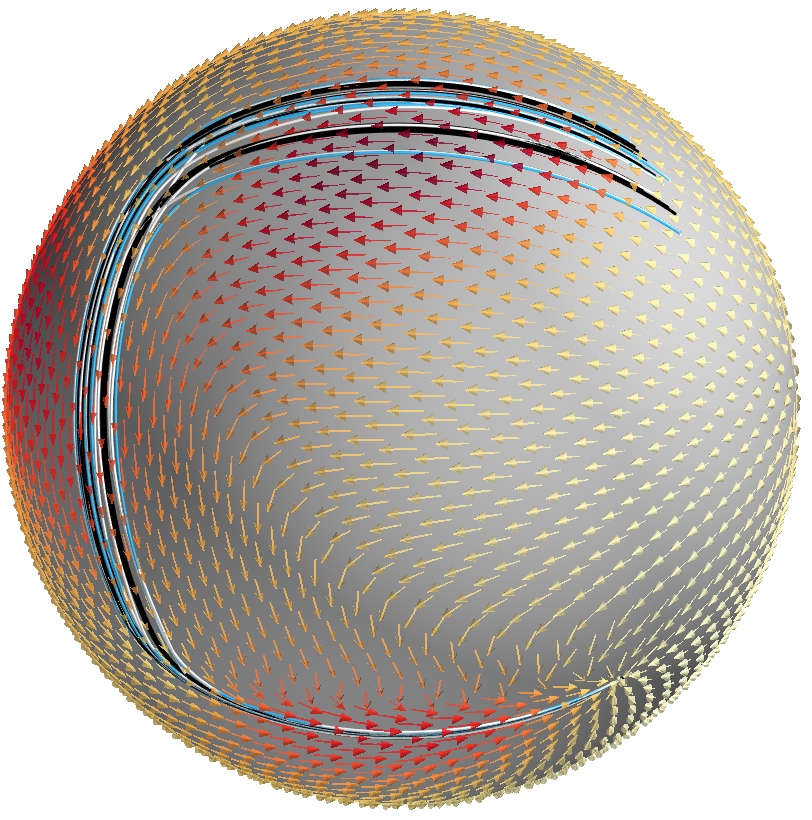}
    \includegraphics[width =.24\linewidth]{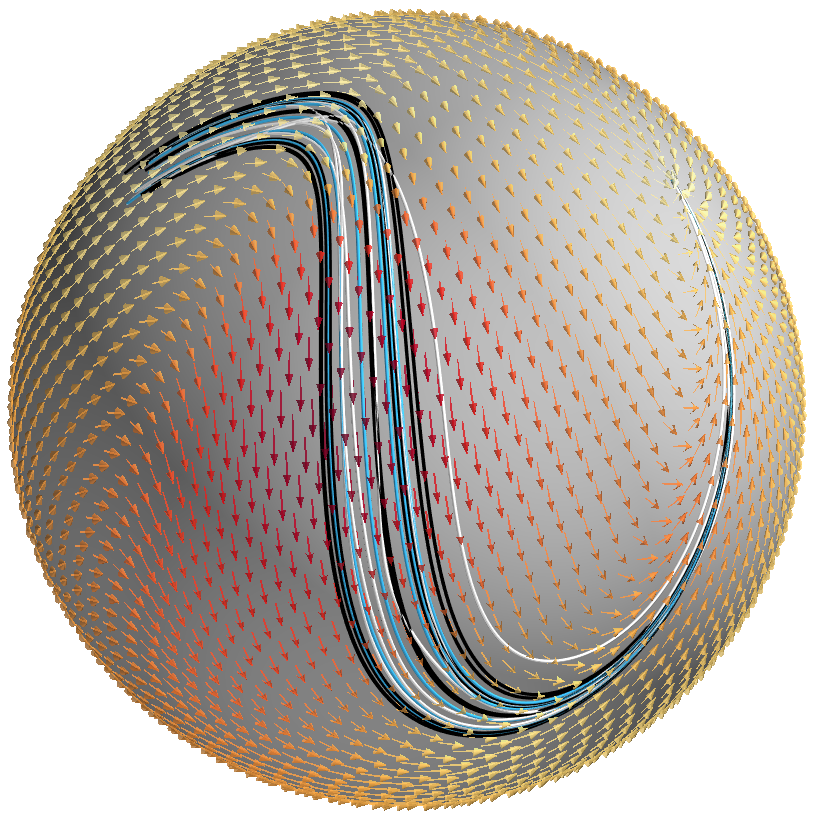}
    \includegraphics[width =.24\linewidth]{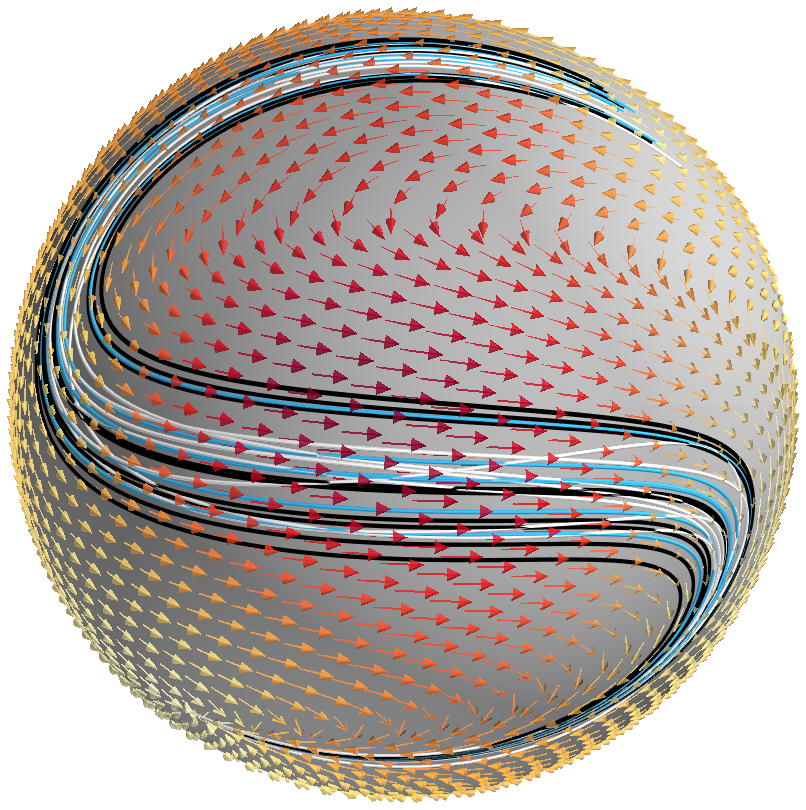}\\
    \includegraphics[width =.24\linewidth]{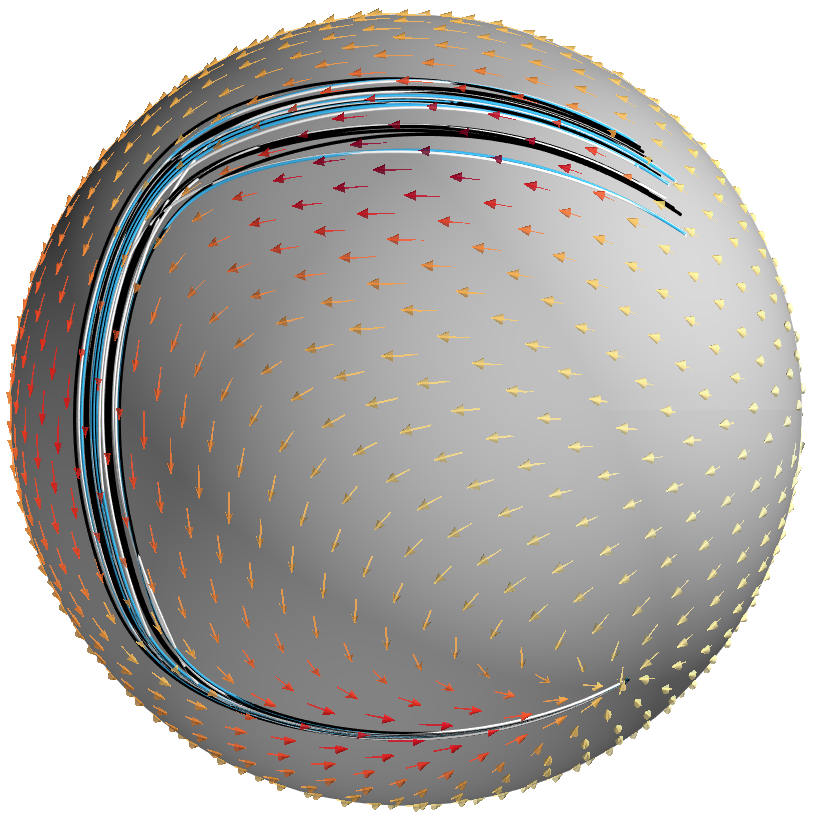}
    \includegraphics[width =.24\linewidth]{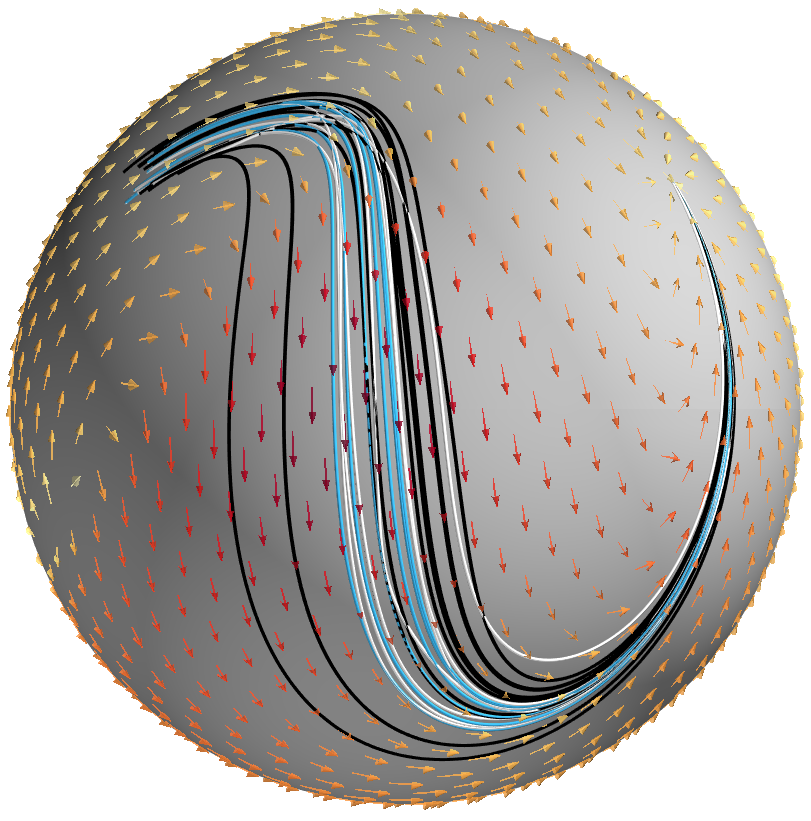}
    \includegraphics[width =.24\linewidth]{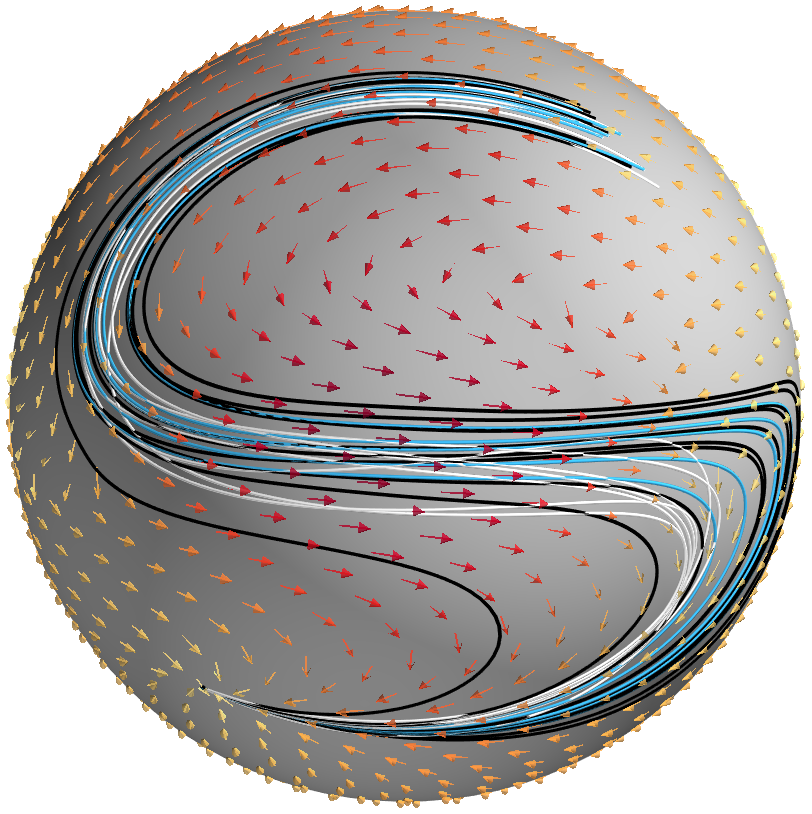}\\
    \includegraphics[width =.24\linewidth]{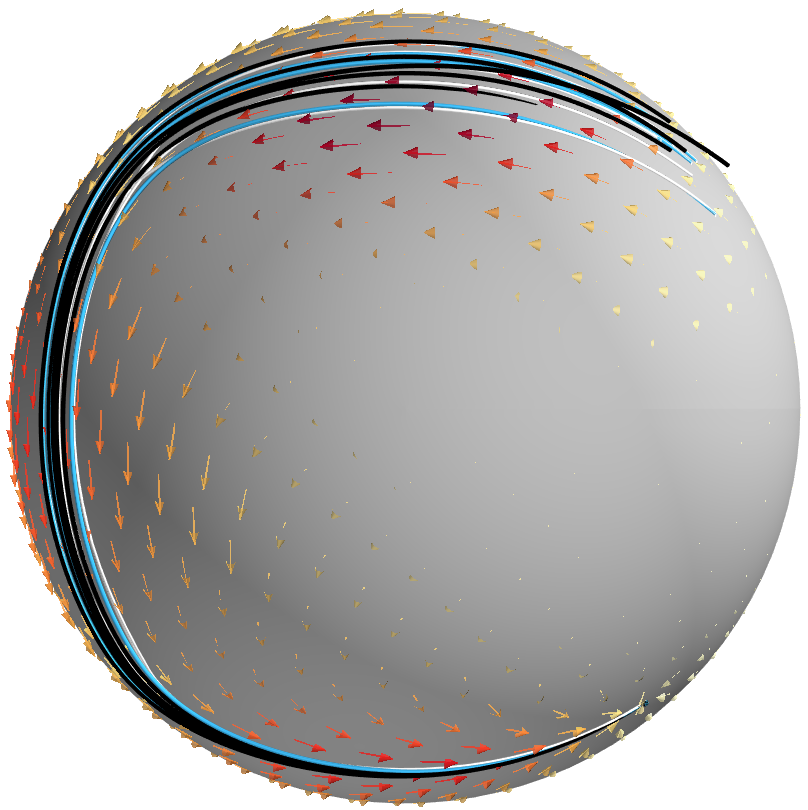}
    \includegraphics[width =.24\linewidth]{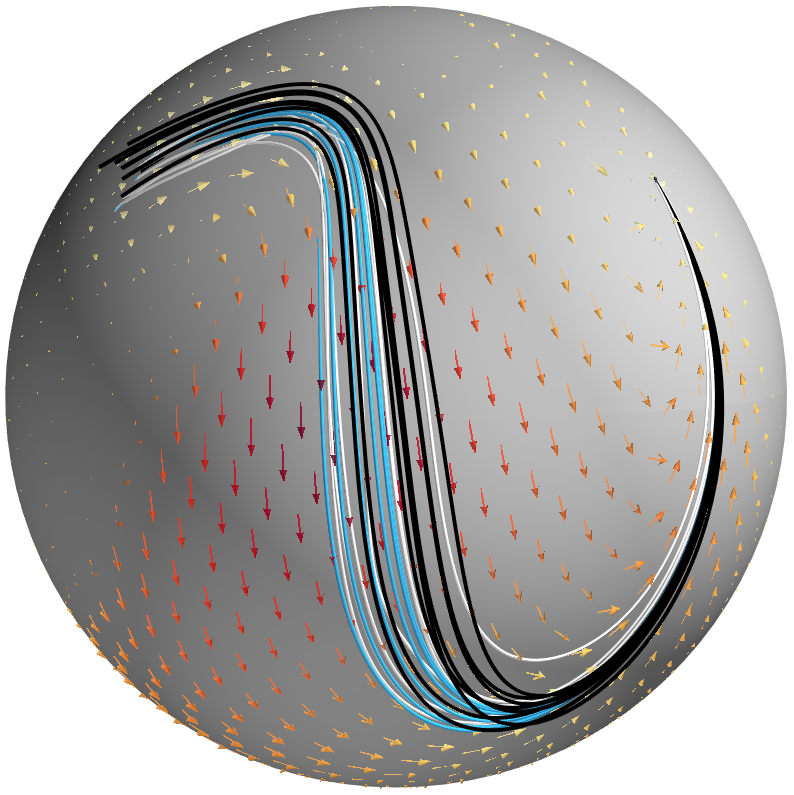}
    \includegraphics[width =.24\linewidth]{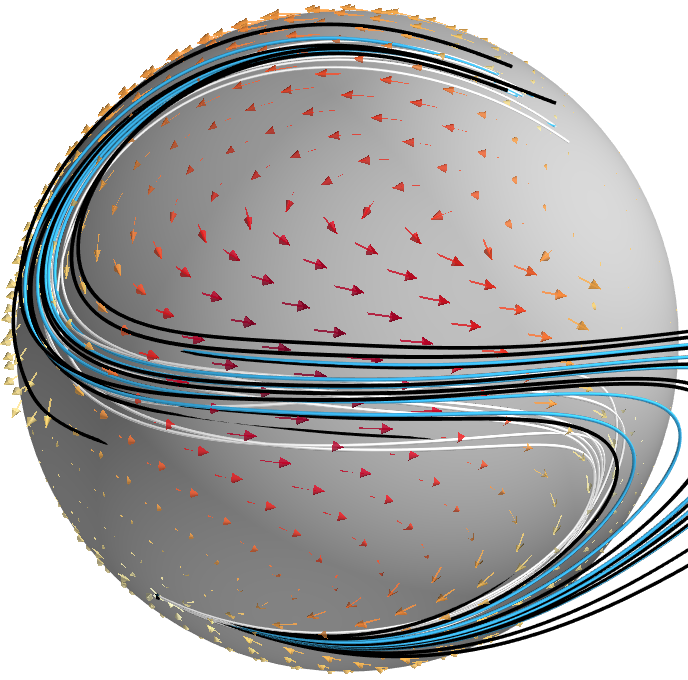}
    \caption{Additional experiments on LASA dataset on $\SphereManifold^2$ for datasets: $\mathsf{SharpC}$, $\mathsf{Spoon}$, and $\mathsf{S}$. The learned vector field is depicted by arrows (color-coded based on the magnitude), and the demonstrations are shown as white curves. The blue and black trajectories are reproductions starting at the same initial points as the demonstrations and randomly-sampled points around them, respectively. The first row shows the results for the RSDS approach, and the next two rows show the \textit{Projected EuclideanFlow} and \textit{EuclideanFlow} results.}  
    \label{fig:lasa_extended_rebuttal}
\end{figure}

In this section, we provide an extended set of experiments to further support the results presented in \S~\ref{sec:result}. The extended results include a complete set of experiments on datasets: $\mathsf{P}$, $\mathsf{G}$, $\mathsf{W}$, and $\mathsf{Multi Models}$, as displayed in Fig.~\ref{fig:lasa_extended}.
Moreover, we carried out three additional experiments on new letters from the LASA dataset, specifically: $\mathsf{SharpC}$, $\mathsf{Spoon}$ and $\mathsf{S}$, whose trajectories significantly differ from the datasets used in the main paper, as shown in Fig.~\ref{fig:lasa_extended_rebuttal}.  
In addition, an extra set of results using \emph{Projected Euclideanflow} is added to the comparison between RSDS and \textit{EuclideanFlow}. 
As mentioned in \S~\ref{sec:result}, this alternative method projects trajectories onto the manifold after computing the integration in the Euclidean space.
Figure~\ref{fig:loss_comparison_appendix} shows the comparison for learning efficiency between the two methods, RSDS and \textit{EuclideanFlow}. Here, the RSDS method generally requires fewer training epochs than \textit{EuclideanFlow}. 
Note that the results regarding \emph{Projected Euclideanflow} and \textit{Euclideanflows} share the same models, therefore, \emph{Projected EuclideanFlow} is omitted from this comparison. 
  
\color{black}

\begin{figure}[t!]
	\centering
	\includegraphics[width =.47\linewidth]{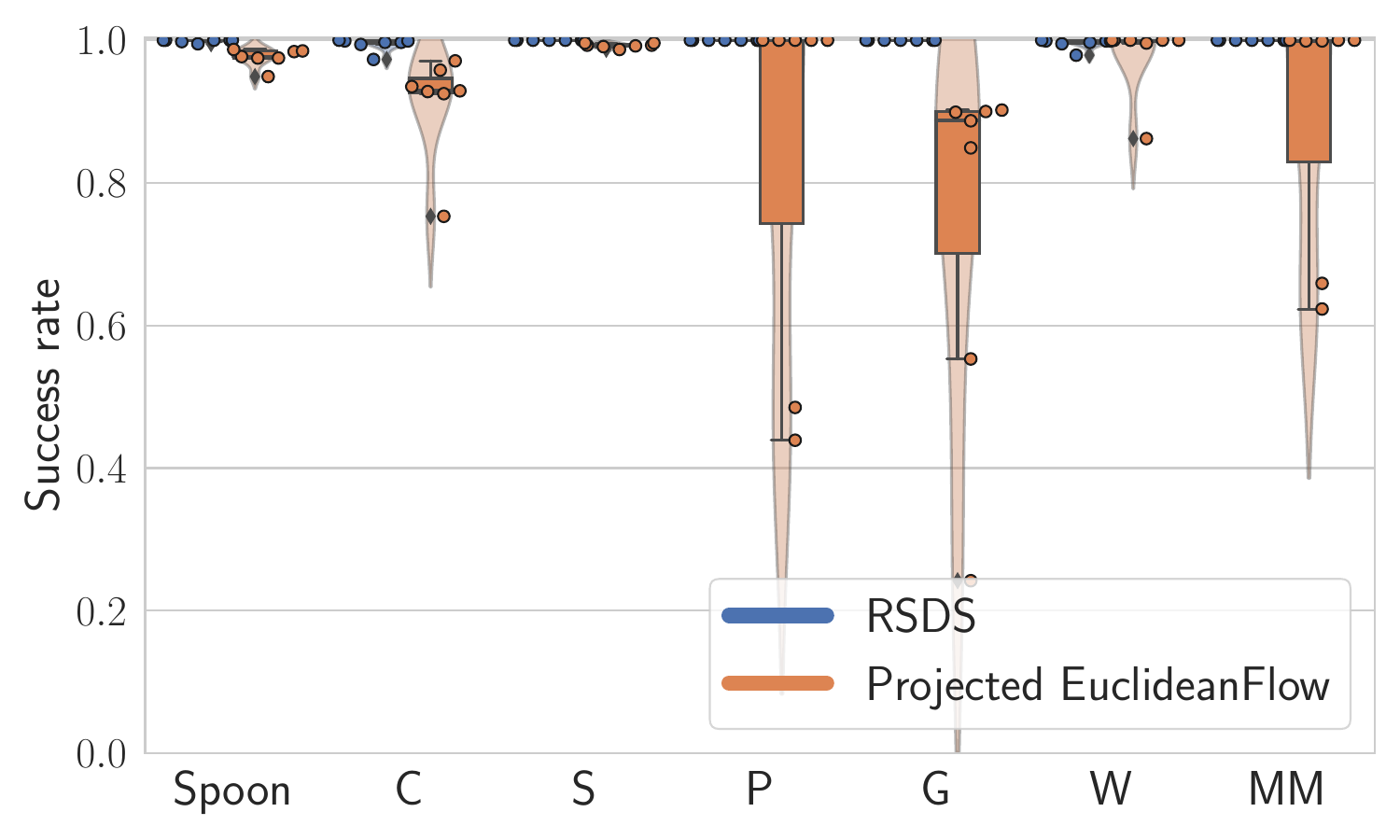}
	\includegraphics[width =.47\linewidth]{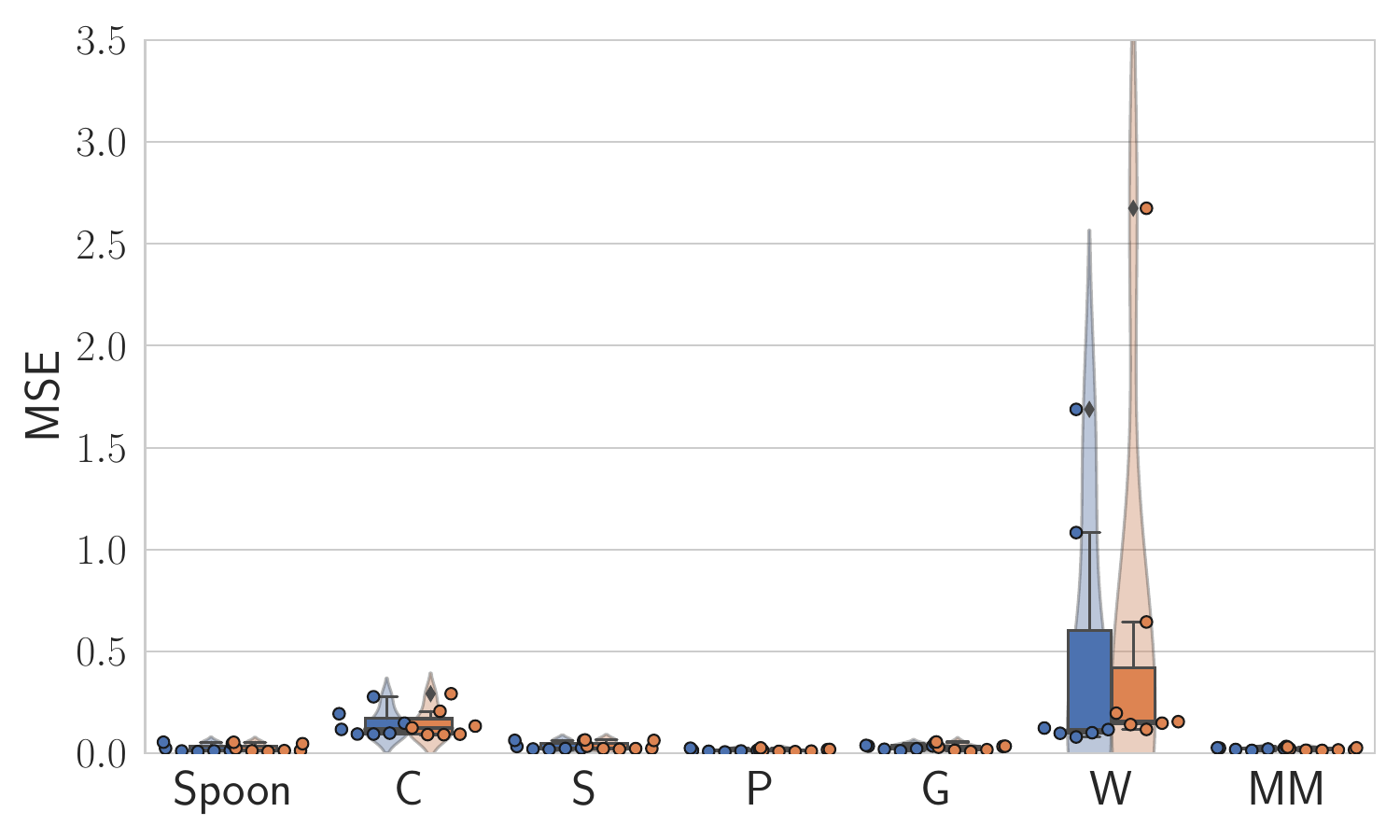}
	\includegraphics[width =.47\linewidth]{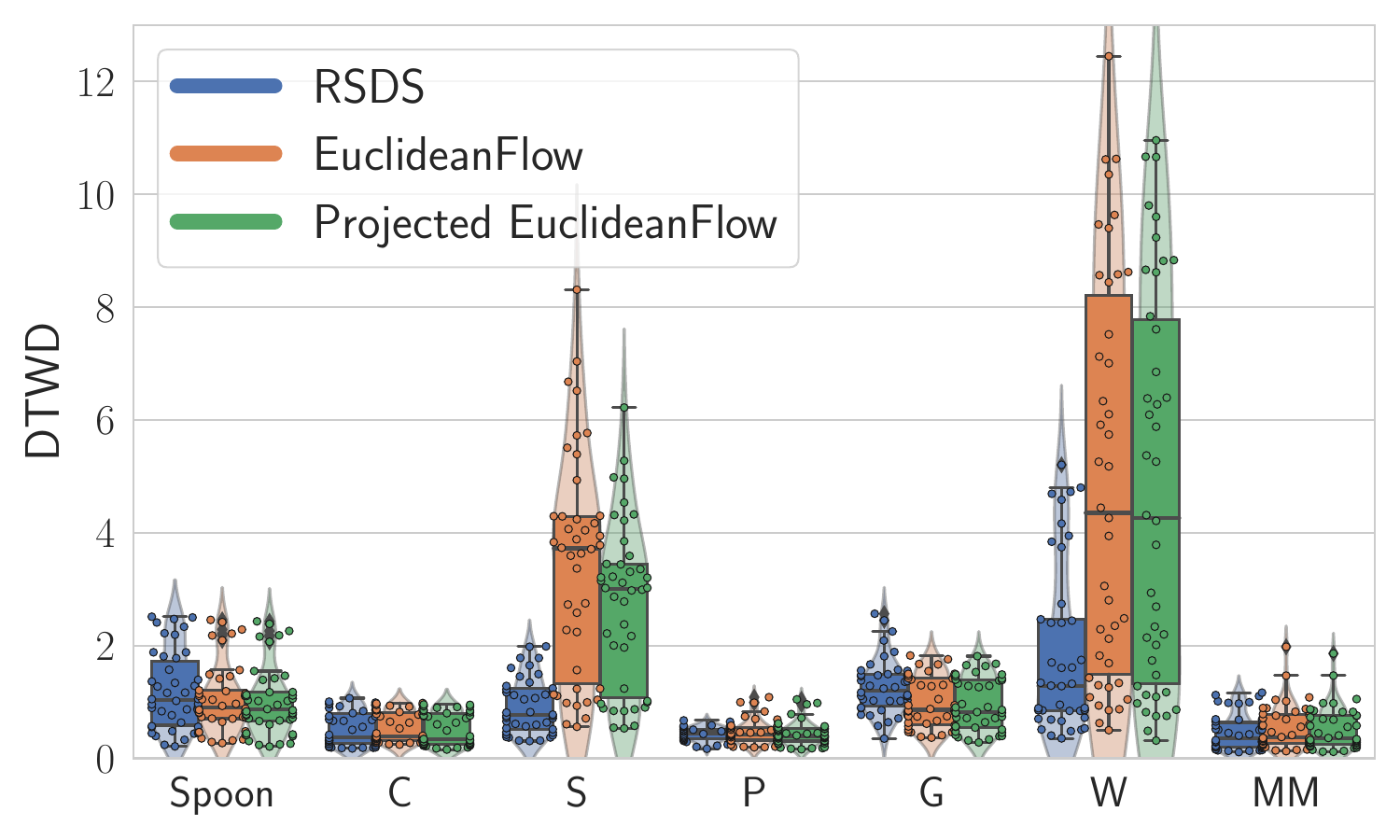}
	\includegraphics[width =.47\linewidth]{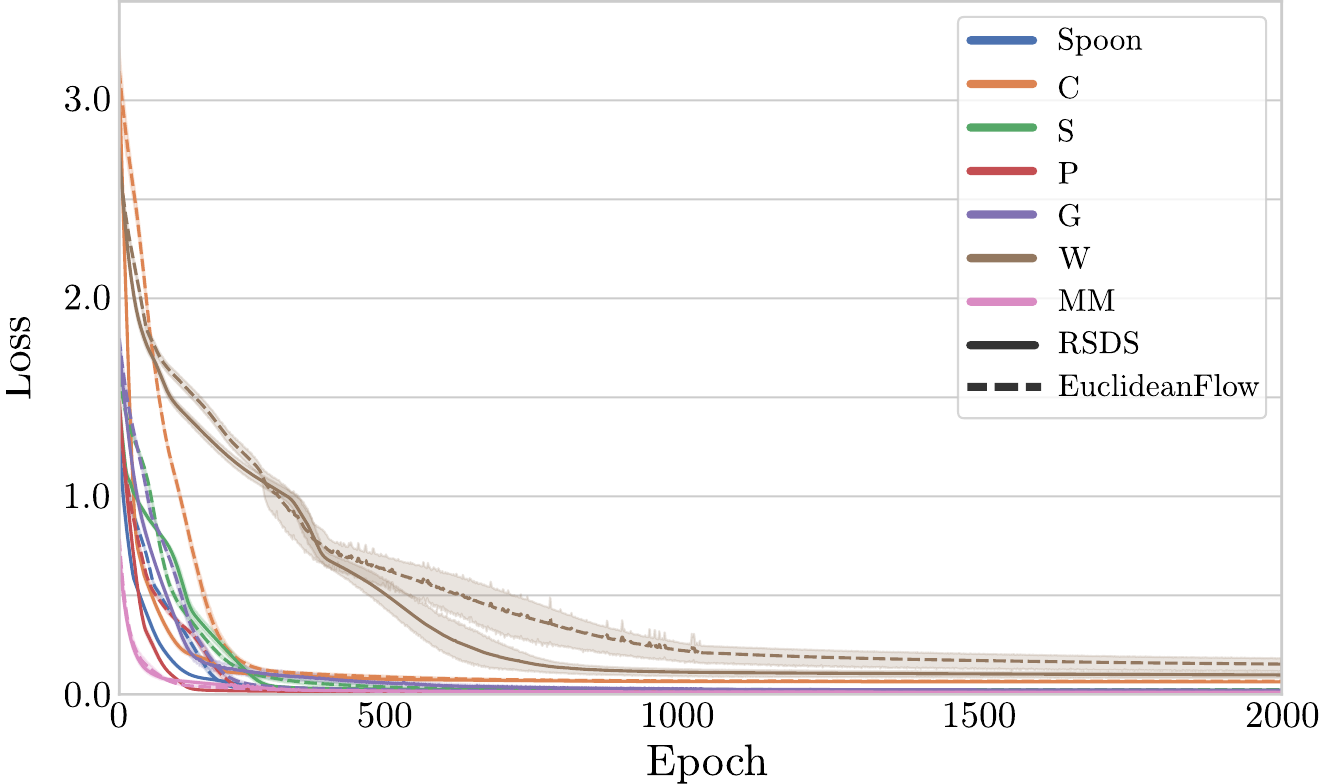}
	\caption{\textit{Top-left}: Average success rate of \textit{RSDS} and \emph{EuclideanFlow} over randomly-sampled initial points on $\SphereManifold^2$ using $7$ different trained models indicated as points. \textit{Top-right:} Average mean square error (MSE) between observed and predicted velocity over data points in the test trajectories indicated as points. \textit{Bottom-left:} Dynamic time warping distance (DTWD) between demonstrations and reproductions. \textit{Bottom-right}: Loss values over $2000 epochs$ for all tested LASA datasets. Several models were trained for each dataset, with the lines and the shaded regions representing the mean and standard deviation.}
	\label{fig:metrics_rebuttal}
\end{figure}

Figure~\ref{fig:lasa_extended} shows the demonstrations as white curves for all datasets, the corresponding learned vector fields and the reproduced trajectories.
The latter, depicted as blue and black trajectories, are rollouts starting from the same initial position as the demonstrations and randomly-sampled points around them, respectively. 
From top to bottom, the rows correspond to trajectories computed using RSDS, \textit{Projected EuclideanFlow}, and \textit{EuclideanFlow}.  
In Fig.~\ref{fig:lasa_extended}, the datasets are ordered from left to right (i.e. from $\mathsf{P}$ to $\mathsf{W}$) based on their difficulty. For example, the $\mathsf{W}$ dataset contains multiple sharp turns, demanding a more expressive learning model in comparison to $\mathsf{P}$ dataset.
Note that the more complex the vector field is, the more prominent the shortcomings of Euclidean approaches are.

\begin{figure}[t!]
\centering
    \includegraphics[width =.24\linewidth]{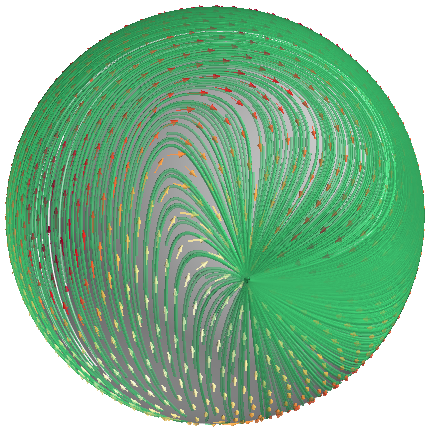}
    \includegraphics[width =.24\linewidth]{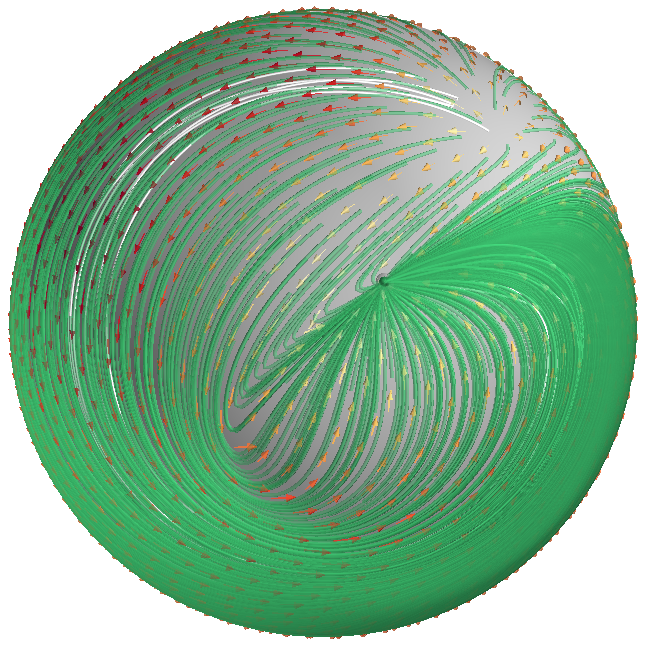}
    \includegraphics[width =.24\linewidth]{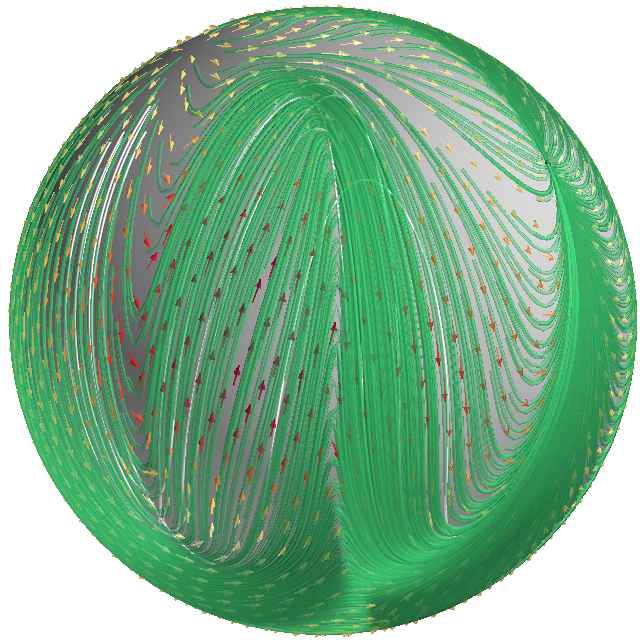}
    \includegraphics[width =.24\linewidth]{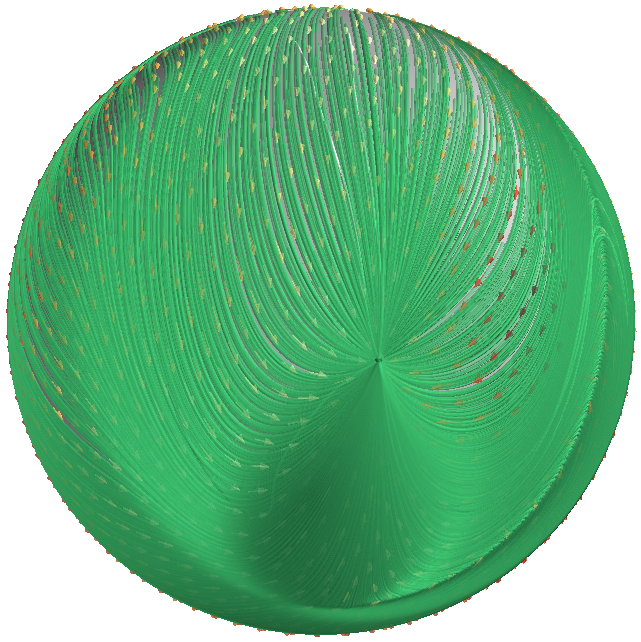}\\
    \includegraphics[width =.24\linewidth]{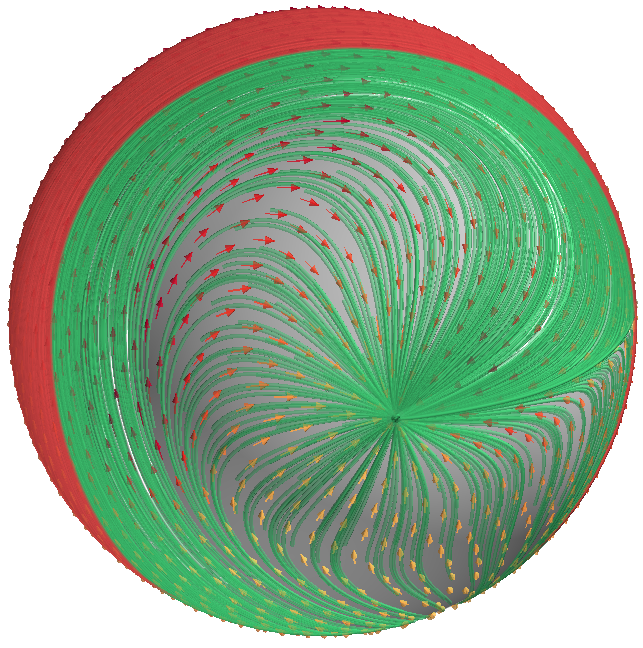}
    \includegraphics[width =.24\linewidth]{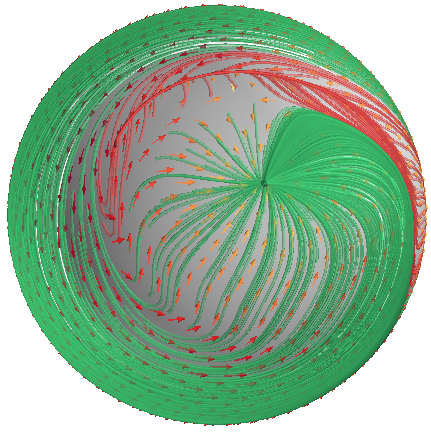}
    \includegraphics[width =.24\linewidth]{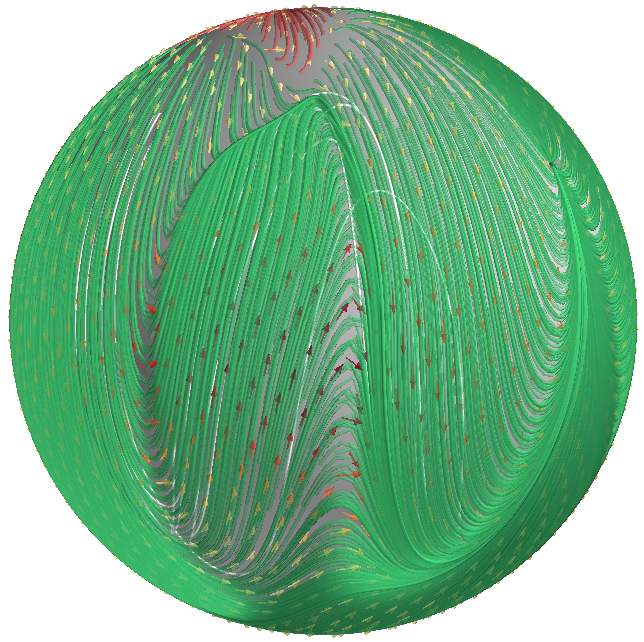}
    \includegraphics[width =.24\linewidth]{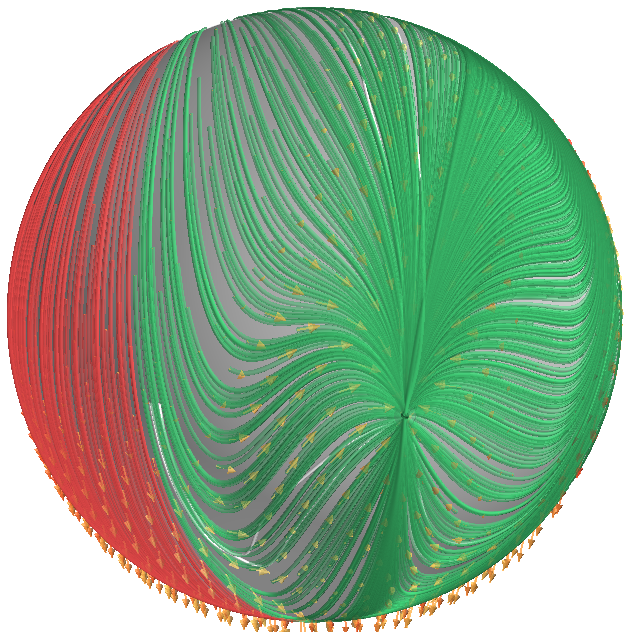}\\
    \includegraphics[width =.24\linewidth]{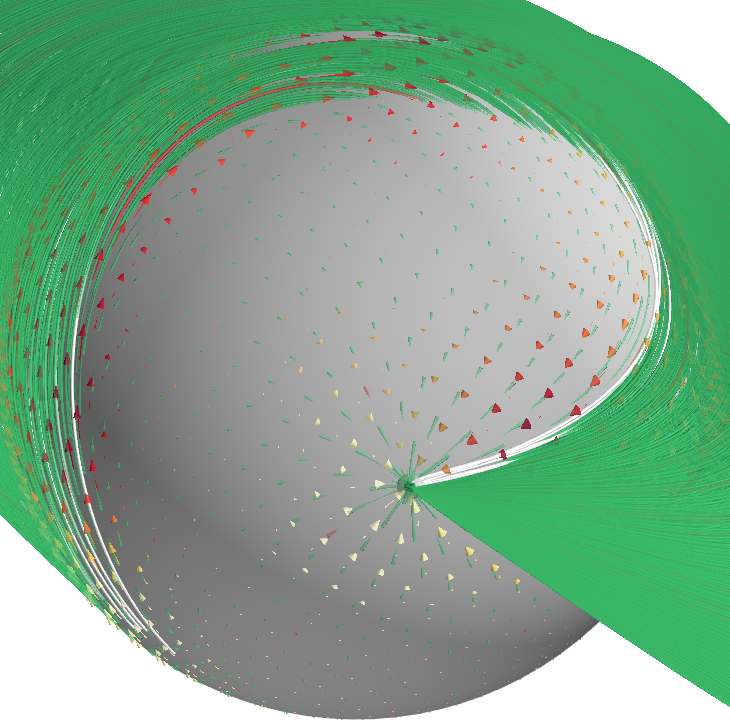}
    \includegraphics[width =.24\linewidth]{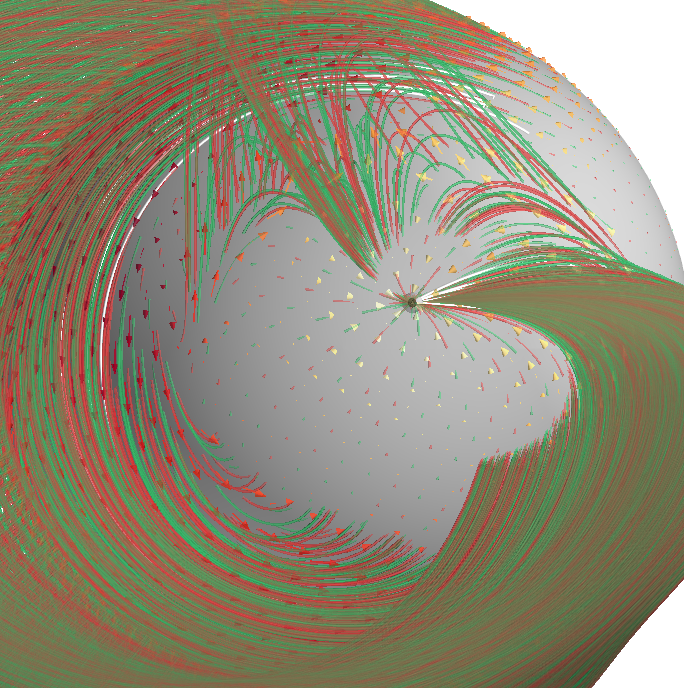}
    \includegraphics[width =.24\linewidth]{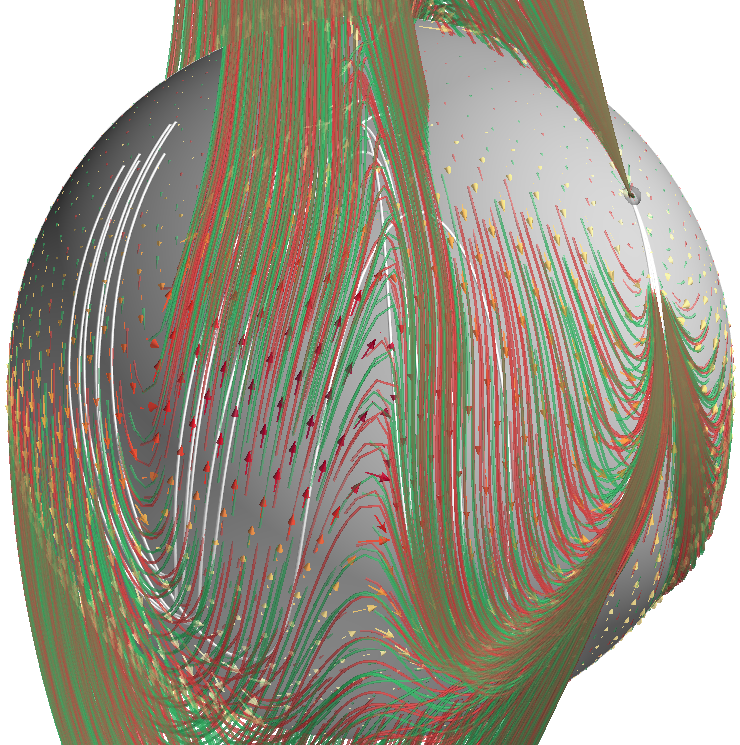}
    \includegraphics[width =.24\linewidth]{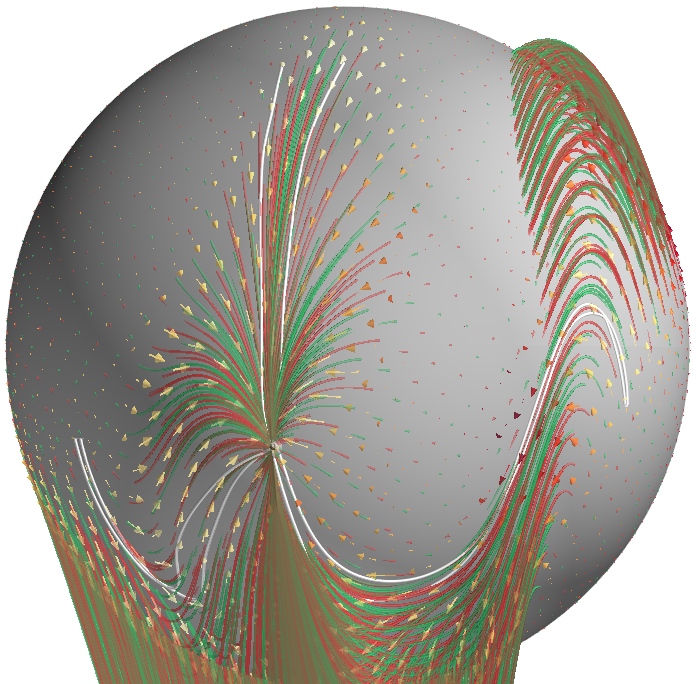}
    \caption{Evaluation of the reproduction stability on the learned vector fields. 
    $1000$ trajectories were generated using initial states uniformly sampled on $\SphereManifold^2$. The path integral for each initial state was computed on the vector field during a fixed time.  
    The successful and failed trajectories are indicated in green and red, respectively.
    From top to bottom, the rows corresponds to trajectories computed using RSDS, \textit{Projected EuclideanFlow}, and \textit{EuclideanFlow}.  
    From left to right, each column corresponds to $\mathsf{P}$, $\mathsf{G}$, $\mathsf{W}$, and $\mathsf{Multi Models}$ demonstrations.
    } 
    \label{fig:stability_appendix}
    \vspace{-0.2cm}
\end{figure}

\begin{figure}[h!]
	\centering
	\includegraphics[width =.24\linewidth]{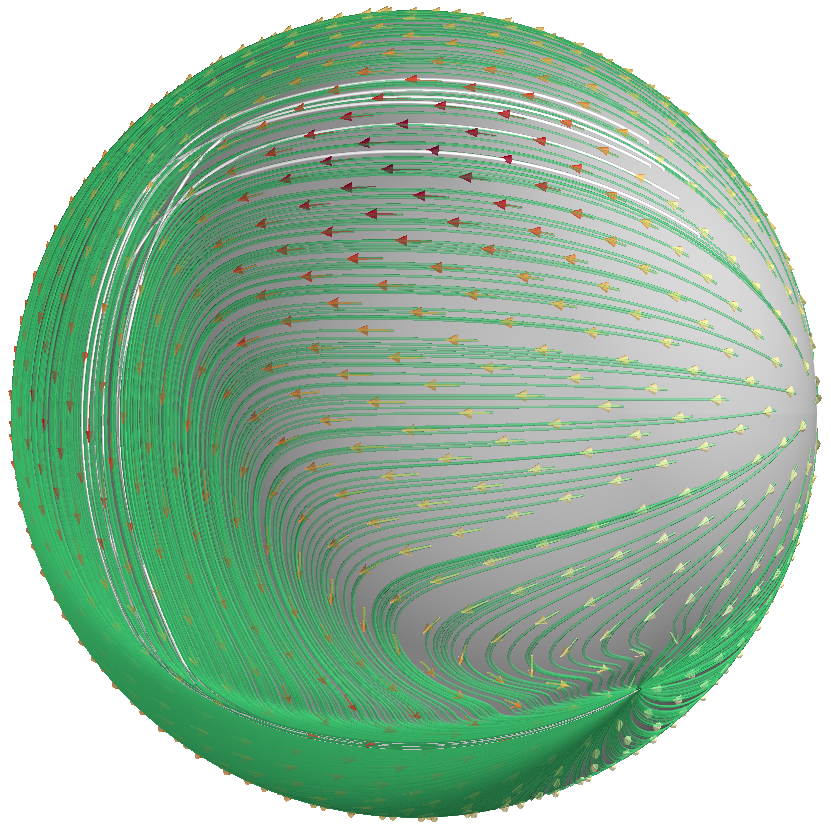}
	\includegraphics[width =.24\linewidth]{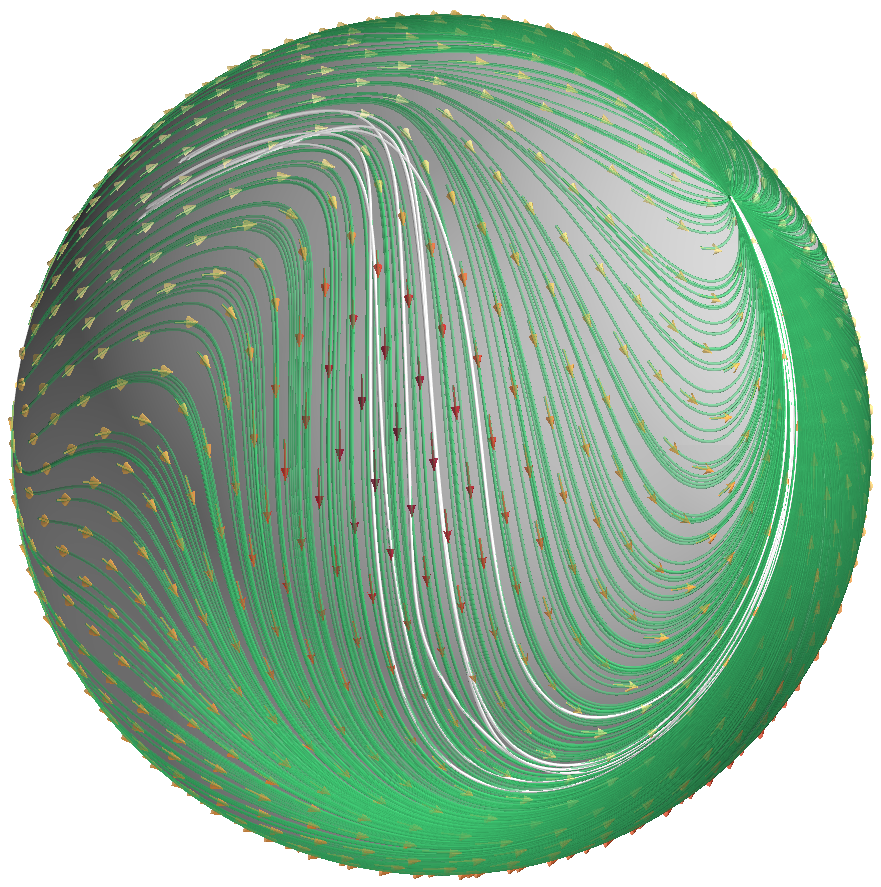}
	\includegraphics[width =.24\linewidth]{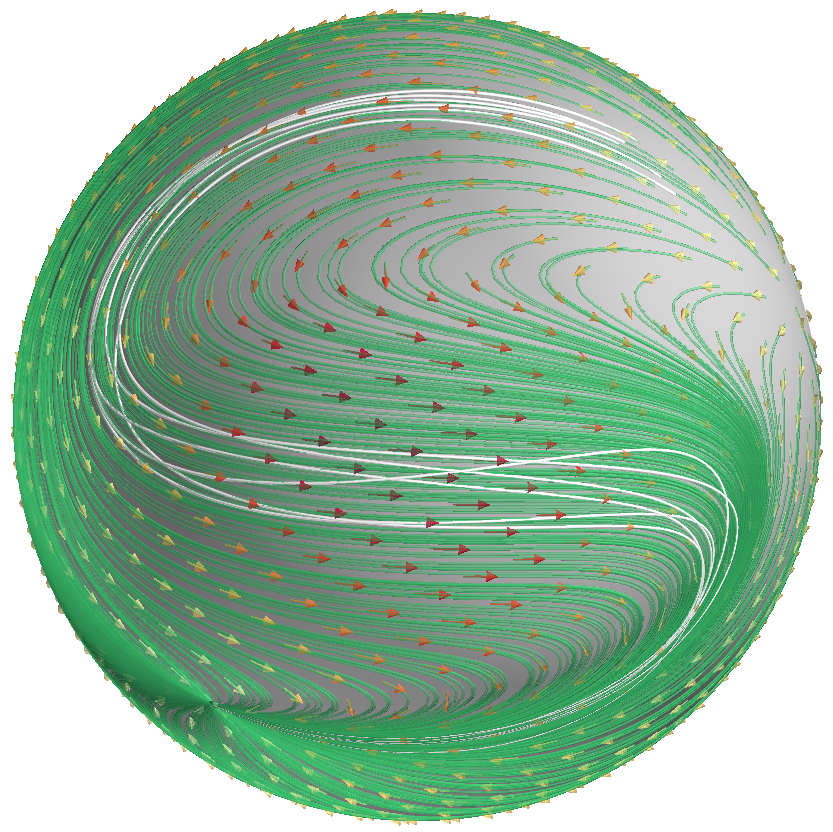}\\
	\includegraphics[width =.24\linewidth]{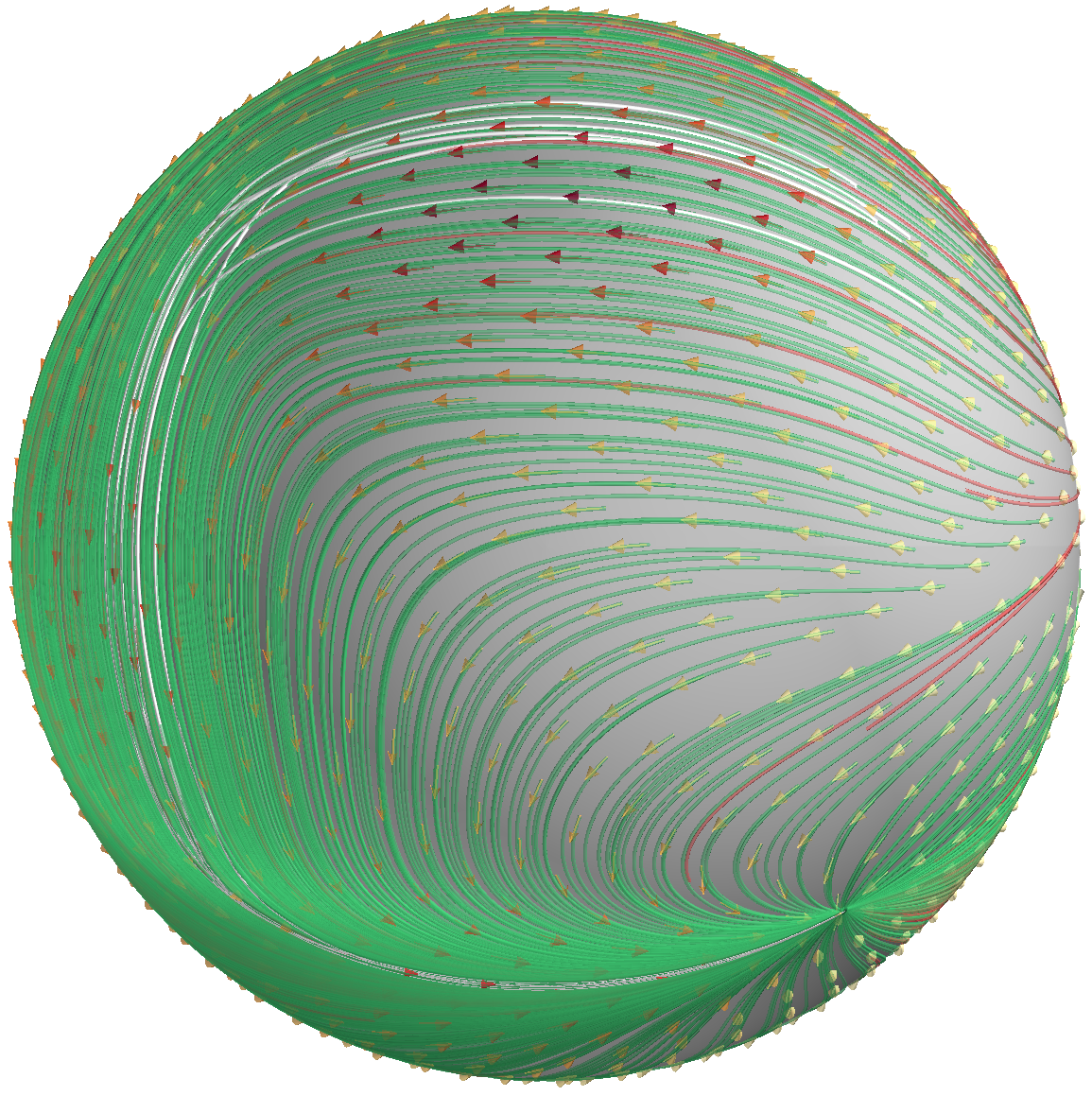}
	\includegraphics[width =.24\linewidth]{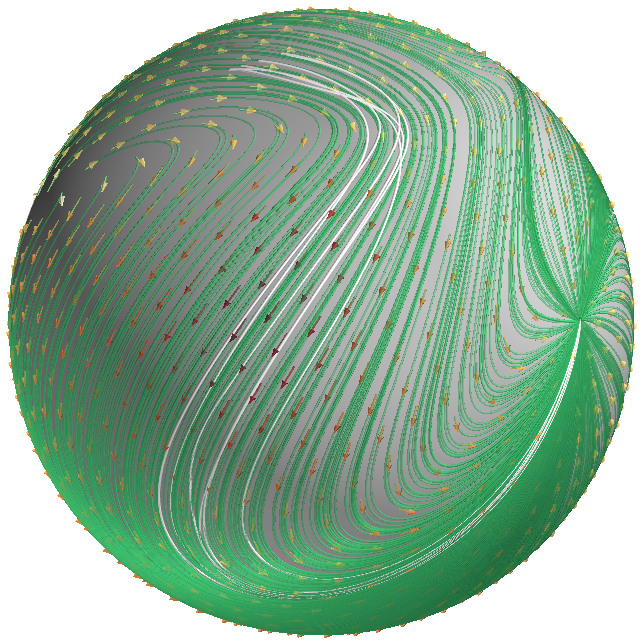}
	\includegraphics[width =.24\linewidth]{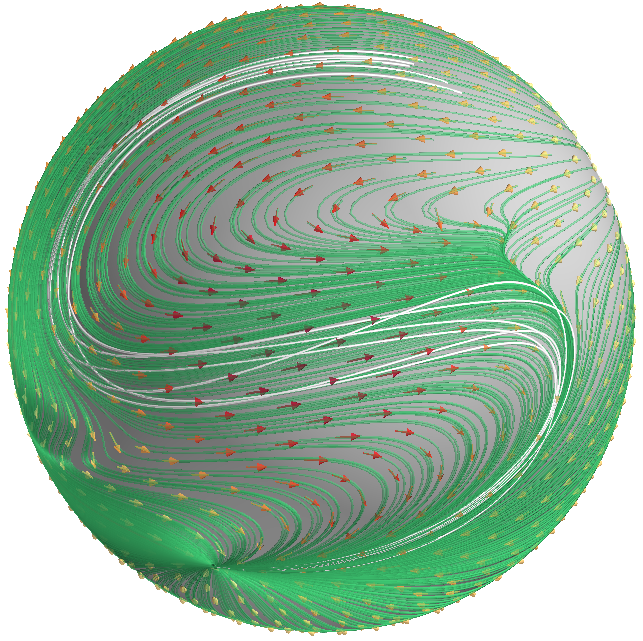}\\
	\includegraphics[width =.24\linewidth]{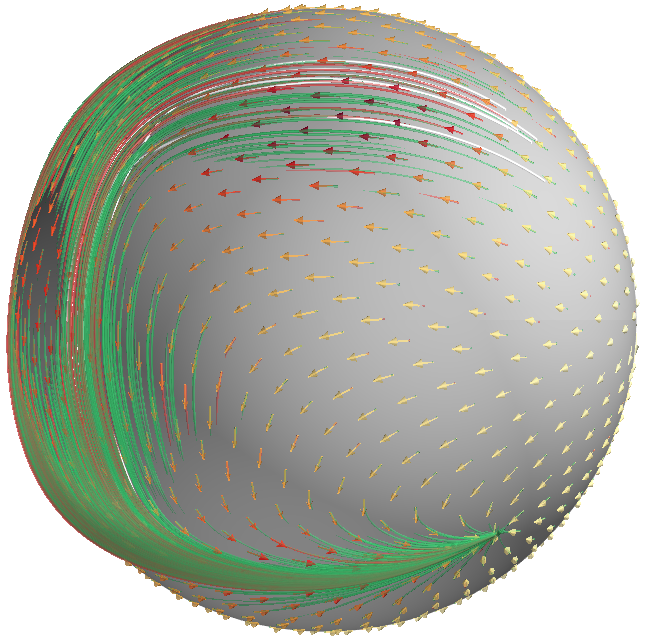}
	\includegraphics[width =.24\linewidth]{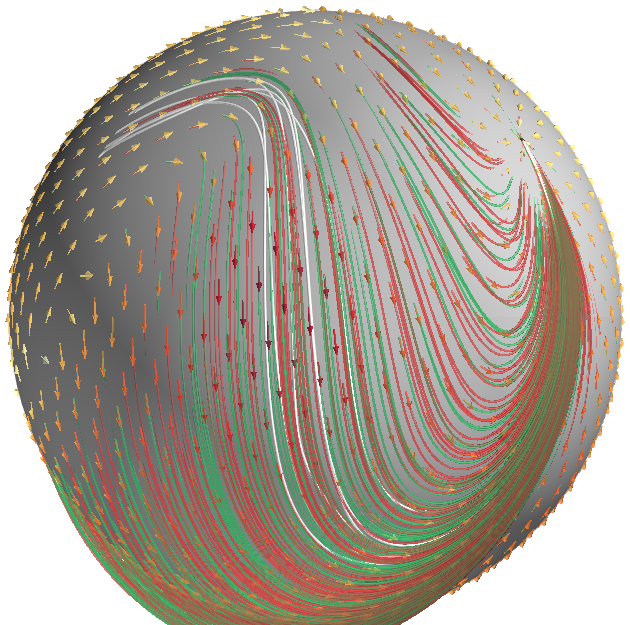}
	\includegraphics[width =.24\linewidth]{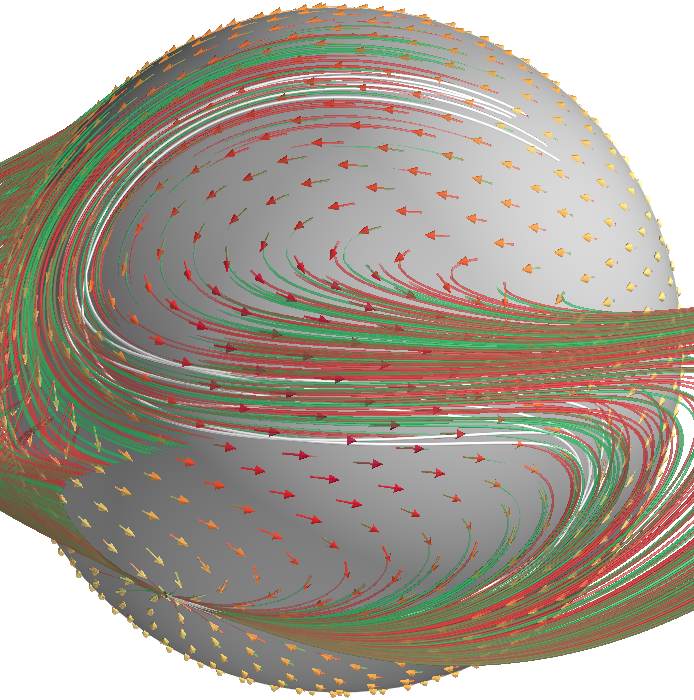}
	\caption{Additional evaluation of the reproduction stability on the learned vector fields. 
		Here, $1000$ trajectories were generated using initial states uniformly sampled on $\SphereManifold^2$. The path integral for each initial state was computed on the vector field during a fixed time.  
		The successful and failed trajectories are indicated in green and red, respectively.
		From top to bottom, the rows corresponds to trajectories computed using \textit{RSDS}, \textit{Projected EuclideanFlow}, and \textit{EuclideanFlow}.  
		From left to right, each column corresponds to $\mathsf{SharpC}$, $\mathsf{Spoon}$, and $\mathsf{S}$ demonstrations.
	} 
	\label{fig:stability_rebuttal}
\end{figure}

To quantitatively demonstrate the differences between \textit{RSDS} and Euclidean approaches (i.e., \textit{EuclideanFlow} and \textit{Projected EuclideanFlow}), Fig.~\ref{fig:metrics_rebuttal}-\emph{bottom-left}, and -\emph{top-right} show the \textit{dynamic time warping distance} (DTWD) as a measure of reproduced position trajectory accuracy, and the \textit{mean squared error} (MSE) of the velocities reproduction. Additionally, the success rate (i.e., convergence to the attractor) of our approach and the baseline can be seen in Fig.~\ref{fig:metrics_rebuttal}-\emph{top-left}, where \textit{RSDS} always reproduces stable trajectories. 
As mentioned in \S~\ref{sec:result}, for a fair comparison in stability analysis, we used the \emph{Projected EuclideanFlow} instead of \textit{EuclideanFlow}. 
To clearly show the intention behind this, Fig.~\ref{fig:stability_appendix} and Fig.~\ref{fig:stability_rebuttal} provide the results computed by \emph{EuclideanFlow} in the last row.  
Figures~\ref{fig:stability_appendix} and~\ref{fig:stability_rebuttal} show the stability evaluation of the reproduced trajectories on the learned vector field. 
Here, $1000$ trajectories were generated using initial states uniformly sampled on $\SphereManifold^2$. 
The successful and failed trajectories are colored as green and red, respectively. 
From top to bottom, the rows correspond to trajectories computed using RSDS, \textit{Projected EuclideanFlow}, and \textit{EuclideanFlow}, respectively.
It is evident that all the \textit{RSDS} trajectories succeeded, while some the \textit{EuclideanFlow} trajectories failed to converge despite the projection.

\begin{figure}
	\centering
	\includegraphics[width =.24\linewidth]{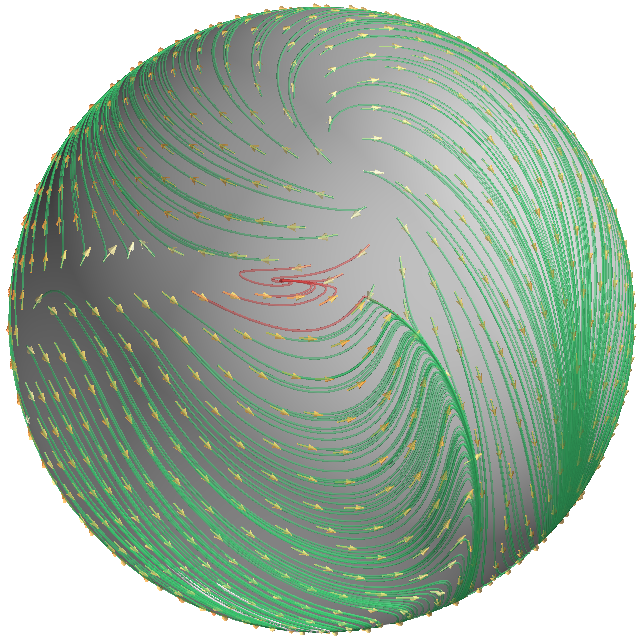}
	\includegraphics[width =.24\linewidth]{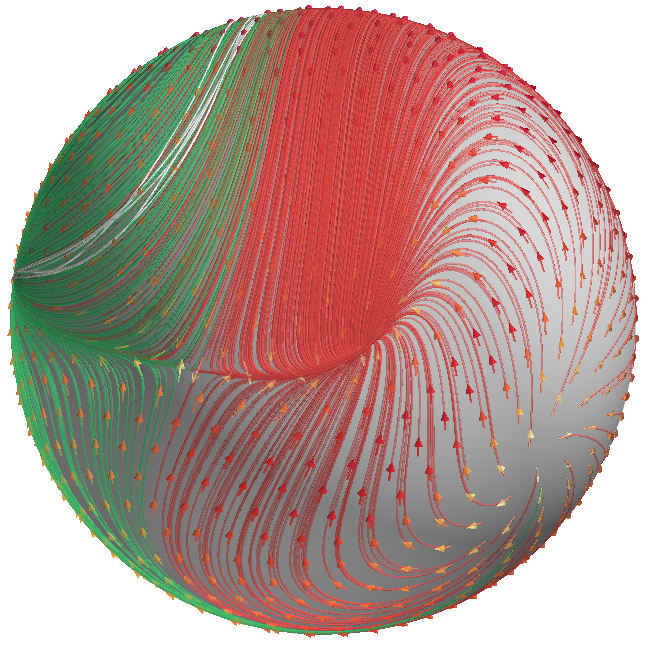}
	\includegraphics[width =.24\linewidth]{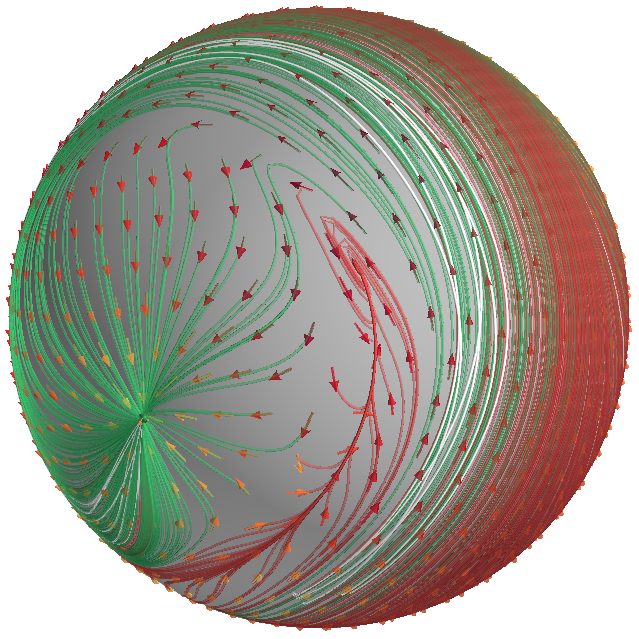}  
	\includegraphics[width =.24\linewidth]{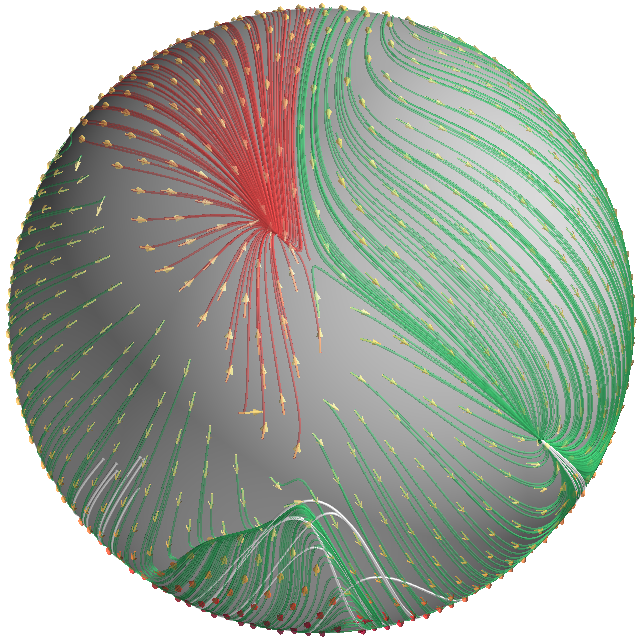} 
	\caption{Spurious attractors on vector fields learned with geometry-unaware \textit{Projected EuclideanFlow}. Plots show a different view angle of the reproduction stability on the learned vector fields using \textit{Projected EuclideanFlow}. From left to right, each plot corresponds to $\mathsf{Spoon}$, $\mathsf{P}$, $\mathsf{G}$, and $\mathsf{W}$ datasets.}
	\label{fig:spurious_attractor}
\end{figure}

In order to further show how unstable trajectories arise, we provide additional plots displaying a different view of some examples of $\textit{Projected EuclideanFlow}$ in Fig~\ref{fig:spurious_attractor}.
We can see that additional spurious attractor points can appear on $\SphereManifold^2$ if we do not explicitly consider the geometric constraints of the data. Then, instead of converging to the true attractor, unstable trajectories diverge to these undesired points.
This phenomenon might not occur when learning vector fields displaying simple dynamics, but in general, geometry-unaware methods may be prone to it.
Therefore, when using $\textit{EuclideanFlow}$, we cannot provide stability guarantees on Riemannian manifolds and this may lead to catastrophic results for real-world applications. 
However, this issue can be theoretically and practically overcome by $\textit{RSDS}$.
Additionally, when compared against $\textit{projected EuclideanFlow}$, $\textit{RSDS}$ usually generates more consistent and smoother trajectories around the demonstrations, which can be clearly observed, for example, for the $\mathsf{S}$ shape.

\subsection{Real robot experiments on $\RtimeS$}
\subsubsection{Data Pre-processing:}
\begin{figure}[th!]
\centering
    \includegraphics[width =.31\linewidth]{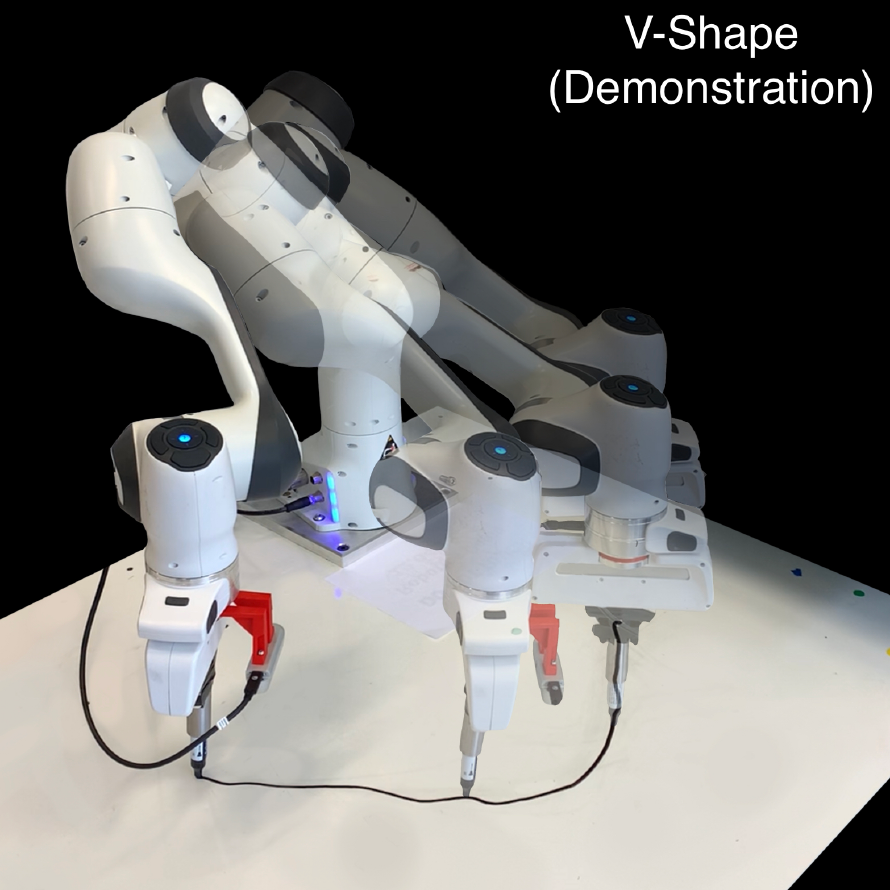}
    \includegraphics[width =.31\linewidth]{Images/Experiments/Robot/V/Reproduction_V.pdf}
    \includegraphics[width =.31\linewidth]{Images/Experiments/Robot/V/Perturbation_V.pdf}\\
    \includegraphics[width =.31\linewidth]{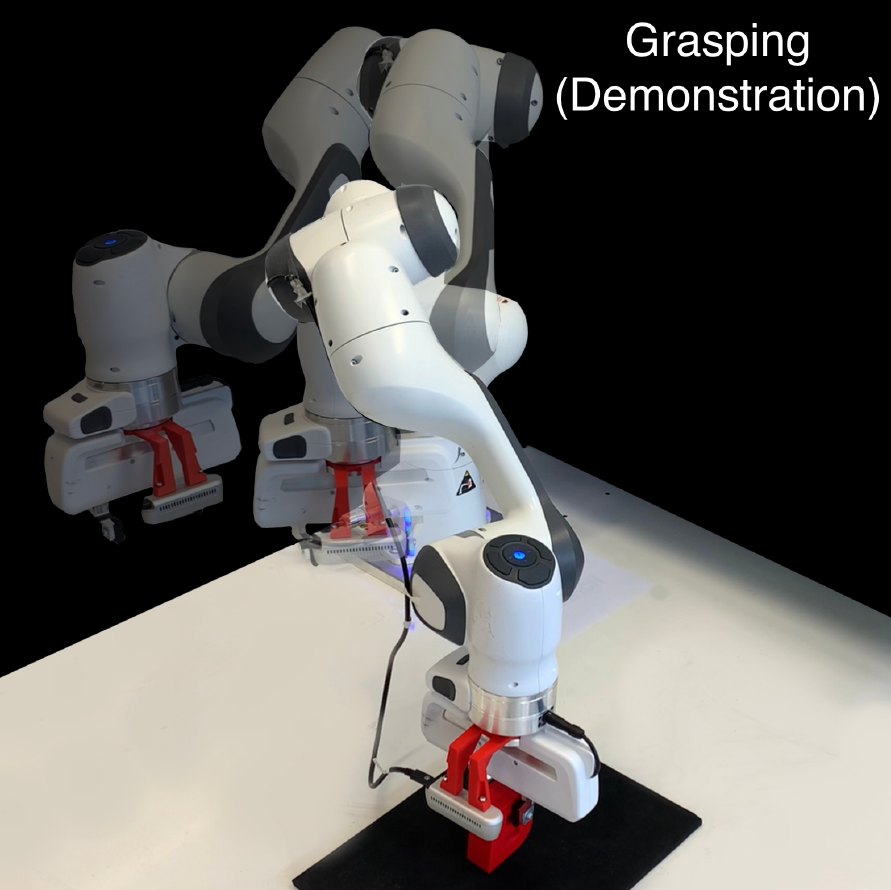}
    \includegraphics[width =.31\linewidth]{Images/Experiments/Robot/Grasp/Reproduction_Grasping_2.pdf}
    \includegraphics[width =.31\linewidth]{Images/Experiments/Robot/Grasp/Perturbation_Grasping_2.pdf}
    \caption{\textit{Left:} The demonstrated trajectories depicted by superimposing images from different time frames. The transparent robot arms depict the trace of the motion trajectory. Here, the robot performs $\mathsf{V}$-shape $\mathsf{DrawingTask}$ (top row) and $\mathsf{GraspingTask}$ with $90$ degrees rotation (bottom row). \textit{Middle:} The motion reproduced using learned vector field on the product manifold $\RtimeS$. \textit{Right:} The motion reproduced using learned vector field on the product manifold $\RtimeS$ when applying perturbation.}
    \label{fig:robot_exp_appendix}
\end{figure}

Before using the demonstrations collected through kinesthetic teaching, a low-pass filter was applied to smooth out the trajectories.
In addition, the trajectories are slightly shifted to guarantee that all demonstrations converge to common target. 
Although, this alteration in position trajectories can be achieved using classical parallel shifting, for quaternion trajectories the \textit{parallel transport} operator on a $3$-Sphere is required to accomplish the shifting operation. 
For sake of completeness, Fig.~\ref{fig:robot_exp_appendix} shows the replay of one of the demonstrations (left picture), as well as the normal and perturbed reproductions shown in Fig.~\ref{fig:robot_exp}. 

\subsection{Stability on Robotic Tasks}
\begin{figure}
	\centering
	\begin{subfigure}[b]{\textwidth}
		\centering
		\includegraphics[width=\textwidth]{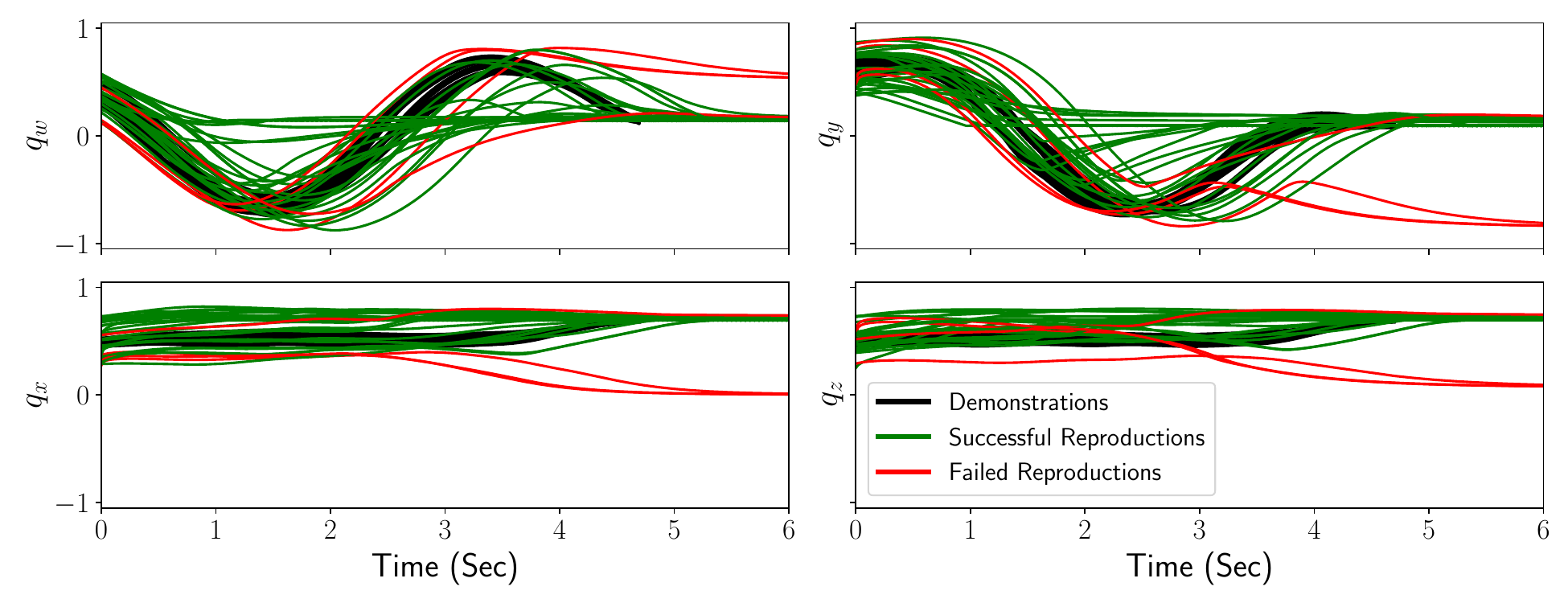}
		\caption{Reproduction of quaternion trajectories with \textit{Projected EuclideanFlow}.}
		\label{fig:quaternion_trajectory_G_EF}    
	\end{subfigure}
	\begin{subfigure}[b]{\textwidth}
		\centering
		\includegraphics[width=\textwidth]{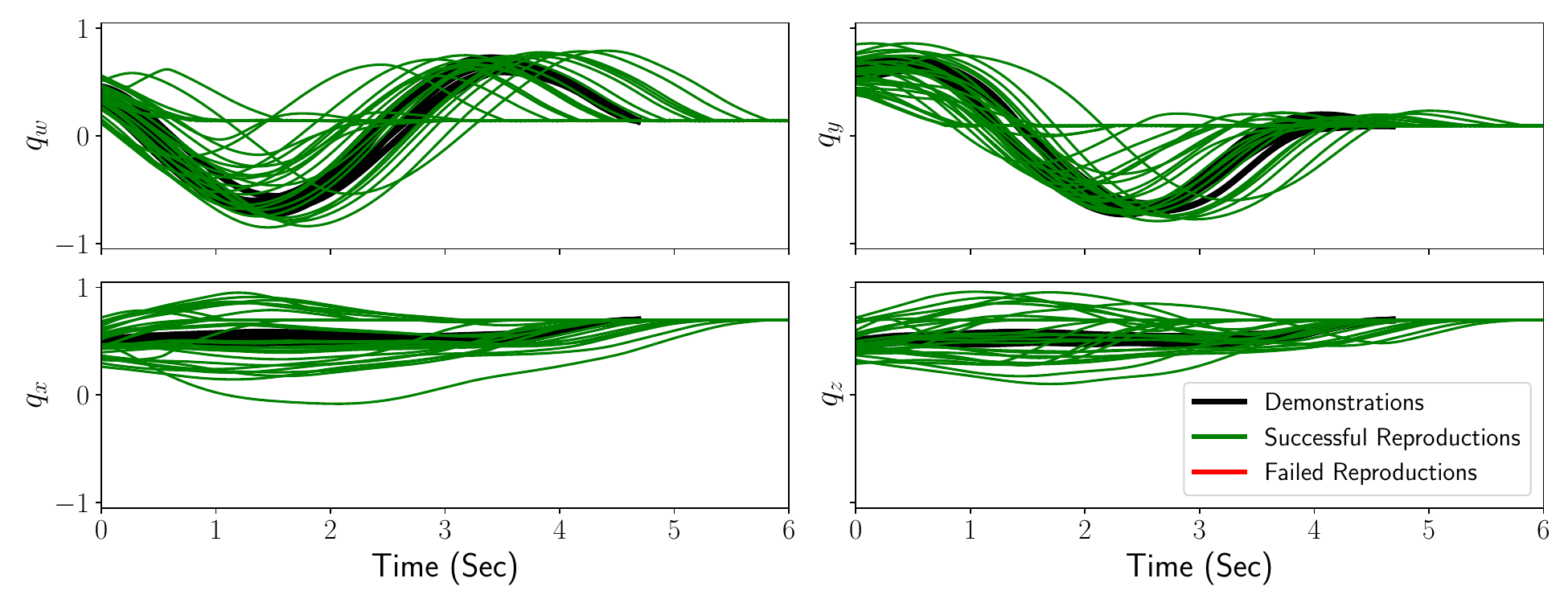}
		\caption{Reproduction of quaternion trajectories with \textit{RSDS}.}
		\label{fig:quaternion_trajectory_G_RSDS}    
	\end{subfigure}
	\caption{Time-series plot of quaternion trajectories obtained from the first-order dynamical systems learned by \textit{Projected EuclideanFlow} and \textit{RSDS}, corresponding to Experiment $1$ (see Fig.~\ref{subfig:stability_robot1}). The quaternion trajectories reproduced by \textit{Projected EuclideanFlow} reveal several integral curves that diverge, therefore winding up distant from the target. These results show the inability to reconstruct stable quaternion trajectories when disregarding the geometric constraints arising from $\mathcal{S}^3$. In contrast, the \textit{RSDS} integral curves succeed to converge and reach the target, demonstrate its ability to reconstruct stable quaternion trajectories on $\mathcal{S}^3$.}
	\label{fig:quaternion_trajectory_G}    
\end{figure}

To provide further evidence for the practical use of \textit{RSDS}, we design new experiments on simulated robotic tasks, where we only require the end-effector to follow demonstrated quaternion trajectories while keeping its position fixed. 
The quaternion demonstrations are synthetically generated 
on $\mathcal{S}^3$.
We apply the same experimental settings as the ones described for the illustrative LASA dataset experiments on $\mathcal{S}^2$, detailed in \S~\ref{subsec:ExpSphere}, for both \textit{RSDS} and \textit{EuclideanFlow}, except that an additional hidden layer is added to improve models' expressiveness.
We reproduce several quaternion trajectories, using the learned vector fields, starting from randomly-sampled initial points around the demonstrations for both \textit{RSDS} and \textit{Projected EuclideanFlow}. 
For the latter, we project the learned vector fields onto $\mathcal{S}^3$, similarly to the experiments reported in \S~\ref{subsec:ExpSphere} of the main paper.
Note that this projection is necessary to guarantee that the resulting integral curves lead to proper unit quaternion references. 
Additionally, we use a Cartesian impedance controller to track the reference signals.
Figure~\ref{fig:quaternion_trajectory_G} shows the reconstructed trajectories on $\SphereManifold^3$ using \textit{Projected EuclideanFlow} corresponding to the $\mathsf{G}$ dataset. 

\begin{figure}
	\centering
	\begin{subfigure}[]{0.495\textwidth}
		\centering
		\includegraphics[width = .49\textwidth]{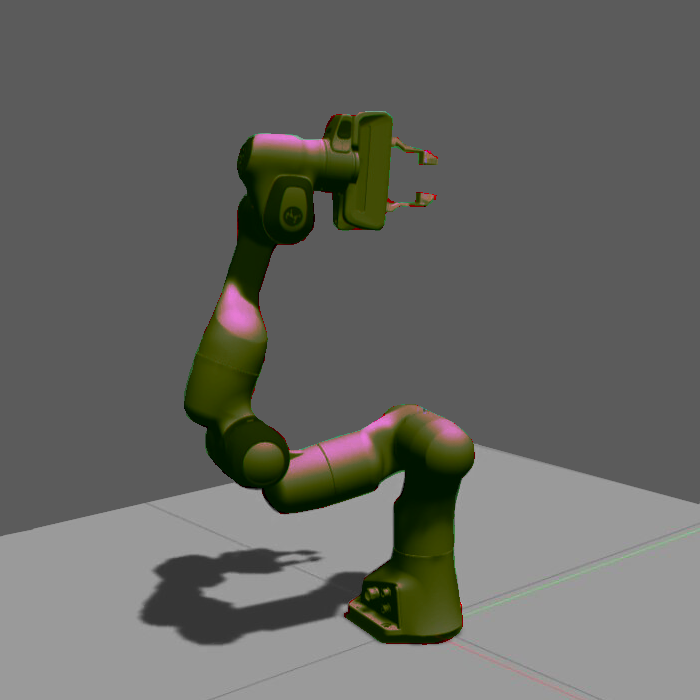}
		\includegraphics[width = .49\textwidth]{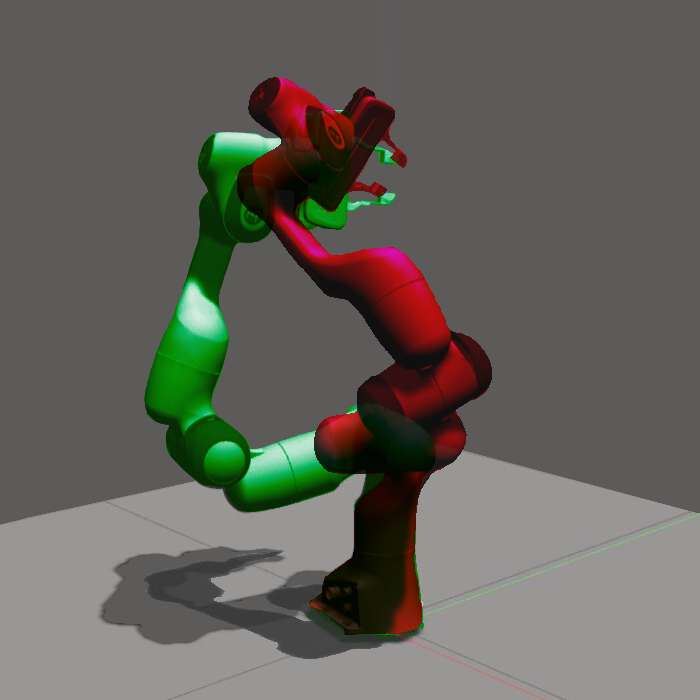}
		\caption{Experiment $1$}
		\label{subfig:stability_robot1}
	\end{subfigure}
	\begin{subfigure}[]{0.495\textwidth}
		\centering
		\includegraphics[width = .49\textwidth]{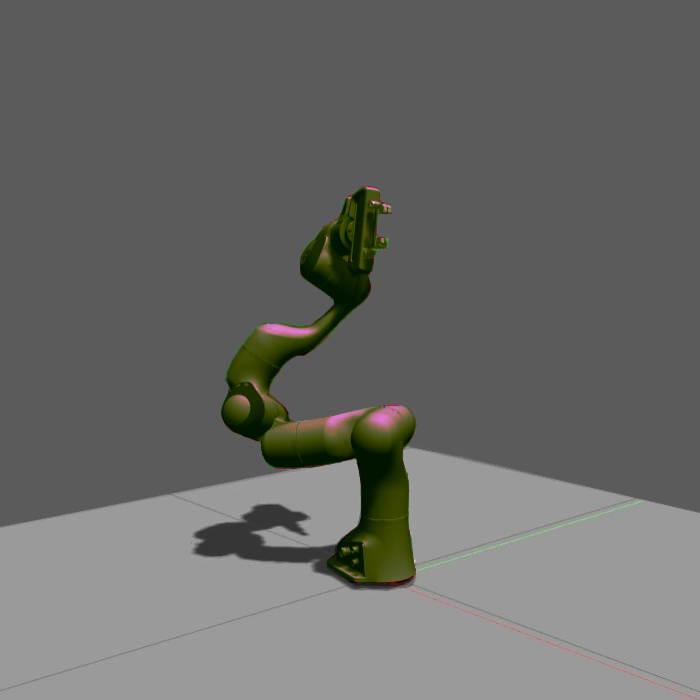}
		\includegraphics[width = .49\textwidth]{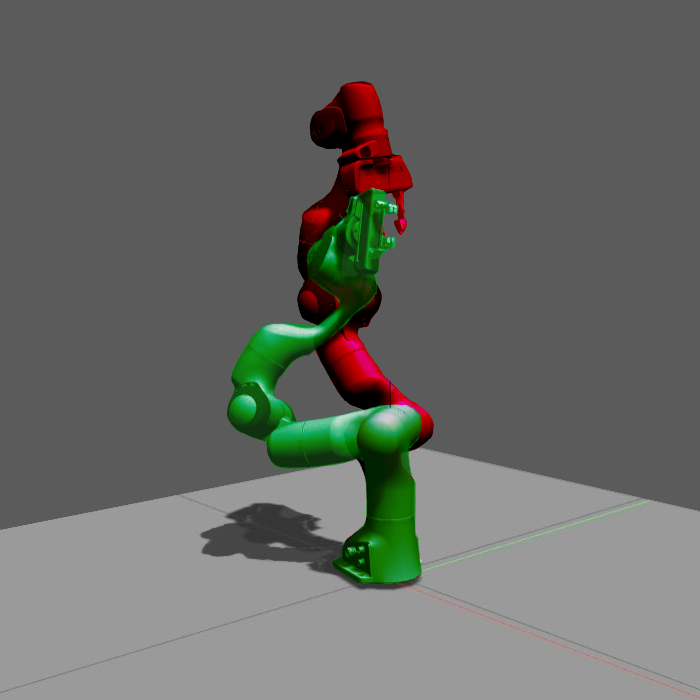}
		\caption{Experiment $2$}
		\label{subfig:stability_robot2}
	\end{subfigure}
	\caption{Stability experiments on a simulated Franka-Emika robot. The green robotic arm represents the end-effector pose target extracted from demonstrations, and the red robot displays the final end-effector pose achieved during reproduction. For each experiment, the \textbf{left} snapshot displays the \textit{RSDS} reproduction, and the \textbf{right} picture shows the \textit{EuclideanFlow} result.}
	\label{fig:stability_robot}
\end{figure}

In Fig.~\ref{fig:quaternion_trajectory_G_EF}, we can observe that some trajectories (depicted in red) generated by \textit{Projected EuclideanFlow} do not to reach the target if the initial point moderately differs from the demonstrations, which often happens in real-world settings. In contrast, \textit{RSDS} succeeds in reproducing stable trajectories despite the initial conditions being different to the training data, as shown in Fig.~\ref{fig:quaternion_trajectory_G_RSDS}.
To visualize the effects that this may have in real-world applications, we show the final robot end-effector pose reached by our \textit{RSDS} method and \textit{Projected EuclideanFlow} for two different simulated experiments in 
Fig.~\ref{fig:stability_robot}, where Figs.~\ref{subfig:stability_robot1} and~\ref{subfig:stability_robot2} correspond to the experiments reported in Figs.~\ref{fig:quaternion_trajectory_G_EF} and~\ref{fig:quaternion_trajectory_G_RSDS}, respectively.
As pointed out previously, \textit{Projected EuclideanFlow} diverges and therefore reaches an undesired final end-effector pose. However, the final end-effector poses reached by \textit{RSDS} are in sharp contrast to the \textit{Projected EuclideanFlow} results, as our method successfully converges to the desired target (represented by the green robot arm in Fig.~\ref{fig:stability_robot}).

\subsection{Runtime of RSDS}
We measured the runtime for \textit{RSDS} and \textit{EuclideanFlows} on a PC with Intel Xeon W-1250P CPU and $31$ Gigabytes of memory.    
\color{black}
Table~\ref{tab:runtime} provides reference values of runtime and shows that \textit{RSDS} runs around $2$ times slower than \textit{EuclideanFlow}, as the \textit{RSDS} model requires solving IVPs on Riemannian manifolds. However, there is still a lot of room for improvement in terms of speeding up the \textit{RSDS} implementation, which is worthwhile for future research.
\begin{table}[h]
	\centering
	\begin{tabular}{@{}lllll@{}}
		\toprule
		& 1 & 2 & 3 & Mean \\ \midrule
		RSDS & 198 & 200 & 197 & 198 \\
		EF & 90 & 89 & 91 & 90 \\ \bottomrule
	\end{tabular}
	\caption{The average runtime (in milliseconds) of \textit{RSDS} and \textit{EuclideanFlow} over $100$ iterations in three separate experiments.}
	\label{tab:runtime}
\end{table}

\end{document}